\documentclass[lettersize,journal]{IEEEtran}
\usepackage{amsmath,amsfonts}
\usepackage{algorithmic}
\usepackage{algorithm}
\usepackage{array}
\usepackage{textcomp}
\usepackage{stfloats}
\usepackage{url}
\usepackage{verbatim}
\usepackage{graphicx}
\usepackage{cite}
\usepackage{hyperref}
\usepackage{booktabs}
\usepackage{multirow}
\usepackage{makecell}
\usepackage{color}

\begin{document}
	
	\title{TRNR: Task-Driven Image Rain and Noise Removal with a Few Images Based on Patch Analysis}
	
	\author{Wu~Ran,~\IEEEmembership{Graduate Student Member,~IEEE,}
		Bohong~Yang,~\IEEEmembership{Student Member,~IEEE,}
		Peirong~Ma, ~\IEEEmembership{Student Member,~IEEE}
		and~Hong~Lu,~\IEEEmembership{Member,~IEEE}}

%

\maketitle

\begin{abstract}
	The recent success of learning-based image rain and noise removal can be attributed primarily to well-designed neural network architectures and large labeled datasets. However, we discover that current image rain and noise removal methods result in low utilization of images. To alleviate the reliance of deep models on large labeled datasets, we propose the task-driven image rain and noise removal (TRNR) based on a patch analysis strategy. The patch analysis strategy samples image patches with various spatial and statistical properties for training and can increase image utilization. Furthermore, the patch analysis strategy encourages us to introduce the N-frequency-K-shot learning task for the task-driven approach TRNR. TRNR allows neural networks to learn from numerous N-frequency-K-shot learning tasks, rather than from a large amount of data. To verify the effectiveness of TRNR, we build a Multi-Scale Residual Network (MSResNet) for both image rain removal and Gaussian noise removal. Specifically, we train MSResNet for image rain removal and noise removal with a few images (for example, 20.0\% train-set of Rain100H). Experimental results demonstrate that TRNR enables MSResNet to learn more effectively when data is scarce. TRNR has also been shown in experiments to improve the performance of existing methods. Furthermore, MSResNet trained with a few images using TRNR outperforms most recent deep learning methods trained data-driven on large labeled datasets. These experimental results have confirmed the effectiveness and superiority of the proposed TRNR. The source code is available on \url{https://github.com/Schizophreni/MSResNet-TRNR}.
\end{abstract}


\section{Introduction}
\label{sec:intro}
\IEEEPARstart{C}{omputer} vision algorithms such as autonomous driving\cite{AutonomousDriving}, semantic segmentation\cite{SemanticSeg}, and object tracking\cite{ObjectTracking} require clean images as input, and thus tend to fail under bad weather conditions. Image rain and noise removal, which are dedicated to restoring clean images from degraded observations, serve as the indispensable pre-processing process for those computer vision algorithms. Specifically, image rain and noise removal focus on removing the rain streaks \cite{Li2016GMM,Yang2019acm,Fu2011Decomp,Fu2017DDN,RenGF2018,JORDERE,JDDGD} and removing noise \cite{DIP2018,Zhang2018FFDNet,KaiZhangDnCNN2017,LiuMWCNN} from contaminated images respectively via solving a linear decomposition problem.

Recent research on image rain and noise removal can be categorized into the prior-based approaches and the data-driven approaches \cite{Derainsurvey}. The prior-based approaches remove rain and noise from images by solving a linear decomposition problem with manually designed priors \cite{Li2016GMM, Gu2017JCAS, Fu2011Decomp, dabovBM3D}. While the data-driven approaches, e.g., \cite{Fu2017DDN, KaiZhangDnCNN2017, Zhang2018FFDNet, Yang2017Jorder} enable models to learn more robustly and more flexibly with the help of large labeled datasets, which usually result in robust models with better performance. However, these data-driven approaches for image rain removal \cite{Xia2018Rescan, Wang2019SPAnet,Yang2019acm, JDNet,JDDGD,BRN2020TIP} and noise removal \cite{KaiZhangDnCNN2017, Zhang2018FFDNet, LiuMWCNN,PaCNet,BUIFD} can not be adapted to limited data occasions due to over-fitting and low generalization ability.

\begin{figure}[!t]
	\centering
	\setlength{\abovecaptionskip}{0.01in}
	\begin{minipage}[t]{0.24\linewidth}
		\centering
		\includegraphics[width=2.1cm]{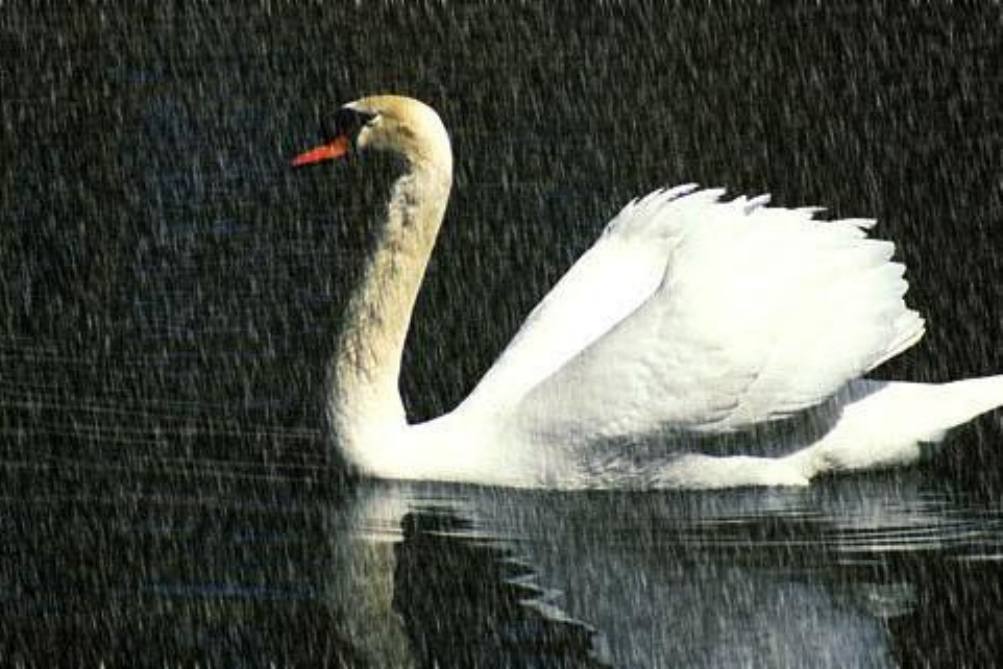}\vspace{3pt}
		\footnotesize{(a)}\vspace{5pt}
	\end{minipage}
	\begin{minipage}[t]{0.24\linewidth}
		\centering
		\includegraphics[width=2.1cm]{figures/rain-015.pdf}\vspace{3pt}
		\footnotesize{(b)}\vspace{5pt}
	\end{minipage}
	\begin{minipage}[t]{0.24\linewidth}
		\centering
		\includegraphics[width=2.1cm]{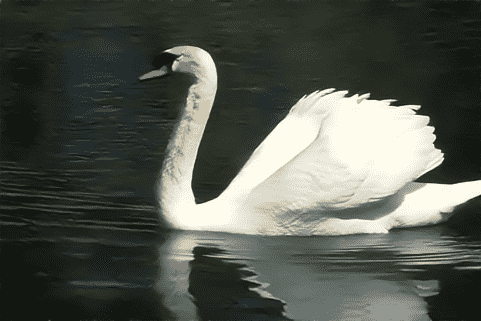}\vspace{3pt}
		\footnotesize{(c)}\vspace{5pt}
	\end{minipage}
	\begin{minipage}[t]{0.24\linewidth}
		\centering
		\includegraphics[width=2.1cm]{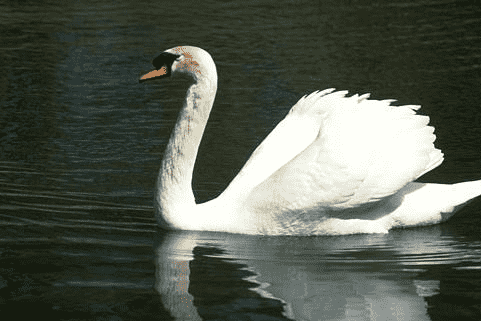}\vspace{3pt}
		\footnotesize{(d)}\vspace{5pt}
	\end{minipage}\\ 
	\begin{minipage}[t]{0.24\linewidth}
		\centering
		\includegraphics[width=2.1cm]{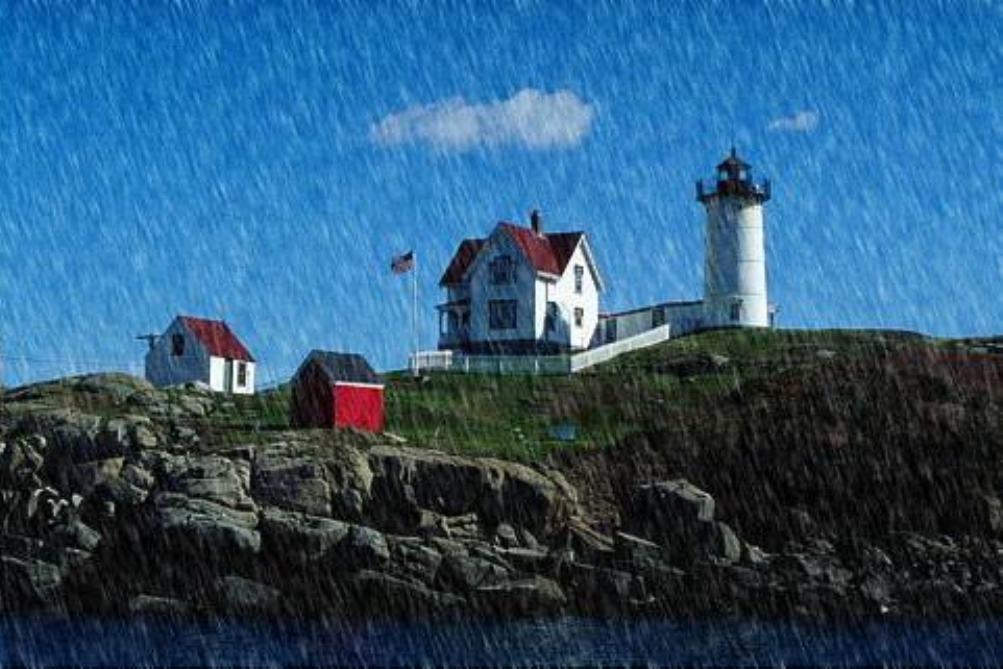}\vspace{-1pt}
		\footnotesize{(e). Rainy}\vspace{5pt}
	\end{minipage}
	\begin{minipage}[t]{0.24\linewidth}
		\centering
		\includegraphics[width=2.1cm]{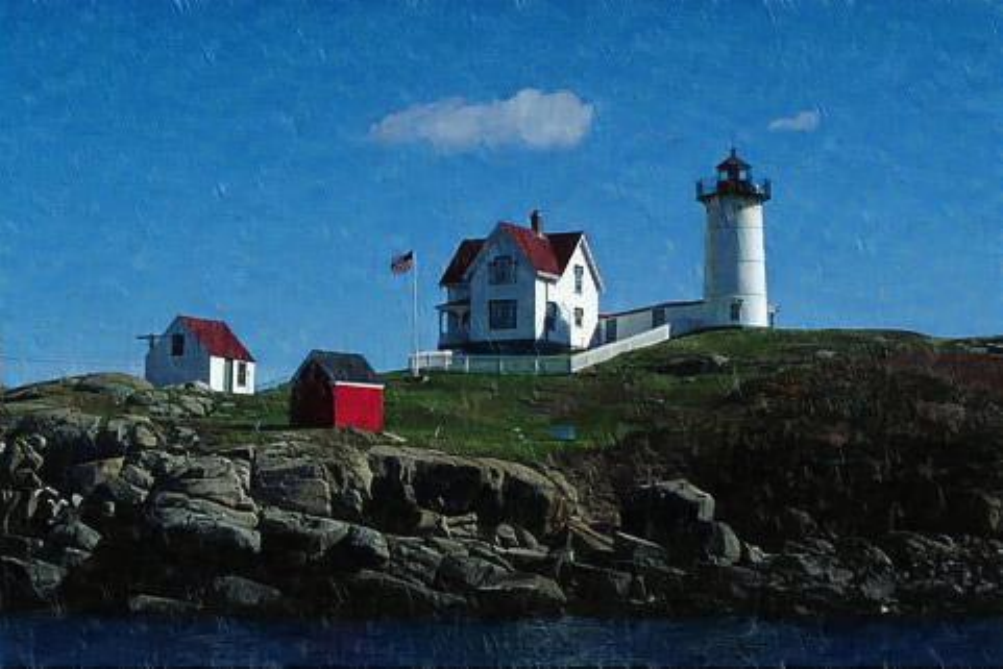}\vspace{-1pt}
		\footnotesize{(f). ReHEN~\cite{Yang2019acm}}\vspace{5pt}
	\end{minipage}
	\begin{minipage}[t]{0.24\linewidth}
		\centering
		\includegraphics[width=2.1cm]{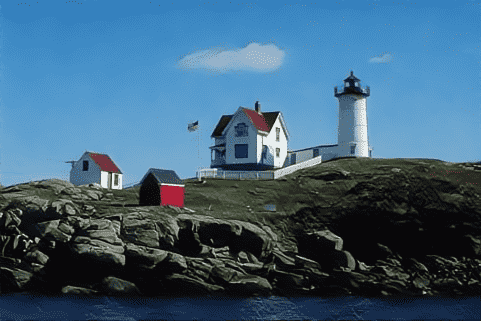}\vspace{-1pt}
		\footnotesize{(g). Ours}
	\end{minipage}
	\begin{minipage}[t]{0.24\linewidth}
		\centering
		\includegraphics[width=2.1cm]{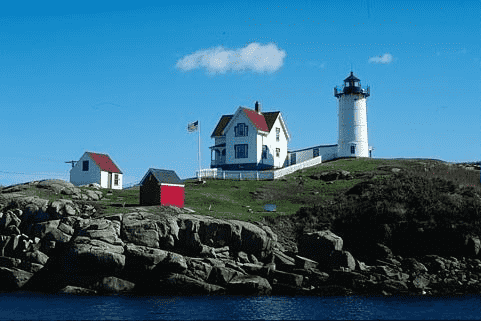}\vspace{-1pt}
		\footnotesize{(h). GT}
	\end{minipage}
	\caption{Image rain removal examples. (a) \& (e) represent the image contaminated by complex and accumulated rain streaks. (b) \& (f) indicate the image rain removal results of Yang et al. \cite{Yang2019acm} that use the full train-set (700 images) for training. (c) \& (g) are results of MSResNet (ours) that are trained with only 280 images. (d) \& (h) are the Ground Truth.}
	\label{fig: visual-comparison}
\end{figure}

Though data-driven approaches have made splendid progress on different benchmarks~\cite{Derainsurvey,Derainbenchmark}, they have not made full use of large labeled datasets by employing the \textit{random image sampling} (RIS) strategy to provide data for training. RIS leads to low utilization of image patches and neglects the spatial and statistical differences between different kinds of image patches. In practice, we observe that the performances of image rain and noise removal models fluctuate when processing image patches with various spatial and statistical properties. This observation motivates us to discriminatively sample image patches for training. Therefore, we propose a patch analysis strategy that groups image patches into different clusters based on the spatial and statistical properties. Then we introduce the \textit{random cluster sampling} (RCS) strategy to provide data for training by randomly choosing image patches from different clusters, which is verified to make better use of image patches.

To overcome the data limitation problem for learning image rain and noise removal, we propose a task-driven approach to substitute the current data-driven approach for training neural networks. By considering image patches discriminatively according to spatial and statistical properties, we introduce the N-frequency-K-shot learning task. The N-frequency-K-shot learning task aims to learn image rain and noise removal on a tiny dataset consisting of image patches from $N$ clusters and each cluster with $K$ examples. During the training process, the neural network can learn from numerous N-frequency-K-shot tasks instead of learning from plenty of data. Concretely, the neural network first adapts itself to a batch of N-frequency-K-shot learning tasks, then it summaries the knowledge obtained from all tasks to make a significant update. The training process of neural networks with the TRNR approach is similar to the Model Agnostic Meta-Learning (MAML) which focuses on fast adaptation to unknown tasks. To prove the effectiveness of the proposed TRNR approach, we design a light Multi-Scale Residual Network (MSResNet) with about 0.9M parameters for image rain removal. And we use MSResNet with about 2.4M parameters for Gaussian noise removal. In practice, MSResNet using the TRNR strategy with 100 images from Rain100L\cite{Yang2017Jorder} (50.0\% of train-set), 250 images from Rain100H\cite{Yang2017Jorder} (20.0\% of train-set), 280 images from Rain800\cite{Rain800} (40\% of train-set), 600 images from RainLight~\cite{JORDERE}, RainHeavy~\cite{JORDERE} (33.3\% of train-set), 1400 images from Rain1400~\cite{Fu2017DDN} (11.1\% of train-set), and 500 images from BSD500\cite{BSD} and Waterloo \cite{ma2017waterloo} (9.7\% of train-set) respectively have achieved impressive performances when compared to recent image rain removal \cite{Fu2017DDN,Yang2017Jorder,Xia2018Rescan,Zhang2018DID,DualCNN,Yang2019acm,JORDERE,MPRNet,DCSFN,ADN,RLNet,JDDGD} and Gaussian noise removal~\cite{dabovBM3D,KaiZhangDnCNN2017,Zhang2018FFDNet, BUIFD, PaCNet,NLNet2017Left} models.

Fig. \ref{fig: visual-comparison} presents two image rain removal examples from Rain800 \cite{Rain800} dataset. ReHEN\cite{Yang2019acm} learns from the entire train-set (700 images), whereas MSResNet learns from only 280 images. Fig. \ref{fig: visual-comparison} indicates that MSResNet trained using fewer images can provide better results. To summarize, the contributions in this paper are summarized below:

\begin{figure*}[htbp]
	\centering
	\setlength{\abovecaptionskip}{0.01in}
	\begin{minipage}[t]{0.32\linewidth}
		\centering
		\includegraphics[width=4.6cm]{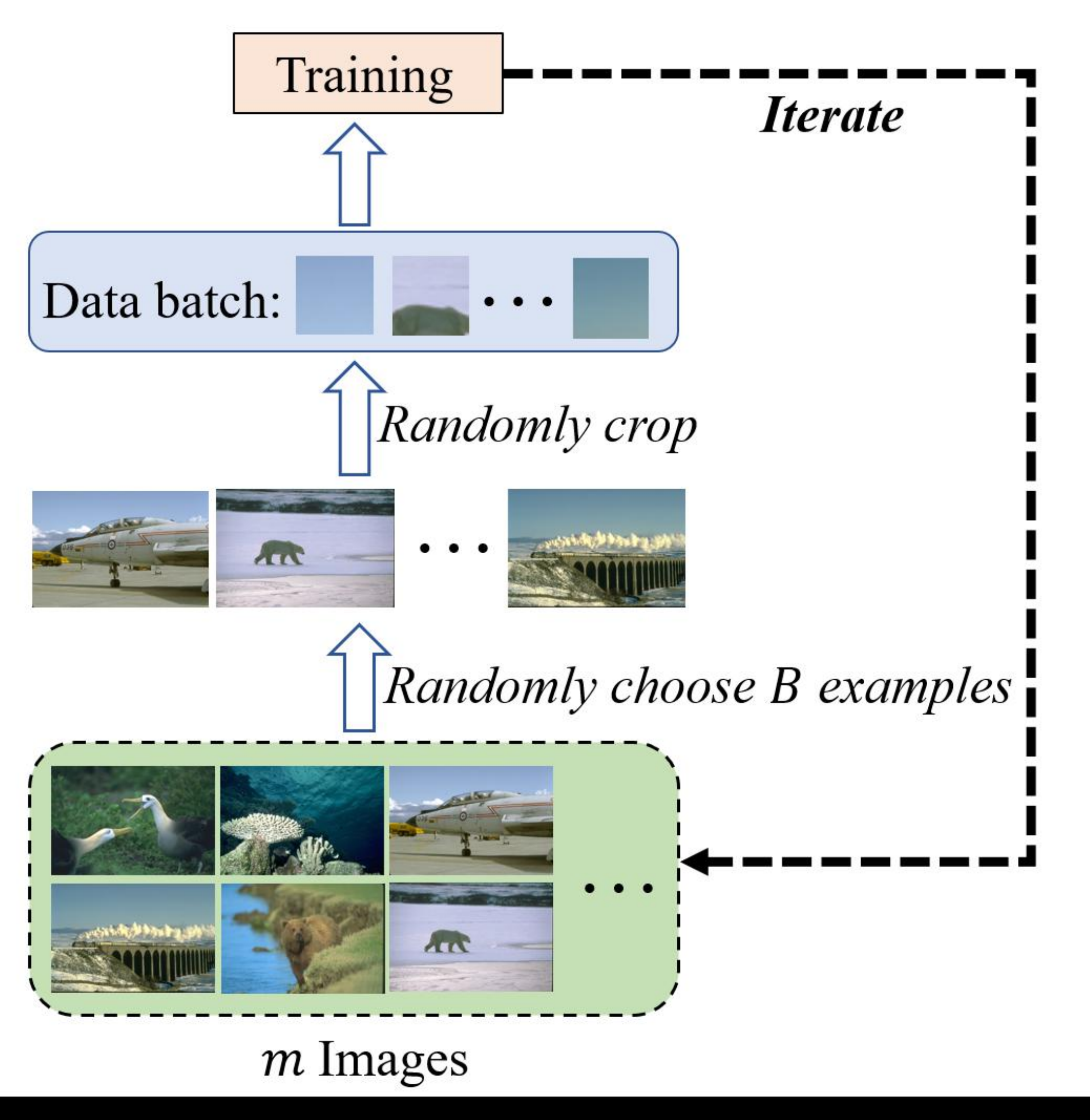}\\
		\footnotesize{(a). Random Image Sampling (RIS)}
	\end{minipage}
	\begin{minipage}[t]{0.32\linewidth}
		\centering
		\includegraphics[width=4.2cm]{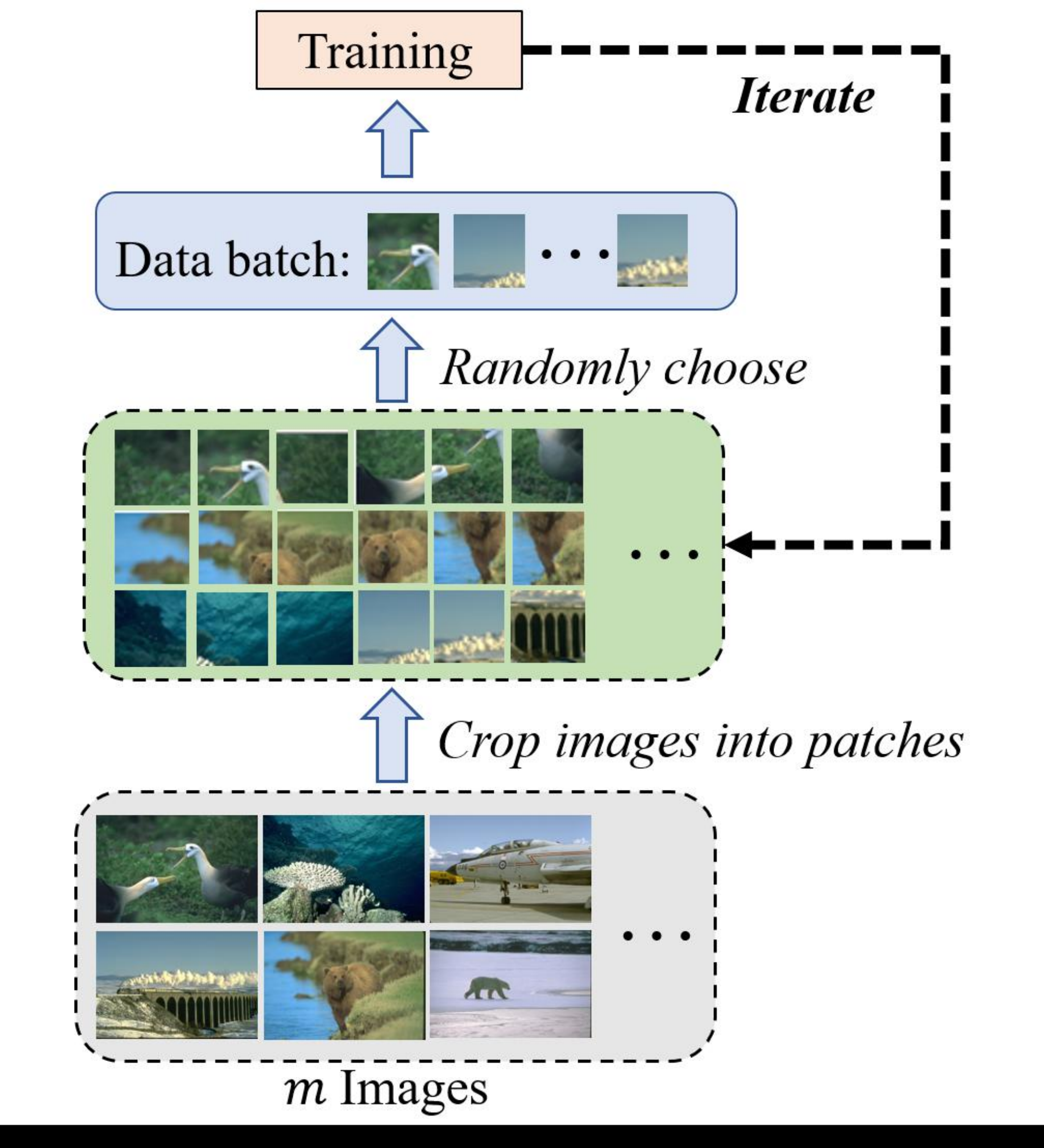}\\
		\footnotesize{(b). Random Patch Sampling (RPS)}
	\end{minipage} 
	\begin{minipage}[t]{0.32\linewidth}
		\centering
		\includegraphics[width=4.2cm]{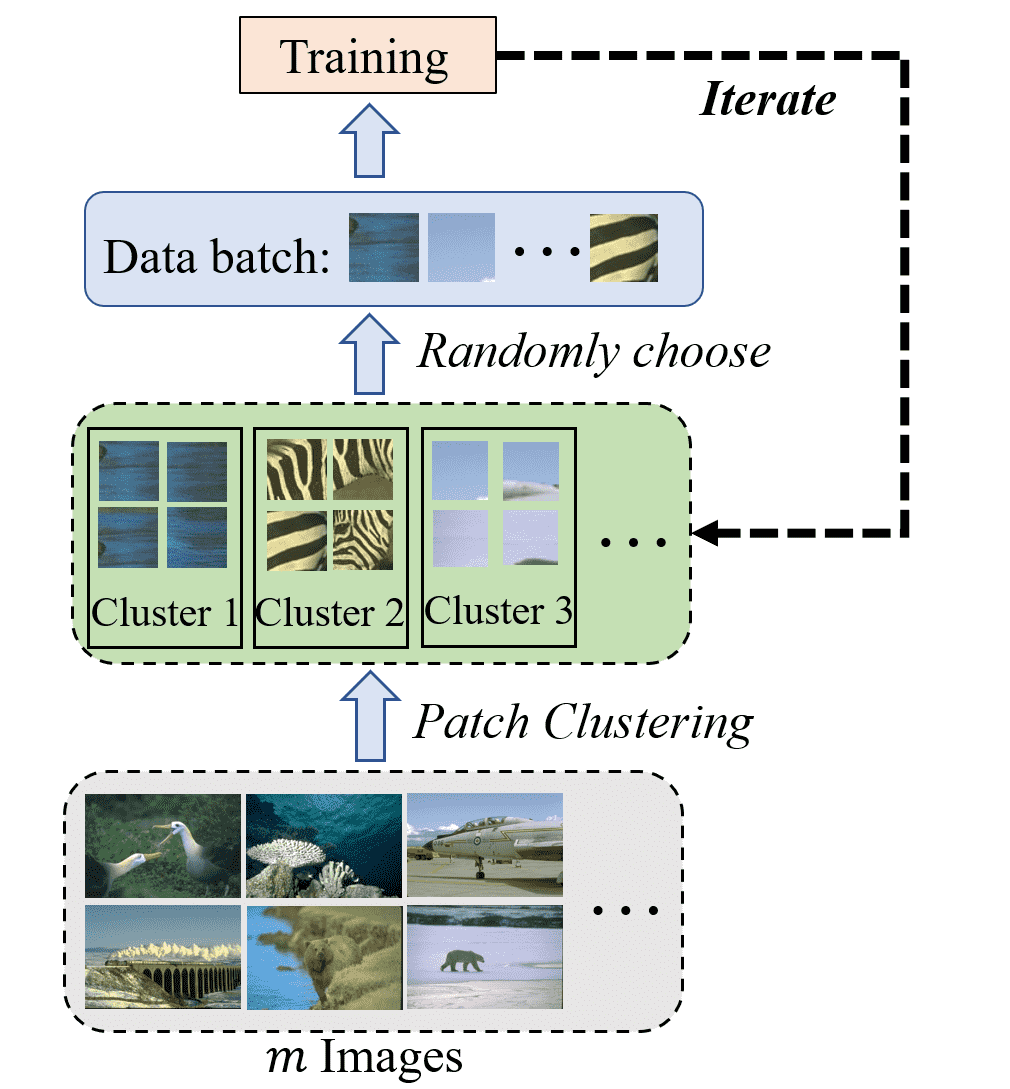}\\
		\footnotesize{(c). Random Cluster Sampling (RCS)}
	\end{minipage}
	\vspace{3.0pt}
	\caption{Comparison of random sampling strategies. The green shaded box indicates the data used for sampling. (a) random image sampling, which is widely used by recent image rain removal models. (b) random patch sampling, which is utilized in DnCNN~\cite{KaiZhangDnCNN2017}. (c) random cluster sampling.}
	\label{fig:sampling}
\end{figure*}

\begin{itemize}
	\item We propose a patch analysis strategy to cluster image patches according to spatial and statistical properties. The patch analysis strategy utilizes RCS to sample image patches discriminatively for training. RCS analytically increases the utilization of image patches.
	\item We propose the task-driven approach dubbed TRNR for learning image rain and noise removal with a few images. TRNR trains neural networks with abundant N-frequency-K-shot learning tasks, and it can boost performances when data is limited. Practically, TRNR boosts the performance of recent image rain removal models DDN~\cite{Fu2017DDN} and ADN~\cite{ADN}.
	\item To prove the effectiveness and superiority of TRNR, we design a MSResNet for both image rain removal and Gaussian noise removal. Experimental results on synthetic datasets have shown that MSResNet performs better than most recent models while requiring fewer images for training. Experimental results on real-world rainy images demonstrate that TRNR can guarantee the generalization ability of MSResNet when data is scarce.
\end{itemize}

The organization of the remaining chapters is as follows. In \S\ref{sec: related-work}, we review recent image rain and noise removal works together with few-shot learning. And in \S\ref{sec: method}, we first elaborate on the patch analysis strategy theoretically, then we detail the proposed task-driven approach TRNR and the architecture of MSResNet. Next, we carry out ablation studies and experiments on image rain and Gaussian noise removal in \S\ref{sec: experiments} for the verification of the proposed TRNR. The conclusion is demonstrated in \S\ref{sec: conclusion}.

\section{Background and Related Work}
\label{sec: related-work}
\subsection{Image Rain and Noise Removal}
\label{subsec:low-level}
Given degraded image $\mathrm{y}$, image rain and noise removal focus on restoring clean background $\mathrm{x}$ via solving the ill-posed linear decomposition problem formulated as Eq. (\ref{eq:lineardecompo}). For image rain removal, $\mathrm{n}$ represents the rain layer\cite{Fu2017DDN,Yang2019acm,DCSFN} and $\mathrm{n}$ symbolizes random noise for image noise removal~\cite{dabovBM3D,KaiZhangDnCNN2017,LiuMWCNN,AdaFM} respectively.
\begin{equation}
	\mathrm{y}=\mathrm{x}+\mathrm{n}
	\label{eq:lineardecompo}
\end{equation}

As both described by the linear decomposition equation, image rain and noise removal are studied similarly. The fundamental assumption for solving Eq. (\ref{eq:lineardecompo}) is that the information required for processing a degraded region comes from its neighbors or non-local regions with similar textures. In the prior-based perspective, the local information can be extracted using dictionary-learning \cite{Fu2011Decomp, Gu2017JCAS, Zhu2017Jointbi}, probabilistic modeling \cite{Li2016GMM, Wang2018Quasi}, etc. And \cite{dabovBM3D,Dong2013NCSR} focus on capturing non-local information to process degraded images. 
These prior-based methods solve the linear decomposition problem iteratively with high time complexity.

With the rise of Deep Learning (DL), image rain and noise removal are coped with deep Convolutional Neural Networks (CNNs). CNNs can exploit rich information from image patches via skillfully stacking convolutional layers. The successes of CNNs on image rain and noise removal mainly stem from (1) fine-grained models \cite{TaiMemnet2017, LiuMWCNN, CycleISP, Yang2017Jorder, Xia2018Rescan, Yang2019acm, LicGAN2018, RenGF2018}, (2) large labeled datasets \cite{KaiZhangDnCNN2017, Fu2017DDN, Yang2017Jorder, Zhang2018DID, Xia2018Rescan, Yang2019acm}, and (3) incorporation of hand-craft priors into CNNs~\cite{DIP2018,LiuMWCNN,Syn2Real,Noise2Self,Noise2Void}. For instance, Wang \textit{et al.} \cite{DCSFN} propose a cross-scale multi-scale module to fuse features at different scales. Zhang \textit{et al.} \cite{Zhang2018DID} build a large labeled dataset which provides 12,000 images for image rain removal. Liu \textit{et al.}~\cite{LiuMWCNN} embed traditional wavelet transformation into CNN for learning image noise removal with a multi-scale wavelet CNN.

Generally, fine-grained models require large labeled datasets for training. The ingredients for building fine-grained models are residual connections \cite{TaiMemnet2017, Xia2018Rescan}, recurrent units \cite{Yang2017Jorder, Xia2018Rescan, Yang2019acm}, attention mechanism \cite{Wang2019SPAnet}, graph convolution~\cite{IGNN,DGCN,CPNet,DAGL}, etc. \cite{Yang2017Jorder, Zhang2018DID} have further introduced rain density labels. Albeit successful and powerful, designing CNN and building large labeled datasets bring huge human labor costs. In addition, the previous RIS strategy leads to insufficient utilization of large labeled datasets and ignores the differences between image patches. Therefore, we propose a patch analysis strategy in this paper to enable CNNs to learn discriminatively from image patches. Then we introduce a task-driven approach named TRNR for image rain and noise removal. Consequently, CNNs can learn image rain and noise removal from plentiful N-frequency-K-shot learning tasks rather than from large labeled datasets.

\begin{figure*}[htbp]
	\centering
	\setlength{\abovecaptionskip}{0.01in}
	\includegraphics[width=0.95\textwidth]{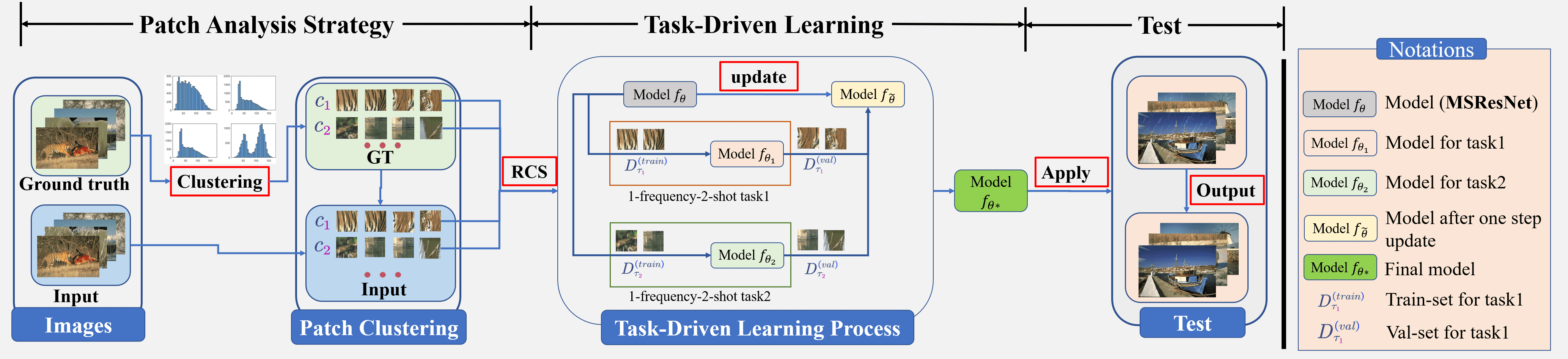}
	\caption{Overview of the proposed TRNR (image rain removal as an example). We first use the patch analysis strategy to crop images into patches and cluster all image patches. Then RCS is used to sample N-frequency-K-shot tasks for training (two 1-frequency-2-shot tasks as an example). At test time, we use the final model for testing.}
	\label{fig:TPAFI}
\end{figure*}

\subsection{Meta-Learning}
\label{subsec:meta-learning}
Meta-Learning, interpreted as \textit{learning to learn}, aims to exploit meta-knowledge via learning on a set of tasks \cite{MetaSurvey}. The learned meta-knowledge can help a neural network adapt fast to unknown tasks. Meta-Learning algorithms learn meta-knowledge via accumulating experience from multiple learning episodes. Specifically, the meta-learning algorithm first learns professional knowledge on a set of tasks with specific objectives and datasets (base-learning). Then, it integrates the learned professional knowledge to update meta-knowledge (meta-learning). Finn \textit{et al.} \cite{Finn2017MAML} propose MAML to learn a good initialization of neural network parameters. MAML has been shown to work well for human pose estimation \cite{MAMLHumanPose} and visual navigation \cite{MAMLNavigation}. In general, the proposed 
TRNR and MAML both learn from numerous tasks. However, MAML aims to explore meta-knowledge for fast adaptation, but the proposed TRNR is devoted to exploiting rich information from a few images and improving the generalization ability.

\subsection{Few-Shot Learning}
\label{subsec:few-shot}
Few-Shot Learning (FSL) focuses on learning a model with a few examples. With the help of prior knowledge, FSL can rapidly generalize to new tasks that contain only a few samples with supervision information \cite{wang2020FSL}.
FSL has made splendid progress on character generation, image classification, visual navigation, etc. The classical FSL problem is the N-way-K-shot classification, where the training data contains $NK$ examples from $N$ classes and each class with $K$ examples \cite{wang2020FSL}. MAML has shown great potential in the N-way-K-shot classification. However, to the best of our knowledge, FSL for pixel-level image processing has rarely been explored, which is helpful when data is scarce. To deal with FSL for image rain and noise removal, we propose the task-driven approach TRNR.

\section{Task-Driven Image Rain and Noise Removal}
\label{sec: method}
In this section, we first detail the proposed patch analysis strategy in \S\ref{subsec: pa}. In \S\ref{subsec: task-driven}, we introduce the N-frequency-K-shot learning task based on the patch analysis strategy. Then we demonstrate the proposed task-driven approach for image rain and noise removal. In \S\ref{subsec: MSResNet}, we present the architecture of the MSResNet.

\subsection{Patch Analysis Strategy}
\label{subsec: pa}
Most recent DL based image rain removal models \cite{Fu2017DDN,Syn2Real,Yang2017Jorder,Yang2019acm,Xia2018Rescan}  and noise removal models \cite{DIP2018,KaiZhangDnCNN2017,LiuMWCNN,NLNet2017Left,Zhang2018FFDNet} are trained with image patches utilizing RIS. Suppose a dataset $D=\{(x_i, y_i)\}_{i=1}^{m}$ contains $m$ examples, where $(x_i, y_i)$ indicates the $i$-th clean and degraded image pair following Eq. (\ref{eq:lineardecompo}) respectively. Typically, we need to sample a data batch composed of $B$ clean and degraded image patch pairs at each training step, where $B$ is the batch size. 

Take sampling $B$ clean image patches as an example, the most widely used RIS as shown in Fig. \ref{fig:sampling}(a), provides $B$ image patches via cropping an image patch from each randomly sampled $B$ images. Analytically, RIS leads to low utilization of image patches. Generally, let $p^{unuse}_{k}$ denote the probability that an image patch has not been sampled through $k$ iterations, $T_e$ denotes the total iterations of an epoch, and $P$ represents the number of patches that can be cropped out of an image. We can 
calculate $p^{unuse}_k$ analytically for RIS in Eq. (\ref{eq:random image sampling}).
\begin{align}
	\label{eq:random image sampling}
	\text{(RIS): }\ &p^{unuse}_k = (1-\frac{B}{mP})^k, \notag\\
	&\text{s.t. }\ k\leq T_e=\frac{m}{B},\  B\leq m
\end{align}
\indent Note that $p^{unuse}_k$ have minimum near to $\mathrm{e}^{-1/P}$for RIS when $k=T_e$ and $P$ is always a large number for RIS. For example, an image of size $321\times481$ can produce $P=\lfloor \frac{321-64}{16}+1\rfloor\lfloor\frac{481-64}{16}+1\rfloor=459$ image patches of size $64\times 64$ with stride as $16$. One way to increase utilization of image patches is \textit{random patch sampling} (RPS) as shown in Fig. \ref{fig:sampling}(b), which is often used in image noise removal~\cite{KaiZhangDnCNN2017,Zhang2018FFDNet,LiuMWCNN}. RPS first crops $m$ images into $mP$ image patches for sampling, which is still feasible when $m\leq B$ while RIS fails. $p^{unuse}_k$ for RPS is formulated in Eq. (\ref{eq:random patch sampling}). Both RIS and RPS regard image patches equal, which may result in data redundancy. For example, as shown in Fig. \ref{fig:sampling}, the sky scene in data batches sampled from RIS and RPS is redundant, since sky scene exists almost in every image. In practice, the sky scene patches contain limited information and thus will contribute to a plain update of parameters.
\begin{equation}
	\label{eq:random patch sampling}
	\text{(RPS): }\ p^{unuse}_k=1-k\frac{B}{mP}, \text{ s.t. }k\leq T_e=\frac{mP}{B}
\end{equation}
\indent Further, we propose the patch analysis strategy to improve the data utilization as shown in Fig. \ref{fig:TPAFI}. For convenience, we only consider clean images dataset $D^{(c)}=\{x_i\}_{i=1}^m$ since clean image and degraded image are pairwise. First, we crop all $m$ images in $D^{(c)}$ to build clean image patches dataset $D^{(p, c)}=\{p_i\}_{i=1}^{N^{(p)}}$, where $N^{(p)}=mP$ means the total number of patches and $p_i$ denotes the $i$-th patch. Second, we cluster all $N^{(p)}$ image patches into $C$ clusters such that $D^{(p,c)}=\cup_{j=1}^C D^{(p,c)}_j$, and $D^{(p,c)}_j=\{p^{(j)}_i\}_{i=1}^{N^{(p)}_j}$, where $p^{(j)}_i$ denotes the $i$-th patch in $j$-th cluster and $N^{(p)}_j$ means the number of patches in cluster $j$.

To obtain a data batch composed of $B$ image patches, we first choose $B$ clusters among total $C$ clusters, then we randomly sample one image patch with replacement from each cluster to form a data batch. We call this sampling \textit{random cluster sampling} (RCS) as shown in Fig. \ref{fig:sampling}(c). Mathematically, if an image patch belongs to cluster $j$, then $p^{unuse}_k$ can be analytically derived as Eq. (\ref{eq:patch analysis}). Note that when $N^{(p)}_j\leq N^{(p)}/C$, RCS increases the utilization compared to RIS. Furthermore, RCS does not have a data redundancy problem and treats image patches discriminatively.
\begin{align}
	\label{eq:patch analysis}
	\text{(RCS): }&p^{unuse}_k=\left(1-\frac{B}{CN^{(p)}_j}\right)^k \notag\\
	&\text{s.t. }\ k\leq T_e=\frac{N^{(p)}}{B}
\end{align}

As for the clustering, we employ the euclidean distance and the Kullback-Leibler (KL) divergence \cite{KLdivergence} of the color histogram as the similarity measures for image patches. We summarize the clustering method in Algorithm 1.

\begin{figure}[!t]
	\centering
	\setlength{\abovecaptionskip}{0.01in}
	\includegraphics[width=7cm]{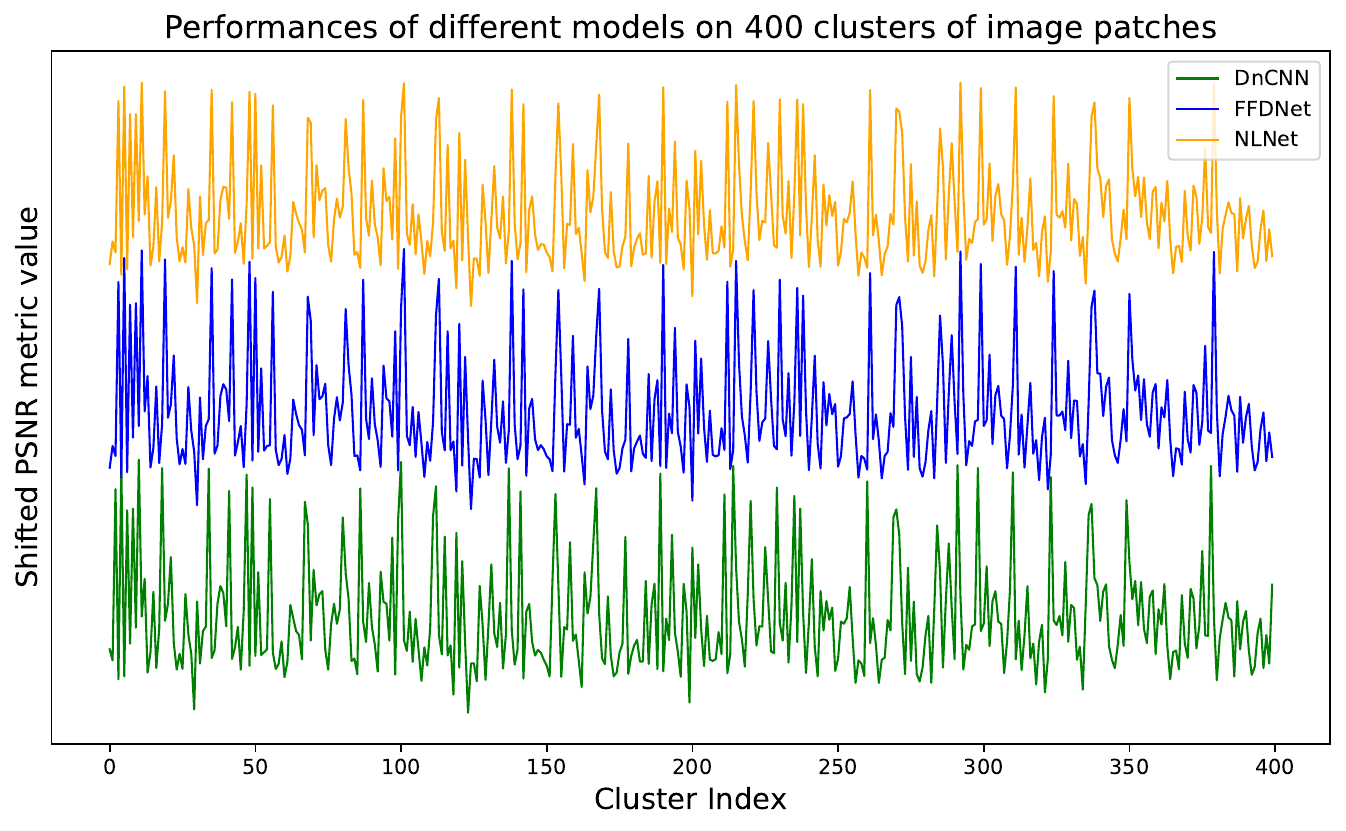}
	\caption{Performances of DnCNN \cite{KaiZhangDnCNN2017}, FFDNet \cite{Zhang2018FFDNet}, NLNet \cite{NLNet2017Left} on 400 clusters of image patches. Different methods tend to simultaneously behave well or bad on the specific cluster of image patches. We shift the curves for the comparison of tendency.}
	\label{fig:tendency}
\end{figure}

\begin{figure*}
	\centering
	\setlength{\abovecaptionskip}{0.01in}
	\includegraphics[width=0.8\linewidth]{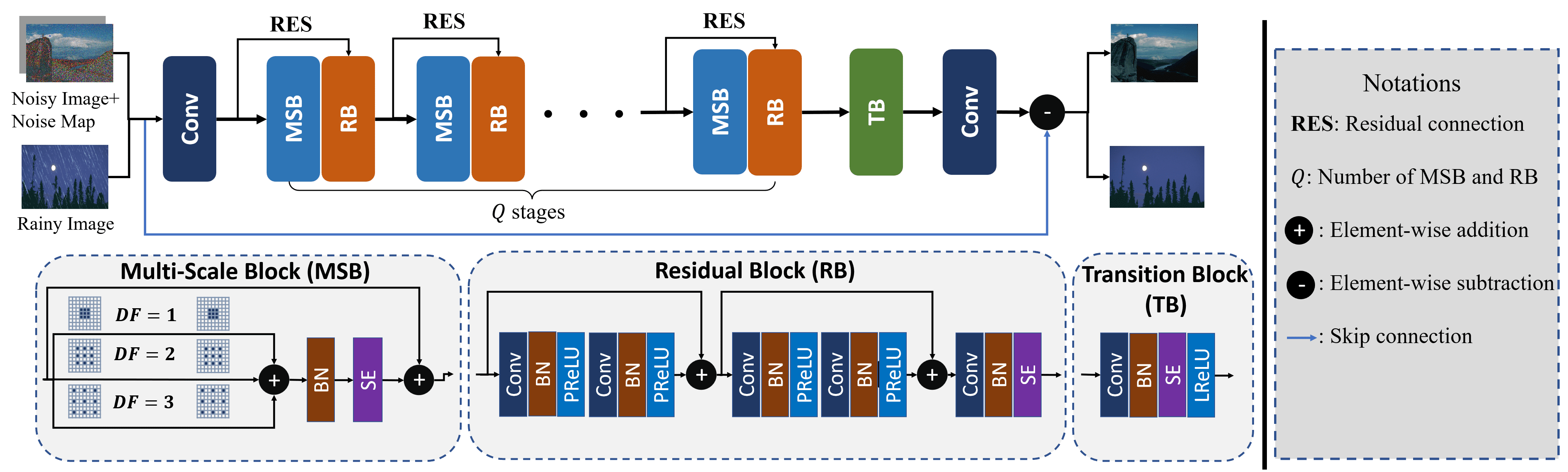}
	\caption{The whole structure of Multi-Scale Residual Neural Network (MSResNet) for image de-raining and denoise. "RES" means adding a residual connection.}
	\label{fig:net}
\end{figure*}

\begin{algorithm}
	\label{algo: pa}
	\caption{Patch Clustering Algorithm}
	\begin{algorithmic}[1]
		\REQUIRE Clean dataset $D^{(c)}=\{x_i\}_{i=1}^{m}$, number of clusters $C$, threshold $T$, momentum $\rho=0.05$.
		
		\ENSURE Patch clusters $D^{(p,c)}=\cup_{j=1}^{C}D_j^{(p,c)}$
		\STATE Initialize $D_j^{(p,c)}=\{\}$ for $j=1$ to $C$. Set cluster centers $r_j=0$ for $j=1$ to $C$.
		
		\STATE Generate clean patches $D^{(p,c)}=\{p_i\}_{i=1}^{N^{(p)}}$ from input dataset $D^{(c)}$.
		\STATE Randomly choose one patch $p_{1}$ from $D^{(p,c)}$. Set $D_1^{(p,c)}=\{p_1\}$, and cluster center $r_1$=$p_1$.
		\STATE Initialize current cluster index $Idx=1$.
		\FOR {$k=2$ to $N^{(p)}$}
		\STATE Compute similarities $score$ between $p_k$ and all cluster centers $r_j$ for $j=1$ to $Idx$. 
		\STATE Find the largest $score$ and corresponded $j$.
		\IF {$score > T $ or $Idx >= C$} 
		\STATE /* add patch to cluster $Idx$ \qquad\qquad\qquad\quad\ */
		\STATE $D_j^{(p,c)} = D_j^{(p,c)} \bigcup \{p_k\}$
		\IF {$Idx<C$}
		\STATE/* update cluster center \qquad\qquad\qquad\qquad*/
		\STATE $r_{Idx}=(1-\rho)\times r_{Idx}+\rho\times p_k$
		\ENDIF
		\ELSE
		\STATE /* new cluster added \qquad\qquad\qquad\qquad\qquad */
		\STATE $Idx = Idx+1$
		\STATE $D_{Idx}^{(p,c)}=\{p_k\}$, and set $r_{Idx}=p_k$
		\ENDIF
		\ENDFOR
	\end{algorithmic}
\end{algorithm}

\subsection{Task-Driven Approach Based on Patch Analysis}
\label{subsec: task-driven}
The goal of the patch analysis strategy is not only to improve the utilization of image patches but also to generate N-frequency-K-shot learning task distribution $p(\mathcal{T})$. Based on the patch analysis strategy, we empirically observe that deep neural networks, such as DnCNN~\cite{KaiZhangDnCNN2017}, FFDNet~\cite{Zhang2018FFDNet}, and NLNet~\cite{NLNet2017Left} tend to simultaneously behave bad or well on image patches from the same cluster as shown in Fig. \ref{fig:tendency}. This indicates that we can regard image rain and noise removal on image patches sharing similar spatial and statistical properties as a sub-task.

\noindent\textbf{N-frequency-K-shot Learning Task.} Based on the above observation, we introduce the N-frequency-K-shot learning task $\tau\sim p(\mathcal{T})$, which aims to quickly learn on $N$ kinds of image patches with parameterized model $f_{\theta}$. Hence, we can embed a batch of N-frequency-K-shot learning tasks as a ``data batch" in the data-driven learning process when data is scarce. This approach results in a bilevel optimization. The inner-level optimization tries to find the optimal parameters $\theta_j^{*}$ for each N-frequency-K-shot task $\tau_j$ by training on the corresponding train-set $\mathcal{D}^{(train)}_{\tau_j}$. The train-set $\mathcal{D}^{(train)}_{\tau_j}$ contains $NK$ examples sampled from $N$ clusters and each cluster with $K$ image patches. Fig. \ref{fig:TPAFI} presents an example of 1-frequency-2-shot learning, where task1 learns on 2 image patches with \textit{tiger stripes} by gradient descent. Formally, the inner-level optimization for task $\tau_j$ can be formed as Eq. (\ref{eq:N-frequency-K-shot learning}).
\begin{equation}
	\label{eq:N-frequency-K-shot learning}
	\theta_j=\theta-\alpha\nabla_{\theta}\mathcal{L}(f_{\theta}, \mathcal{D}_{\tau_j}^{(train)})=\theta-\alpha\mathbf{g}^{(train)}_{j}
\end{equation}
where $\mathcal{L}$ denotes the loss function for $\tau_j$, $\alpha$ is a step size hyperparameter, and $\mathbf{g}_j^{(train)}$ represents the gradients of $\theta$ computed on train-set. Further, we denote the validation set for $\tau_j$ as $\mathcal{D}^{(val)}_{\tau_j}$ which also contains $NK$ examples. 

\noindent\textbf{Task-Driven Approach.} The data-driven methods update parameters by taking a small step along the opposite direction of gradients computed on a batch of data. Analogically, the proposed task-driven approach TRNR updates the parameters $\theta$ by taking a small step along the opposite direction of gradients computed on a batch of tasks, which serves as the outer-level optimization. In practice, TRNR first samples a batch of $R$ N-frequency-K-shot learning tasks $\{\tau_j\}_{j=1}^R$. Then TRNR performs the inner-level optimization for each task $\tau_j$ using Eq. (\ref{eq:N-frequency-K-shot learning}) to obtain the adapted parameters $\theta_j$. 

Although model parameterized with $\theta_j$ can now achieve lower loss value on train-set $\mathcal{D}_{\tau_j}^{(train)}$, its generalization ability on validation set $\mathcal{D}_{\tau_j}^{(val)}$ is not guaranteed. Therefore, we compute the gradients of $\theta$ at $\theta_j$ on the validation set $\mathcal{D}_{\tau_j}^{(val)}$, which can decrease the loss value on both train-set and validation set of $\tau_j$. By summarizing the gradients information over all $R$ tasks, we can make a signifcant update for $\theta$ as formulated in Eq. (\ref{eq:outer-loop}) mathematically. Fig. {\ref{fig:TPAFI}} presents an example of task-driven learning, where two 1-frequency-2-shot tasks are sampled to obtain the modified parameters $\tilde{\theta}$. We summarize the task-driven approach in Algorithm 2. 

\begin{equation}
	\label{eq:outer-loop}
	\tilde{\theta} = \theta-\beta\frac1{R}\sum_{j=1}^{R}\nabla_{\theta}\mathcal{L}(f_{\theta_j}, \mathcal{D}^{(val)}_{\tau_j})
\end{equation}

Analytically, we demonstrate that Eq. (\ref{eq:outer-loop}) corresponds to optimizing the TRNR loss $\mathcal{L}^{TRNR}$ in Eq. (\ref{eq: TRNR-loss}), 

\begin{align}
	\mathcal{L}^{TRNR}&=\frac1{R}\sum_j\mathcal{L}(\theta, \mathcal{D}^{(val)}_{\tau_j}) \notag \\
	&-\alpha\frac1{R}\sum_j (\mathbf{g}^{(train)}_j)^T\phi(\mathbf{g}^{(val)}_j)
	\label{eq: TRNR-loss}
\end{align}
where $\mathbf{g}_j^{(val)}$ is the gradient of $\theta$ computed on validation set and $\phi(\cdot)$ means to remove the derivative of $\theta$. From Eq. (\ref{eq: TRNR-loss}), we conclude that TRNR can explicitly optimize on the validation sets, and can also simultaneously decrease the loss values on both train-sets and validation sets since $\mathbf{g}_{j}^{(train)}$ and $\mathbf{g}_j^{(val)}$ need to be aligned. Hence, TRNR serves as a superior learning approach, which can improve the generalization ability of models. The proof of Eq. (\ref{eq: TRNR-loss}) is presented in the Appendix.

\begin{algorithm}
	\caption{Proposed TRNR Learning Algorithm}
	\begin{algorithmic}[1]
		\REQUIRE Deep neural network $f_{\theta}$, step size hyperparameters $\alpha,\beta$, task distribution $p(\mathcal{T})$, Clustered image patches dataset $D^{(p,c)}=\cup_{j=1}^C D_j^{(p,c)}$
		\ENSURE well-trained model $f_{\theta*}$
		\WHILE {not converge}
		\STATE Sample $R$ N-frequency-K-shot learning tasks $\{\tau_j\}$ from $p(\mathcal{T})$ 
		\FOR{$j=1$ to $R$}
		\STATE /*\textit{\ \quad Learning on $R$ tasks}\qquad\qquad\qquad\qquad\quad\ \   */
		\STATE Sample $\mathcal{D}^{(train)}_{\tau_j}$ from $D^{(p,c)}$ for task $\tau_j$ using RCS strategy
		\STATE Adapt $\theta$ to task $\tau_j$ as an initialization
		\STATE Compute adapted parameters $\theta_j$ for $\tau_j$ in Eq. (\ref{eq:N-frequency-K-shot learning}) 
		\STATE 	Sample $\mathcal{D}^{(val)}_{\tau_j}$ from $D^{(p,c)}$ for task $\tau_j$
		\ENDFOR
		\STATE /*\textit{\ \quad Summarize gradients from $R$ tasks}\qquad\qquad\quad */
		\STATE Compute updated model parameters $\tilde{\theta}$ in Eq. (\ref{eq:outer-loop}) using all $\theta_j$ and $\mathcal{D}^{(val)}_{\tau_j}$
		\STATE Initialize the model with parameters $\theta=\tilde{\theta}$
		\ENDWHILE
		\STATE Let $\theta^{*}=\tilde{\theta}$
		\RETURN well-trained model $f_{\theta^{*}}$
	\end{algorithmic}
\end{algorithm}

\subsection{Multi-Scale Residual Neural Network (MSResNet)}
\label{subsec: MSResNet}
To better balance network complexity against the computational overhead, we introduce three basic blocks in the proposed Multi-Scale Residual Neural Network (MSResNet), which include Multi-Scale Block (MSB), Residual Block (RB), and Transition Block (TB). Multi-scale design and residual connection have been proved to be effective in image rain removal \cite{Yang2017Jorder,Xia2018Rescan,Yang2019acm,JDNet} and noise removal \cite{KaiZhangDnCNN2017,LiuMWCNN,TaiMemnet2017,CycleISP}. The details of MSResNet are illustrated in Fig. \ref{fig:net}.

\noindent\textbf{Multi-Scale Block (MSB)} MSB is used for enlarging the receptive field and aggregating feature maps of different scales. As shown in Fig. \ref{fig:net}, at each stage, MSB takes the output from the previous RB as input except for the first MSB. Inspired by the multi-scale design in \cite{yang2019ReMAEN}, we use parallel atrous convolution \cite{AtrousConv} layers with different dilation rates to extract multi-scale information in MSB, and apply squeeze-and-excitation (SE) block \cite{SENet} to exploit cross-channel information.

\noindent\textbf{Residual Block (RB)} Residual connection \cite{ResNet} has become indispensable in modern deep CNNs. In this paper, we construct a simple residual block (RB). As shown in Fig. \ref{fig:net}, each RB takes the output of the previous MSB as input. There are two residual connections in each RB to aggregate features from different Conv-BN-PReLU modules. At the end of each RB, a SE block is used to further explore the cross-channel information. 

To obtain finer features for restoring clean background, we apply a skip connection from each MSB to its adjacent MSB as illustrated in Fig. \ref{fig:net} with ``RES".

\noindent\textbf{Transition Block (TB)} After processing input features via $Q$ MSBs and RBs ($Q$ stages), we utilize a transition block named TB to integrate features for inferring the rain layer and noise layer. The configuration of TB is illustrated in Fig. \ref{fig:net}.

\noindent\textbf{Loss function} We use the combination of L1 loss and SSIM loss as the loss function. Mathematically, let $(p, \tilde{p})$ denote the clean patch and corresponding degraded patch pair in $\mathcal{D}^{(train)}_{\tau_j}$. And $\hat{p}=f_{\theta_j}(\tilde{p})$ indicates the restored patch. Then the loss function $\mathcal{L}$ for task $\tau_j$ is formulated below in Eq. (\ref{eq:loss}),
\begin{align}
	\mathcal{L}(f_{\theta}, \mathcal{D}^{(train)}_{\tau_j})&=\mathbb{E}_{(p,\tilde{p})}\left[||p-\hat{p}||_1\right] \notag \\
	&+\lambda\mathbb{E}_{(p,\tilde{p})}\left[1-\text{SSIM}(p, \hat{p})\right]\notag \\
	&\text{s.t. }\tau_j\sim p(\mathcal{T})
	\label{eq:loss}
\end{align}

\noindent where $\lambda$ is a hyperparameter.

\begin{table*}[t]
	\centering
	\small
	\renewcommand{\arraystretch}{1.2}
	\renewcommand\tabcolsep{1.0pt}
	\caption{Details of Datasets for Data-Driven Rain and Noise Removal. The Test-set for WaterlooBSD is 222 Images from Set12, McMaster, Kodak, BSD68, and Urban100.}
	\begin{tabular}{l | c c c c c c c} 
		\bottomrule
		Datasets & Rain100L~\cite{Yang2017Jorder} & Rain100H~\cite{Yang2017Jorder} & Rain800~\cite{Rain800} & RainLight~\cite{JORDERE} & RainHeavy~\cite{JORDERE} & Rain1400~\cite{Fu2017DDN} & WaterlooBSD \\
		\hline
		Train-set & 200 & 1254 & 700 & 1800 & 1800 & 12600 & 5176 \\
		\hline
		Test-set & 100 & 100 & 100 & 200 & 200 & 1400 & 222 \\
		\bottomrule
	\end{tabular}
	\label{tab: RIS datasets}
\end{table*} 

\begin{table*}[t]
	\centering
	\small
	\renewcommand{\arraystretch}{1.2}
	\renewcommand\tabcolsep{1.3pt}
	\caption{Detail Settings of Datasets for Task-Driven Rain and Noise Removal. Train-set Denotes the Number of Training Examples and the Corresponding Ratio of the Entire Train-set in Table \ref{tab: RIS datasets}. \#Train-patch and \#Train-cluster Indicate Total Training Patches and Clusters for TRNR. As for Rain100L-S, RainLight-S, and Rain1400-S, We Present the Size of Train-set and \#Train-patch for Both Clean and Rainy Images Due to Multiple Rainy Augmentations.}
	\begin{tabular}{l | c c c c c c c} 
		\bottomrule
		Datasets & Rain100L-S & Rain100H-S & Rain800-S & RainLight-S & RainHeavy-S & Rain1400-S & WaterlooBSD-S \\
		\hline
		Train-set & 100 (50\%) & 250 (20\%) & 280 (40\%) & 100/600 (33.3\%) & 100/600 (33.3\%) & 100/1400 (11.1\%) & 500 (9.7\%) \\ 
		\#Train-patch & 22003 & 57329 & 73276 & 22587/135522 &  22602/135612 & 28201/394814 & 162764 \\
		\#Train-cluster & 4003 & 4829 & 5090 & 3187 & 3202 & 3671 & 5968\\
		\hline
		Test-set & 100 & 100 & 100 & 200 & 200 & 1400 & 222 \\
		\bottomrule
	\end{tabular}
	\label{tab: TRNR datasets}
\end{table*} 
\section{Experiments}
\label{sec: experiments}
In this section, we conduct extensive experiments to demonstrate the effectiveness of the proposed TRNR approach. For data-driven image rain removal, we train MSResNet using RIS separately on 6 synthetic datasets Rain100L~\cite{Yang2017Jorder}, Rain100H\footnote{1800 image pairs in train-set, but 546 of them are repeated in test-set.}~\cite{Yang2017Jorder}, Rain800~\cite{Rain800}, RainLight~\cite{JORDERE}, RainHeavy~\cite{JORDERE}, Rain1400~\cite{Fu2017DDN} and denote it as MSResNet-RIS. And we train MSResNet using RIS on WaterlooBSD which contains 5176 images collected from BSD~\cite{BSD} and Waterloo~\cite{ma2017waterloo} for image Gaussian noise removal. As for TRNR, we construct corresponding small datasets from the above synthetic datasets, dubbed Rain100L-S, Rain100H-S, Rain800-S, RainLight-S, RainHeavy-S, Rain1400-S, and WaterlooBSD-S, respectively.  And the model is named as MSResNet-TRNR. Since each clean image in RainLight, RainHeavy, and Rain1400 has multiple rainy augmentations, we randomly select 100 clean images and all the corresponding rainy augmentations to build RainLight-S, RainHeavy-S, and Rain1400-S. Details of datasets for MSResNet-RIS and MSResNet-TRNR are shown in Table \ref{tab: RIS datasets} and Table \ref{tab: TRNR datasets} respectively. We evaluate MSResNet-RIS and MSResNet-TRNR on 6 synthetic datasets Rain100L, Rain100H, Rain800, RainLight, RainHeavy, Rain1400, and real-world images collected from~\cite{Yang2017Jorder,CGAN,Wang2019SPAnet}. Datasets BSD68~\cite{BSD}, Kodak\footnote{\href{http://r0k.us/graphics/kodak/}{http://r0k.us/graphics/kodak/}}, McMaster~\cite{McMaster}, and Urban100~\cite{Urban} are utilized for color image Gaussian noise removal evaluation. Note that following~\cite{KaiZhangDnCNN2017,Zhang2018FFDNet}, the images in Kodak are center cropped into size $500\times500$. Relatively, Set12, BSD68, and Urban100 are used for grayscale image Gaussian noise removal comparison. Image rain and noise removal are evaluated in terms of Peak Signal-to-Noise Ratio (PSNR)~\cite{PSNR}, and Structural Similarity (SSIM)~\cite{SSIM} metrics. We compute PSNR and SSIM metrics over RGB channels for color images.

\subsection{Training Details}
\noindent\textbf{Basic Training Settings.}
We set $Q=4$ in MSResNet for image rain removal and $Q=6$ for Gaussian noise removal. Moreover, the filters of all convolution layers in MSResNet have the size of $3\times 3$ and 48 channels for image rain removal, and have 64 channels for Gaussian noise removal. All random seeds are fixed to 0. And the negative slopes for all LeakyReLU activations are $0.2$. We first train MSResNet-RIS for image rain removal and Gaussian noise removal. Then we train MSResNet-TRNR utilizing the proposed TRNR with corresponding small datasets. The noise level in the training process of MSResNet-RIS and MSResNet-TRNR is uniformly in the range [0, 55]. We implement MSResNet using PyTorch~\cite{pytorchofficial} and train it with Adam optimizer on NVIDIA GeForce RTX 3090 GPU.

\noindent\textbf{Task-Driven Image Rain Removal Settings.}
We train MSResNet-TRNR using TRNR for image rain removal. In the task-learning process, $N$ and $K$ are $12$ and $1$ respectively in N-frequency-K-shot learning tasks for Rain100L-S, Rain100H-S, and Rain800-S. The batch size $R$ of N-frequency-K-shot learning tasks is set as $5$ in Eq. (\ref{eq:outer-loop}). Since each clean image in RainLight-S, RainHeavy-S and Rain1400-S have at least 6 rainy augmentations, we set a larger $N=32$ with $K=1$ and $R=3$. Step size hyperparameters $\alpha$ and $\beta$ are both set to $0.001$. 

\noindent\textbf{Task-Driven Image Noise Removal Settings.}
We employ TRNR to train MSResNet-TRNR for Gaussian noise removal. In the task-learning process, $N$ and $K$ are $18$ and $1$ for N-frequency-K-shot learning tasks due to larger model capacity. The batch size $R$ of N-frequency-K-shot learning tasks is set as $3$ in Eq. (\ref{eq:outer-loop}). We do not impose SSIM loss following DnCNN~\cite{KaiZhangDnCNN2017} and FFDNet~\cite{Zhang2018FFDNet}. Step size hyperparameters $\alpha$ and $\beta$ are both set to $0.001$. 

\subsection{Ablation Study}
In this subsection, we further study the proposed MSResNet and TRNR strategy. Specifically, we primarily analyze the design of MSResNet to find the best combinations of different modules. Then we compare the widely used strategy RIS with RPS, RCS, and TRNR to show that the model trained using RIS suffers a heavy performance drop against using TRNR. We also provide a feature visualization analysis to figure out what TRNR learns. Next, we determine the best configuration for task-driven learning, which includes the choices for $N$ in N-frequency-K-shot learning tasks and the batch size $R$ in Eq. (\ref{eq:outer-loop}). Then we study the effect of the cluster number on TRNR. At last, we take an investigation on RIS, RPS, RCS, and TRNR when the dataset size is from 60 to 280 to demonstrate TRNR's robustness.

\begin{table}[htbp]
	\centering
	\small
	\renewcommand{\arraystretch}{1.1}
	\renewcommand\tabcolsep{3.5pt}
	\caption{Average PSNR and SSIM Metrics on Test-Set of Rain100L of Different Designs. MS Means Multi-Scale Module and SE Indicates the SE Block.}
	\begin{tabular}{l|c c | c c c| c}
		\bottomrule
		Design & MS & SE & Q=3 & Q=4 & Q=5 & PSNR/SSIM  \\
		\bottomrule
		Baseline & $\times$ & $\times$ & $\surd$ & $\times$ & $\times$ & 34.56/0.965 \\
		$N_1$  & $\surd$ & $\times$ & $\surd$ & $\times$ & $\times$ & 34.72/0.967 \\
		$N_2$ & $\times$ & $\surd$ & $\surd$ & $\times$ & $\times$ & 34.59/0.966\\
		$N_3(Q=3)$ & $\surd$ & $\surd$ & $\surd$ & $\times$ & $\times$ & 34.91/0.967\\
		$N_3(Q=4)$ & $\surd$ & $\surd$ & $\times$ & $\surd$ & $\times$ & \textbf{35.12/0.967} \\
		$N_3(Q=5)$ & $\surd$ & $\surd$ & $\times$ & $\times$ & $\surd$ & 34.95/0.966\\
		\bottomrule
	\end{tabular}
	\label{tab: modules}
\end{table}

\noindent\textbf{Analysis of the MSResNet Design.} We begin by examining the importance of different modules in MSResNet. In particular, we regard MSResNet (Q=3) without multi-scale design in MSB, without all SE blocks as the Baseline. And we construct $3$ models listed as follows for comparison:
\begin{itemize}
	\item $N_1$: Baseline with multi-scale design in each MSB.
	\item $N_2$: Baseline with SE blocks at the bottom of all MSB, RB, and TB.
	\item $N_3$: $N_1$ with SE blocks at the bottom of all MSB, RBs, and TB.
\end{itemize}

\begin{figure*}[!t]
	\centering
	\setlength{\abovecaptionskip}{0.01in}
	\begin{minipage}[t]{0.22\linewidth}
		\centering
		\includegraphics[width=3.9cm]{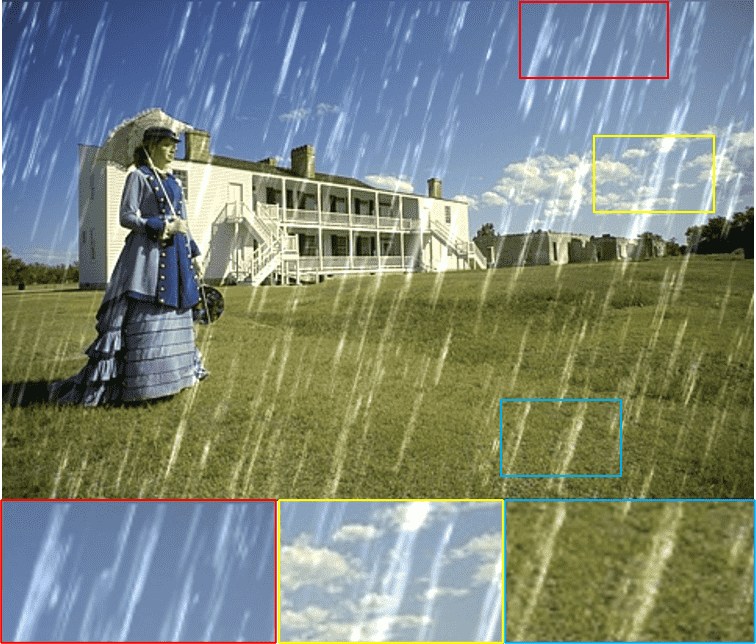}\vspace{-1pt}
		\footnotesize{Rainy Image (22.86/0.767)}\vspace{5pt}
	\end{minipage}
	\begin{minipage}[t]{0.22\linewidth}
		\centering
		\includegraphics[width=3.9cm]{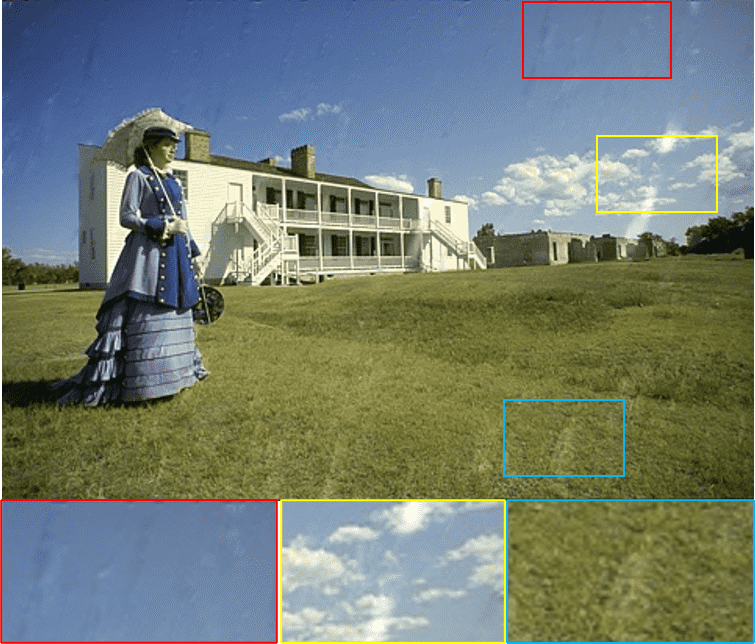}\vspace{-1pt}
		\footnotesize{$Q=3$, $\lambda=0$ (32.10/0.960)}\vspace{5pt}
	\end{minipage} 
	\begin{minipage}[t]{0.22\linewidth}
		\centering
		\includegraphics[width=3.9cm]{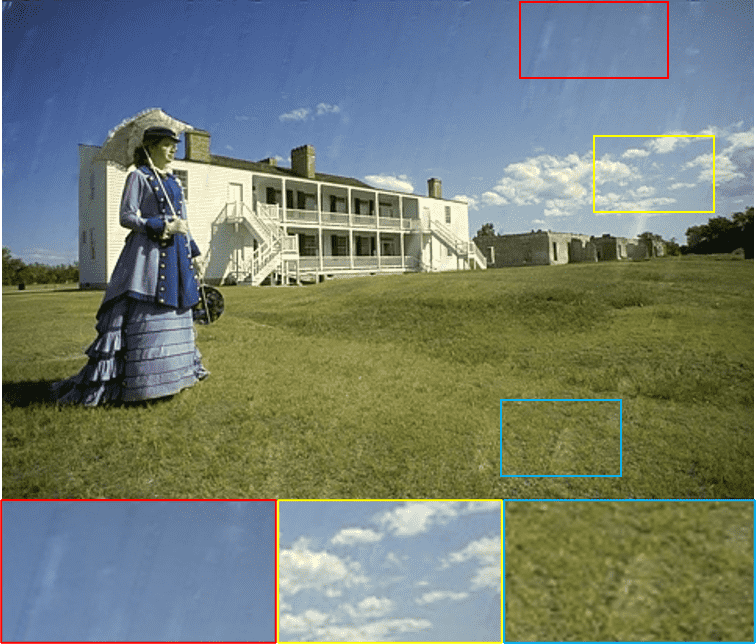}\vspace{-1pt}
		\footnotesize{$Q=4$, $\lambda=0$ (32.93/0.963)} \vspace{5pt}
	\end{minipage}
	\begin{minipage}[t]{0.22\linewidth}
		\centering
		\includegraphics[width=3.9cm]{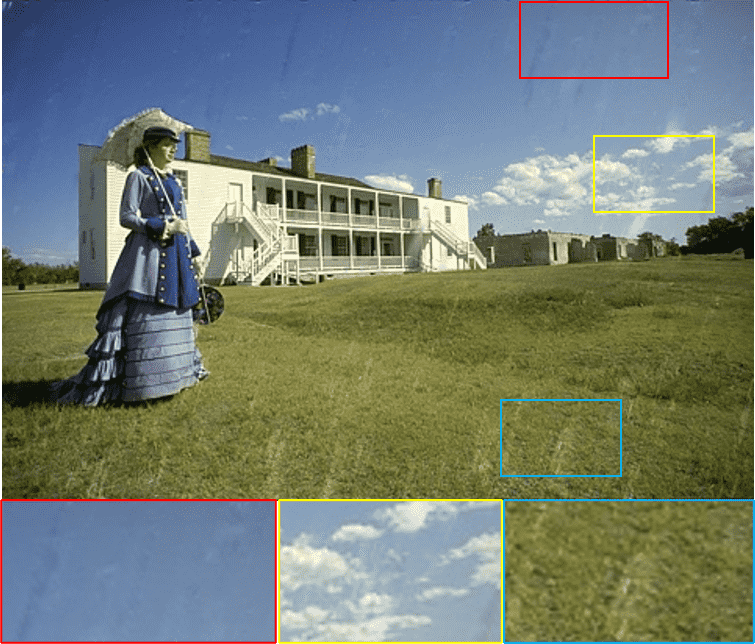}\vspace{-1pt}
		\footnotesize{$Q=5$, $\lambda=0$ (32.30/0.960)}\vspace{5pt}
	\end{minipage}
	\\ 
	\begin{minipage}[t]{0.22\linewidth}
		\centering
		\includegraphics[width=3.9cm]{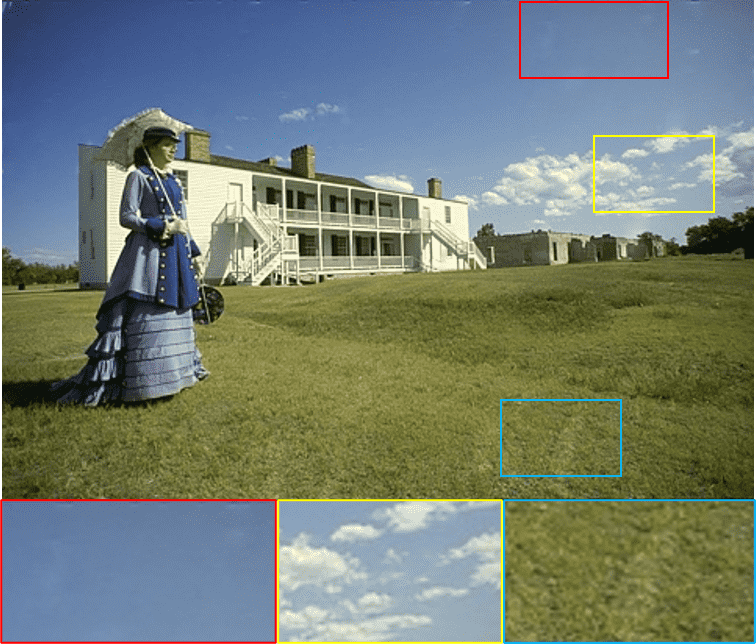}\vspace{-1pt}
		\footnotesize{$Q=4$, $\lambda=1.0$ (34.79/0.977)}\vspace{5pt}
	\end{minipage}
	\begin{minipage}[t]{0.22\linewidth}
		\centering
		\includegraphics[width=3.9cm]{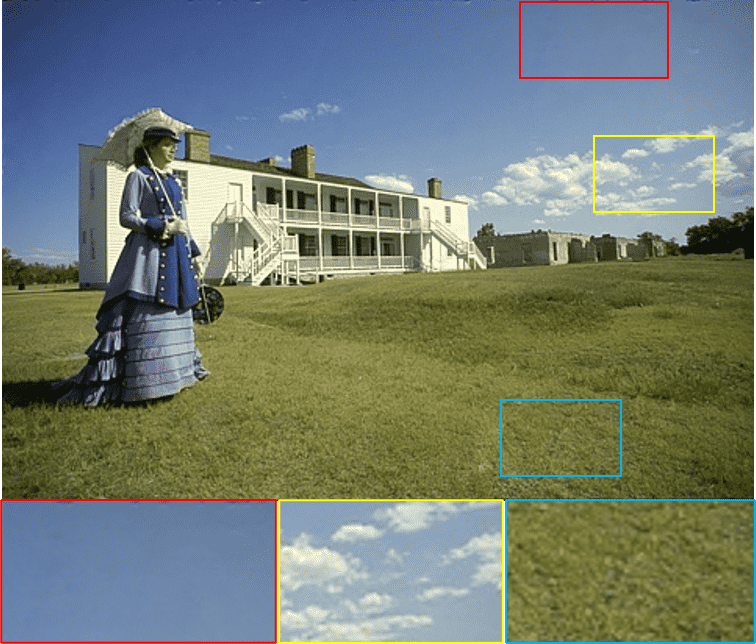}\vspace{-1pt}
		\footnotesize{$Q=4$, $\lambda=5.0$ (\textbf{35.81/0.980})}\vspace{5pt}
	\end{minipage} 
	\begin{minipage}[t]{0.22\linewidth}
		\centering
		\includegraphics[width=3.9cm]{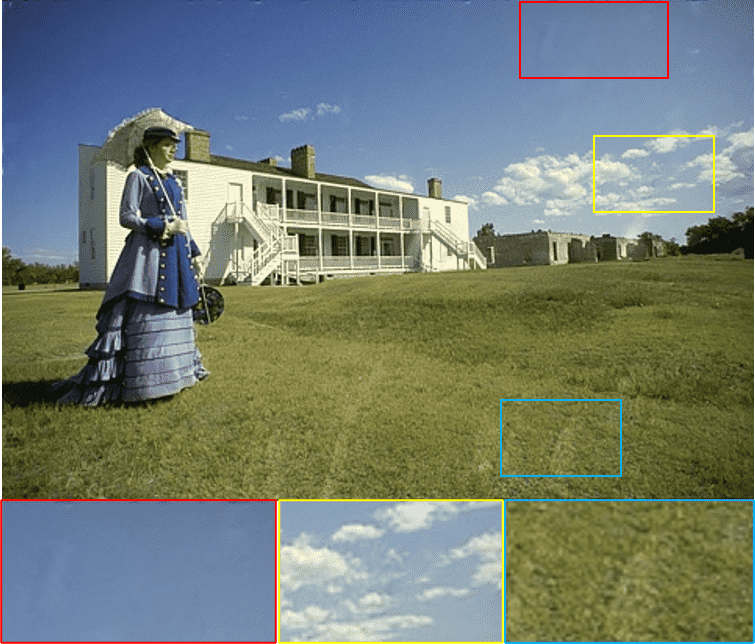}\vspace{-1pt}
		\footnotesize{$Q=4$, $\lambda\to\infty$ (33.42/0.973)} \vspace{5pt}
	\end{minipage}
	\begin{minipage}[t]{0.22\linewidth}
		\centering
		\includegraphics[width=3.9cm]{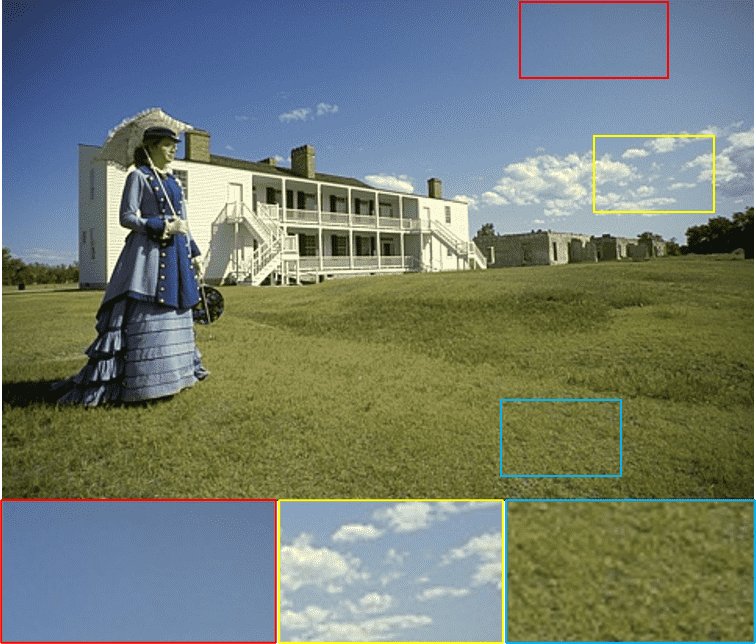}\vspace{-1pt}
		\footnotesize{Ground Truth (Inf/1.0)}\vspace{5pt}
	\end{minipage}
	\caption{Image rain removal results of MSResNet with different $Q$ and with different $\lambda$ in loss function.}
	\label{fig: lambda-example}
\end{figure*}

We use RIS to train all of the above networks on the entire Rain100L train-set and then evaluate them on the Rain100L test-set. Table \ref{tab: modules} displays the results in terms of PSNR/SSIM measurements. Adding multi-scale design or SE blocks can both enhance performance, as illustrated in Table \ref{tab: modules}. Furthermore, multi-scale design and SE blocks have mutually beneficial impacts. We then conduct experiments on the number of MSB and RB in MSResNet, denoted as $Q$. From Table \ref{tab: modules}, we find that $Q=4$ brings the best performance. Visual examples on the first row of Fig. \ref{fig: lambda-example} depict that MSResNet with $Q=3$ tends to under-derain the grass and clouds and $Q=5$ tend to over-derain the sky.

\begin{figure}[b]
	\centering
	\setlength{\abovecaptionskip}{0.01in}
	\includegraphics[width=0.7\linewidth]{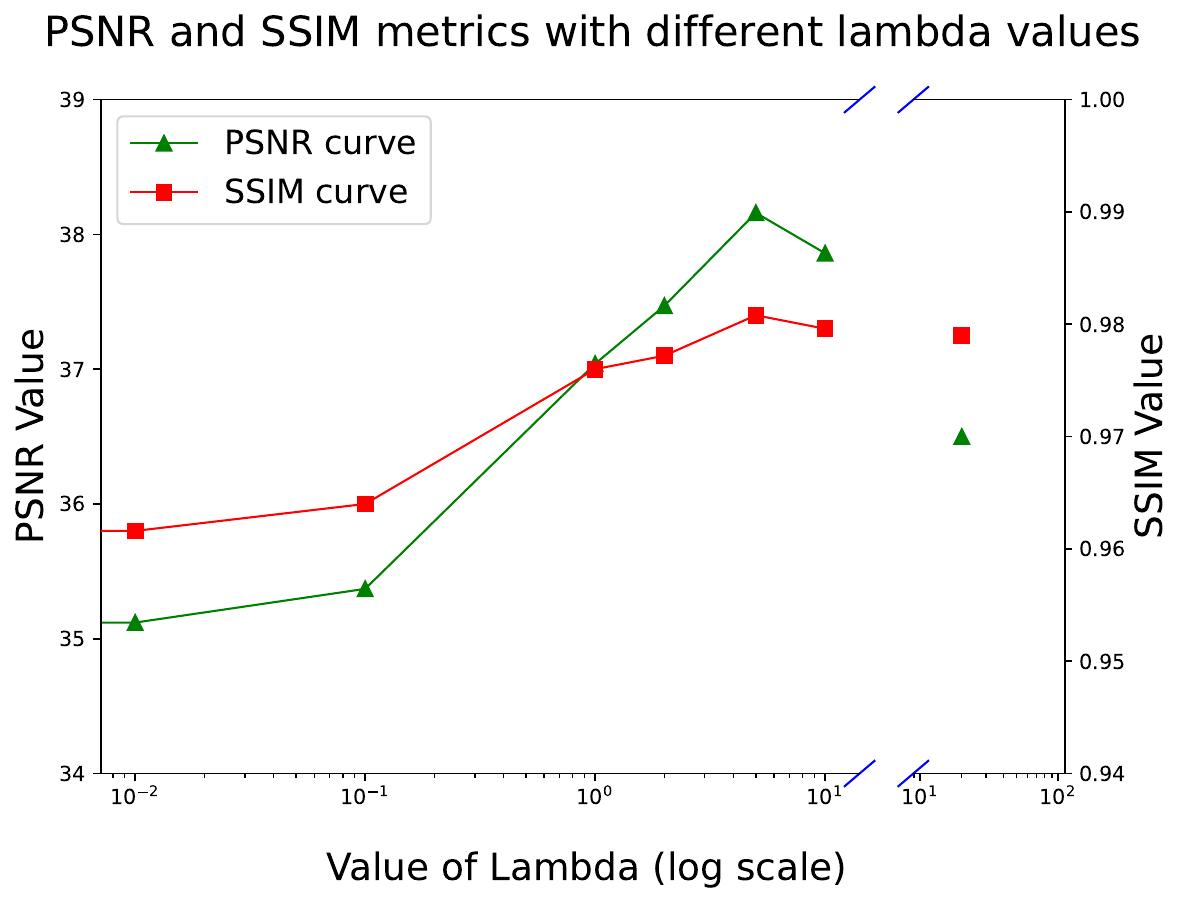}
	\caption{Test PSNR and SSIM Metrics when $\lambda$ in Eq. (\ref{eq:loss}) varies from 0.01 to 10. The points on the right are obtained by using only SSIM loss for training, which indicates $\lambda$ is infinity in Eq. (\ref{eq:loss}).}
	\label{fig: lambda}
\end{figure}

\begin{figure}[htbp]
	\centering
	\setlength{\abovecaptionskip}{0.01in}
	\begin{minipage}[t]{0.45\linewidth}
		\centering
		\includegraphics[width=4.0cm]{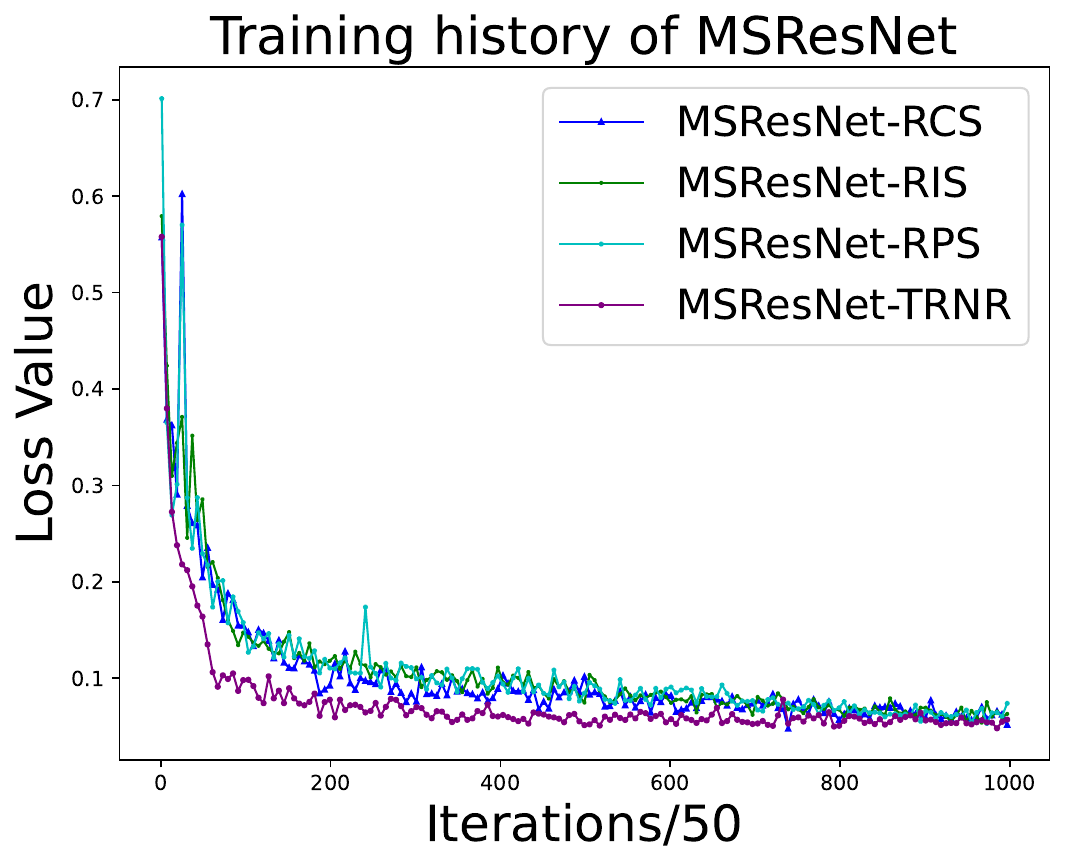}\\
		\footnotesize{(a). MSResNet}
	\end{minipage} 
	\begin{minipage}[t]{0.45\linewidth}
		\centering
		\includegraphics[width=4.0cm]{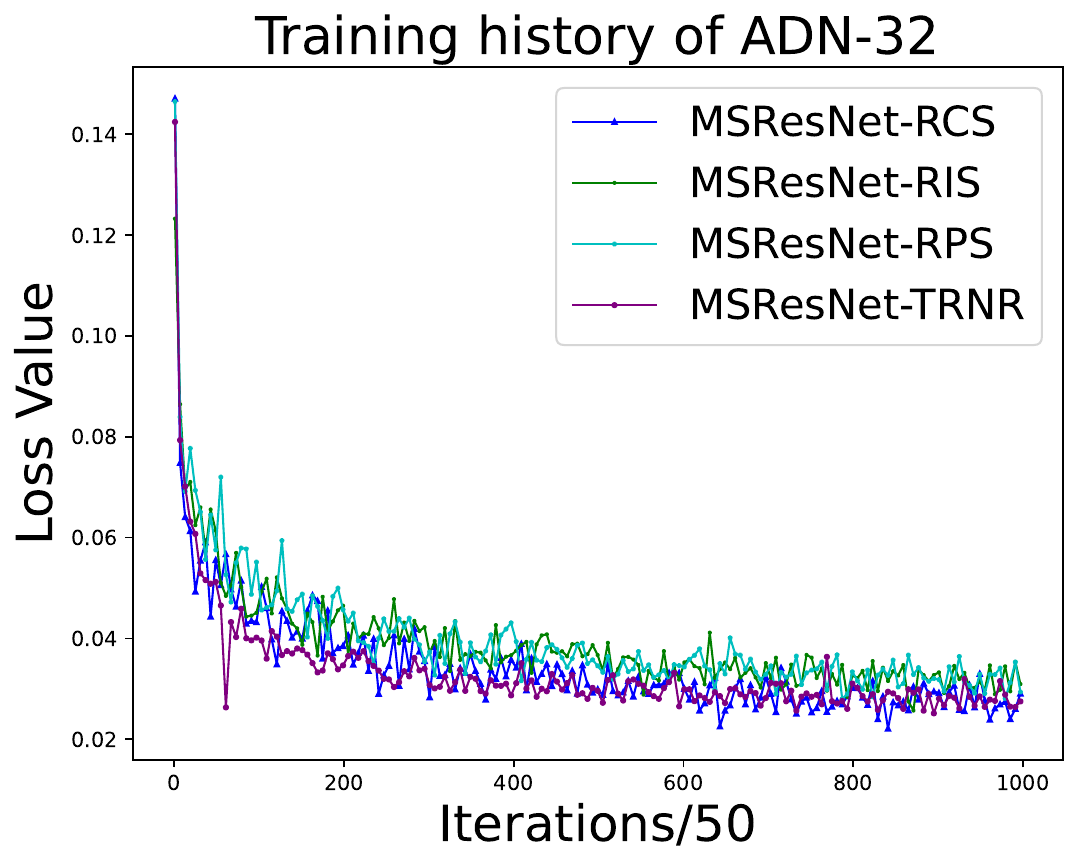}\\
		\footnotesize{(b). ADN-32}
	\end{minipage} 
	\vspace{3.0pt}
	\caption{Loss curve for MSResNet and ADN-32 using RIS, RPS, RCS and TRNR strategies. Both MSResNet and ADN-32 are trained on Rain100L-S.}
	\label{fig:loss}
\end{figure}

\noindent \textbf{Ablation on Loss Function.} Since SSIM loss has been proven to boost image rain removal performance~\cite{Yang2019acm,Wang2019SPAnet,DCSFN,JDNet,BRN2020TIP}, we make an ablation study on $\lambda$ which leverages the SSIM loss term in Eq. (\ref{eq:loss}).  Fig. \ref{fig: lambda} shows the testing PSNR and SSIM metrics on Rain100L when $\lambda$ varies. Fig. \ref{fig: lambda-example} demonstrates that the SSIM loss term leads to more reasonable image rain removal results. However, only employing SSIM loss term tend to under-derain.

\noindent\textbf{Comparison of Learning Strategies.} We compare the data-driven learning strategies using RIS, RPS, and RCS, as well as the task-driven learning strategy TRNR. We train MSResNet using RIS, RPS, RCS, and TRNR for 50,000 iterations on Rain100L-S and use the test-set of Rain100L for evaluation. The evaluation results are presented in Table \ref{tab: strategies}. Quantitative results in Table \ref{tab: strategies} indicate that MSResNet using the RIS strategy offers the worst performance. MSResNet using TRNR has achieved 38.17/0.984 in terms of PSNR and SSIM metrics. TRNR has improved the PSNR metric by 0.99dB and the SSIM metric by 0.4\% when compared to the RIS strategy, demonstrating TRNR's effectiveness. 

To further reveal the effectiveness of TRNR, we also apply TRNR to the existing models DDN~\cite{Fu2017DDN} and ADN-32~\cite{ADN}. We implement the ADN following the statements in~\cite{ADN}. In practice, we train DDN and ADN-32 on Rain100-S using the RIS, RPS, RCS, and the TRNR strategies for 50,000 iterations, and we tabulate the quantitative results in Table \ref{tab: ddn-adn}. 
Note that we employ L1 loss to train DDN to obtain better performances, and we use the default loss setting in ~\cite{ADN} for ADN-32. The quantitative results in Table \ref{tab: ddn-adn} indicate that TRNR can improve performances against the RIS, RPS, and RCS strategies. Unlike most image rain removal methods, such as ADN-32, DDN forces the neural network to learn directly on high-frequency details by omitting the base layer for training. In particular, DDN employs guided filtering to extract low-frequency contexts into the base layer. Hence TRNR provides marginal improvement for DDN as shown in Table V because the redundant background, such as the \textit{sky} scene, is eliminated from the rainy input before being fed into the neural network and the problem of low image utilization no longer exists. However, such a strategy causes DDN to fail to handle accumulated and complex rain occasions, where rain inevitably occurs in the low-frequency base layer. Therefore, for the majority of image rain removal methods learning from both the low-frequency contexts and high-frequency details, TRNR is still appropriate.

We also observe that TRNR can lead to more stable training with lower loss values, as is illustrated by the entire training loss histories of MSResNet and ADN-32 using RIS, RPS, RCS, and TRNR in Fig. \ref{fig:loss}.

\begin{table}[h]
	\centering
	\small
	\renewcommand{\arraystretch}{1.1}
	\renewcommand\tabcolsep{3.5pt}
	\caption{Quantitative Comparison of Data-Driven Learning Strategies and Task-Driven Learning Strategy. The Average PSNR and SSIM Metrics are Computed on the Test-Set of Rain100L.}
	\begin{tabular}{l|c c c c c}
		\bottomrule
		Learning strategies & \multicolumn{4}{c}{Different combinations} \\
		\bottomrule
		Data-Driven+RIS & $\surd$ & $\times$ & $\times$ & $\times$ \\
		Data-Driven+RPS  & $\times$ & $\surd$ & $\times$ & $\times$ \\
		Data-Driven+RCS & $\times$ & $\times$ & $\surd$ & $\times$ \\
		Task-Driven+TRNR & $\times$ & $\times$ & $\times$ & $\surd$ \\
		\hline
		PSNR & 37.18 & 37.88 & 38.06 & \textbf{38.17} \\
		SSIM & 0.980 & 0.983 & 0.983 & \textbf{0.984} \\
		\bottomrule
	\end{tabular}
	\label{tab: strategies}
\end{table}

\begin{table}[htbp]
	\centering
	\small
	\renewcommand{\arraystretch}{1.2}
	\renewcommand\tabcolsep{2.6pt}
	\caption{PSNR/SSIM Evaluation Results of DDN and ADN-32 on Rain100L Test-Set by Training on Rain100L-S using RIS, RPS, RCS, and TRNR.}
	\begin{tabular}{l|c c c c}
		\bottomrule
		Models & RIS & RPS & RCS & TRNR \\
		\bottomrule
		DDN & 34.00/0.957 & 34.02/0.958 & 33.97/0.958 & \textbf{34.03/0.958} \\ 
		ADN-32  & 35.64/0.974 & 35.66/0.974 & 35.71/0.975 & \textbf{35.88/0.975}\\
		\bottomrule
	\end{tabular}
	\label{tab: ddn-adn}
\end{table}

\begin{figure}[htbp]
	\centering
	\setlength{\abovecaptionskip}{0.01in}
	\begin{minipage}[t]{0.9\linewidth}
		\centering
		\includegraphics[width=8.2cm]{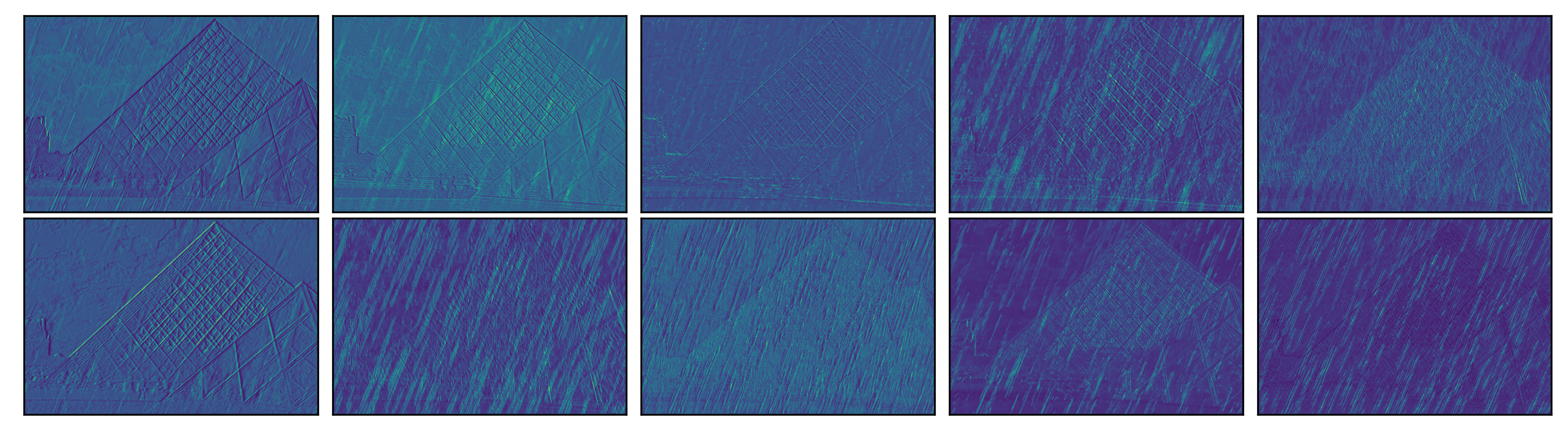}\\
		\footnotesize{(a). Feature of MSResNet using RIS}
	\end{minipage} \\
	\begin{minipage}[t]{0.9\linewidth}
		\centering
		\includegraphics[width=8.2cm]{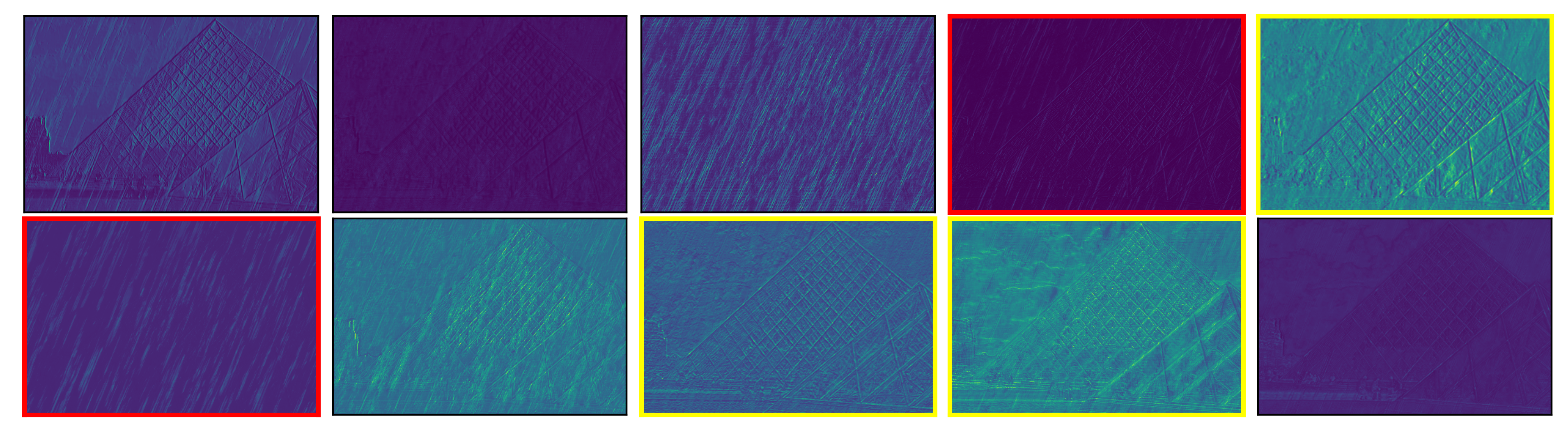}\\
		\footnotesize{(b). Feature of MSResNet using TRNR}
	\end{minipage} \\
	\vspace{3.0pt}
	\caption{Top-10 activated features of the last MSB in MSResNet. (a) features of MSResNet trained on Rain100L-S using RIS. (b) features of MSResNet trained on Rain100L-S using TRNR.}
	\label{fig:feature-vis}
\end{figure}

\noindent\textbf{Feature Visualization Analysis.} TRNR has been verified in Table \ref{tab: strategies} to improve the performance of MSResNet when data is limited. In this section, we investigate what features TRNR can learn. We choose the Top-10 activated features of the last MSB in MSResNet with a large receptive field for visualization. For comparison, MSResNet with RIS and TRNR on Rain100-S is used. As depicted in Fig. \ref{fig:feature-vis}, the benefits of TRNR are threefold: (I). TRNR can increase the diversity of features, (II). TRNR empowers MSResNet to learn discriminative features. The red boxes and yellow boxes in Fig. \ref{fig:feature-vis}(b) indicate that MSResNet with TRNR can capture rain streaks and textures separately, (III). MSResNet with TRNR can retain more texture information at different granularities as presented in the yellow boxes of Fig. \ref{fig:feature-vis} (b).

\noindent\textbf{Task-Driven Learning Settings.} The proposed TRNR has obtained the best performance as shown in Table \ref{tab: strategies}. We then make a further investigation on the task-driven learning settings. In practice, $K$ is fixed to $1$ as some clusters only have 2 image patches (one for $\mathcal{D}^{(train)}_{\tau_j}$ and one for $\mathcal{D}^{(val)}_{\tau_j}$). Therefore, we study various combinations of $N$ in N-frequency-K-shot learning and batch size $R$ of sampled tasks number in Eq. (\ref{eq:outer-loop}). Intuitively, larger $N$ leads to greater memory requirements, and larger $R$ often results in better performance but longer training time as shown in Algorithm 2. Practically, we train MSResNet with $Q=4$ on 5 combinations of $N$ and $R$ and evaluate using test-set of Rain100L. The results in Table \ref{tab: ablation on RN} indicate that $N=12$ with $R=5$ provides the best performance.

\begin{figure}[t]
	\centering
	\setlength{\abovecaptionskip}{0.01in}
	\begin{minipage}[t]{0.45\linewidth}
		\centering
		\includegraphics[width=3.7cm]{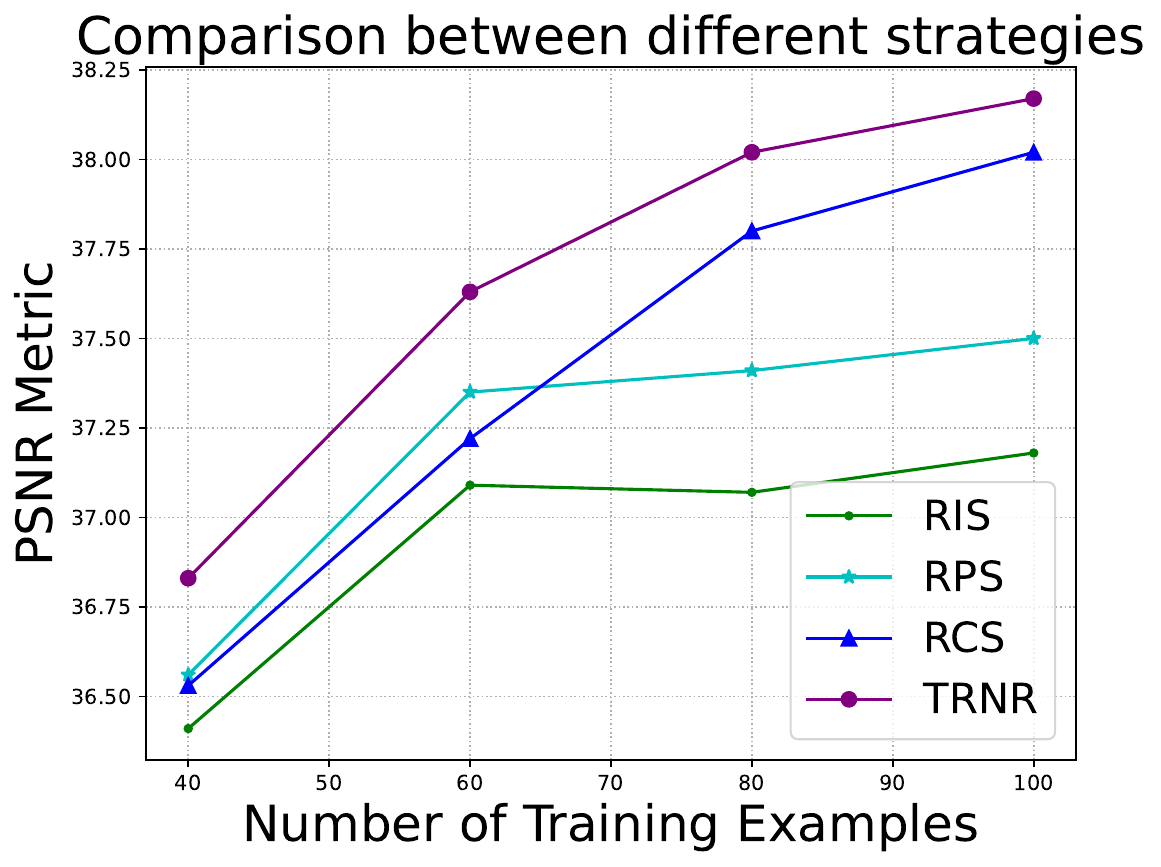}
		\footnotesize{(a). Different train-set size of Rain100L}
	\end{minipage}
	\begin{minipage}[t]{0.45\linewidth}
		\centering
		\includegraphics[width=3.5cm]{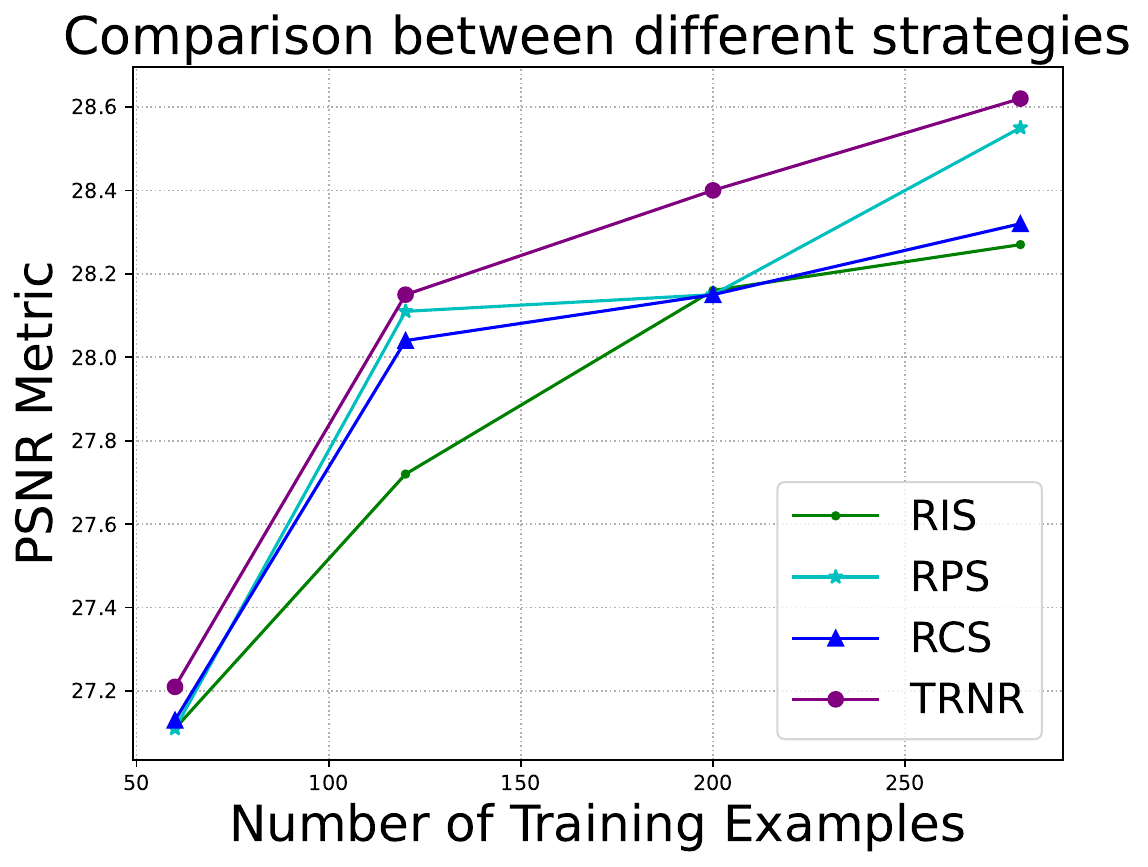}
		\footnotesize{(b). Different train-set size of Rain800}
	\end{minipage} 
	\caption{Comparison of Rain100L and Rain800 RIS, RPS, and RCS with proposed TRNR when train-set size varies. (a) represents the Rain100L testing PSNR metric when the train-set size is increased from 40 to 100 images. (b) depicts the Rain800 testing PSNR metric with train-set sizes ranging from 60 to 280.}
	\label{fig: dataset size}
\end{figure}

\noindent\textbf{Analysis on Number of Clusters. }As elaborated in \S\ref{subsec: pa}, the image patches are clustered according to its spatial and statistical properties. Different clustering result affects RCS and may influence the performance of TRNR. Therefore we use all images in Rain100L-S to build two datasets with different numbers of clusters, which are noted as Rain100L-S1K (992 clusters) and Rain100L-S2K (1958 clusters). For evaluation, we train MSResNet using TRNR on Rain100L-S1K and Rain100L-S2K. The testing PSNR/SSIM metrics of MSResNet-TRNR using Rain100L-S1K and Rain100L-S2K are 37.81/0.983 and 37.91/0.983, respectively. Compared with testing results of MSResNet-TRNR using Rain100L-S in Table \ref{tab: strategies}, we find that a larger cluster number results in better performance. Specifically, fine-grained clustering with more clusters can reduce the gap between $D^{(train)}_{\tau_j}$ and $D^{(val)}_{\tau_j}$, which both contain $NK$ examples from $N$ clusters and each with $K$ examples. Hence $\mathbf{g}^{(val)}_j$ and $\mathbf{g}^{(train)}_{j}$ can be well aligned in Eq. (\ref{eq: TRNR-loss}), resulting in a significant update for $\theta$ with less noise. However, an excessive number of clusters may bring high computation costs, and will hardly boost the performance when all image patches are well clustered.

\noindent\textbf{Analysis on Dataset Size. }How can we obtain a feasible model when the train-set only contains a few samples? In this paper, we propose to maximize the utilization of the dataset and employ a task-driven approach. In this section, we use Rain100L and Rain800 to study the effects of dataset size on RIS, RPS, RCS, and the TRNR strategies. As for Rain100L, we compare the TRNR with RIS, RPS, and RCS strategies when the train-set increases from 40 examples to 100 examples. As illustrated in Fig. \ref{fig: dataset size}(a), TRNR has shown the best performances among all dataset sizes when compared with RIS, RPS, and RCS strategies. For example, when using 80 images for training, TRNR obtains 38.02dB in terms of PSNR metric, with 0.95dB improvement over the RIS strategy, and 0.61dB improvement over the RPS strategy, and 0.22dB improvement over the RCS strategy. We can also observe that RIS suffers a heavy performance drop against the TRNR due to the low utilization of images. 

Furthermore, we utilize Rain800 to investigate the influence of dataset size when faced with accumulated and complex rain streaks. In practice, we increase the train-set size for Rain800 from 60 to 280. The corresponding evaluation results for different strategies are depicted in Fig. \ref{fig: dataset size}(b). Similarly, TRNR has shown the best performance among all strategies when train-set changes. For instance, when using 200 examples in Rain800 for training, TRNR has obtained a 0.24dB gain over RIS, and a 0.25dB improvement over both RPS and RCS strategies, in terms of the PSNR metric.  

\begin{table}[htbp]
	\centering
	\small
	\renewcommand{\arraystretch}{1.2}
	\renewcommand\tabcolsep{3.5pt}
	\caption{Average PSNR and SSIM Metrics on the Test-Set of Rain100L of Different $N$ and $R$ Combinations ($K$ is Set to 1).}
	\begin{tabular}{l|c c c c c}
		\bottomrule
		($N$, $R$) & (32, 1) & (26, 2) & (18, 3) & (14, 4) & (12, 5)\\
		\hline
		PSNR & 37.75 & 37.91 & 38.07 & 37.97 & \textbf{38.17} \\ 
		SSIM  & 0.982 & 0.983 & 0.984 & 0.983 & \textbf{0.984}\\
		\bottomrule
	\end{tabular}
	\label{tab: ablation on RN}
\end{table}

\noindent\textbf{Learn More from Less. }TRNR has shown its superiority over RIS, RPS, and RCS. We then compare the performances of MSResNet trained with TRNR on smaller datasets and RIS on entire datasets. As shown in Table \ref{tab: de-rain comparison}, we observe that MSResNet-TRNR offers higher PSNR/SSIM metrics against MSResNet-RIS on Rain100L, Rain100H, and Rain800 datasets. For instance, testing PSNR/SSIM metrics for MSResNet-TRNR on Rain100H are 28.69/0.899, with 0.95dB/1.1\% gain in terms of PSNR/SSIM against MSResNet-RIS. This observation demonstrates the MSResNet-TRNR can learn more with fewer images. On the other hand, we use the entire train-set of Rain100L to create a clustered dataset termed Rain100L-Full, which contains 45754 patch pairs and 3496 clusters. MSResNet-TRNR with Rain100L-Full has offered 38.59/0.985 testing PSNR/SSIM results, representing 0.42dB and 0.43dB improvements over MSResNet-TRNR using Rain100L-S and MSResNet-RIS using the entire train-set respectively.
\begin{table*}[htbp]
	\centering
	\small
	\renewcommand{\arraystretch}{1.03}
	\renewcommand\tabcolsep{5.5pt}
	\caption{Quantitative Evaluation Results on Rain100L, Rain100H, Rain800, RainLight, RainHeavy, and Rain1400 Datasets of MSResNet-TRNR and Other Recent Learning-Based Methods. Top: Models Trained with Entire Train-set in Table \ref{tab: RIS datasets}. Bottom: Models Trained with Small Train-set in Table \ref{tab: TRNR datasets}.The Best Results for Methods on the Top are \textbf{Bolded} in {\color{cyan}Cyan}, and The Second and Third Results are Highlighted in {\color{red}Red} and {\color{blue}Blue} Colors, Respectively. We \textbf{Bold} the Best Results for Methods at the Bottom.}
		\begin{tabular}{ c | l | c | c | c | c | c | c}
			\bottomrule
			Train-set & Methods  & Rain100L & Rain100H & Rain800 & RainLight & RainHeavy & Rain1400 \\
			\bottomrule
			\multirow{18}*{\makecell[c]{Entire\\Train-set}} & Rainy & 25.52/0.825 & 12.13/0.349 & 21.15/0.651 & 25.34/0.830 & 11.68/0.343 & 23.69/0.757 \\
			& JCAS (ICCV'17) \cite{Gu2017JCAS} & 28.40/0.881 & 13.65/0.459 & 22.19/0.766 & - & - & 28.77/0.819\\
			& DerainNet (TIP'17)~\cite{clearingthesky} & 27.36/0.855 & 13.94/0.403 & 23.14/0.771 & 32.35/0.945 & 23.12/0.735 & 27.53/0.866 \\
			& DDN (CVPR'17) \cite{Fu2017DDN} & 32.04/0.938 & 24.95/0.781 & 21.16/0.732 & 32.86/0.947 & 20.12/0.635 & 27.61/0.901 \\
			& JORDER (CVPR'17) \cite{Yang2017Jorder}  & 36.11/0.970 & 22.15/0.674 & 22.24/0.776 & - & - & 27.55/0.853\\ 
			& RESCAN (ECCV'18) \cite{Xia2018Rescan} & 36.64/0.975 & 26.45/0.846 & 24.09/0.841 & 36.00/0.973 & 25.92/0.823 & 28.57/0.891\\
			& DID-MDN (CVPR'18) \cite{Zhang2018DID} & 25.70/0.858 & 17.39/0.612 & 21.89/0.795 & - & 15.54/0.520 & 27.99/0.869 \\
			& DualCNN (CVPR'18) \cite{DualCNN}  & 26.87/0.860 & 14.23/0.468 &  24.09/0.841 & - & - & 24.98/0.838\\
			& PReNet (CVPR'19) \cite{PreNet}  & 36.28/0.977 & 27.65/0.882 & 25.33/0.860 & 37.58/0.983 & 28.32/0.895 & 30.73/0.918\\
			& ReHEN (ACM'MM'19) \cite{Yang2019acm}  & {\color{red}37.41}/{\color{blue}0.980} & 27.97/0.864 & 26.96/0.854  & {\color{red}38.06}/{\color{red}0.983} & 27.76/0.868 & {\color{blue}31.33}/0.918 \\
			& JORDER-E (TPAMI'19) \cite{JORDERE}  & 37.09/0.979 & 27.84/0.862  & 25.90/0.844 & 37.19/0.981 & 27.53/0.862& - \\
			& {\color{magenta}DCSFN} (ACM'MM'20) \cite{DCSFN}  & 35.25/0.976 & \textbf{{\color{cyan}29.70}}/\textbf{{\color{cyan}0.907}} & 26.49/0.855 & {\color{blue}37.73}/{\color{blue}0.983} & 28.28/0.895 & 30.38/0.912  \\
			& {\color{magenta}BRN} (TIP'20) \cite{BRN2020TIP}  & 36.75/0.979 & {\color{red}29.02}/{\color{red}0.900} & - & 37.45/0.983 & {\color{blue}28.84}/{\color{red}0.904} & 30.74/0.916 \\
			& MPRNet{$^{\dagger}$} (CVPR'21) \cite{MPRNet}  & 35.01/0.960 & 28.54/0.872 & \textbf{{\color{cyan}28.75}}/{\color{red}0.876} & 34.41/0.957 & 27.65/0.871 & {\color{red}32.04}/{\color{red}0.928}\\
			& RLNet (CVPR'21) \cite{RLNet}  & 37.38/0.980 & - & {\color{blue}27.95}/{\color{blue}0.870} & 36.56/0.978 & {\color{red}28.87}/{\color{blue}0.895} & 30.72/0.916\\ 
			& JDDGD (TCSVT'21) \cite{JDDGD}  & 36.66/0.975 & {\color{blue}28.95}/0.886 & - & - & - & \textbf{{\color{cyan}33.74}}/\textbf{{\color{cyan}0.953}}\\
			& {\color{magenta}ADN-64} (ICASSP'21) \cite{ADN}  & {\color{blue}37.39}/{\color{red}0.981} & - & 27.22/0.866 & - & - & 31.19/0.920\\
			& MSResNet-RIS (\textbf{Ours}) & \textbf{{\color{cyan}38.16}}/\textbf{{\color{cyan}0.984}} & 27.74/0.888 & {\color{red}28.49}/\textbf{{\color{cyan}0.885}} & \textbf{{\color{cyan}38.66}}/\textbf{{\color{cyan}0.986}} & \textbf{{\color{cyan}28.94}}/\textbf{{\color{cyan}0.909}} & 31.26/{\color{blue}0.921}\\
			\hline
			\multirow{4}*{\makecell[c]{Partial\\Train-set}} & {\color{magenta}DCSFN} (ACM'MM'20)~\cite{DCSFN} & 31.69/0.942 & 23.86/0.798 & 24.46/0.838 & 35.44/0.974 & 25.08/0.841 & 29.42/0.898 \\
			& {\color{magenta}BRN} (TIP'20)~\cite{BRN2020TIP}  & 35.47/0.974 & 26.05/0.855 & 25.45/0.868 & 36.26/0.979 & 27.13/0.879 & 26.68/0.881 \\
			& {\color{magenta}ADN-64} (ICASSP'21)~\cite{ADN} & 36.43/0.976 & 26.94/0.874 & 26.43/0.866 & 37.55/0.983 & 27.21/0.882 & 26.30/0.873  \\
			& MSResNet-TRNR (\textbf{Ours}) & \textbf{38.17}/\textbf{0.984} & \textbf{28.69}/\textbf{0.899}& \textbf{28.62}/\textbf{0.892} & \textbf{38.38}/\textbf{0.985} & \textbf{28.48}/\textbf{0.907} & \textbf{30.66}/\textbf{0.917} \\
			\bottomrule
		\end{tabular}
	\label{tab: de-rain comparison}
\end{table*}
\begin{figure*}[htbp]
	\centering
	\setlength{\abovecaptionskip}{0.01in}
	\begin{minipage}[t]{0.15\linewidth}
		\centering
		\includegraphics[width=2.6cm]{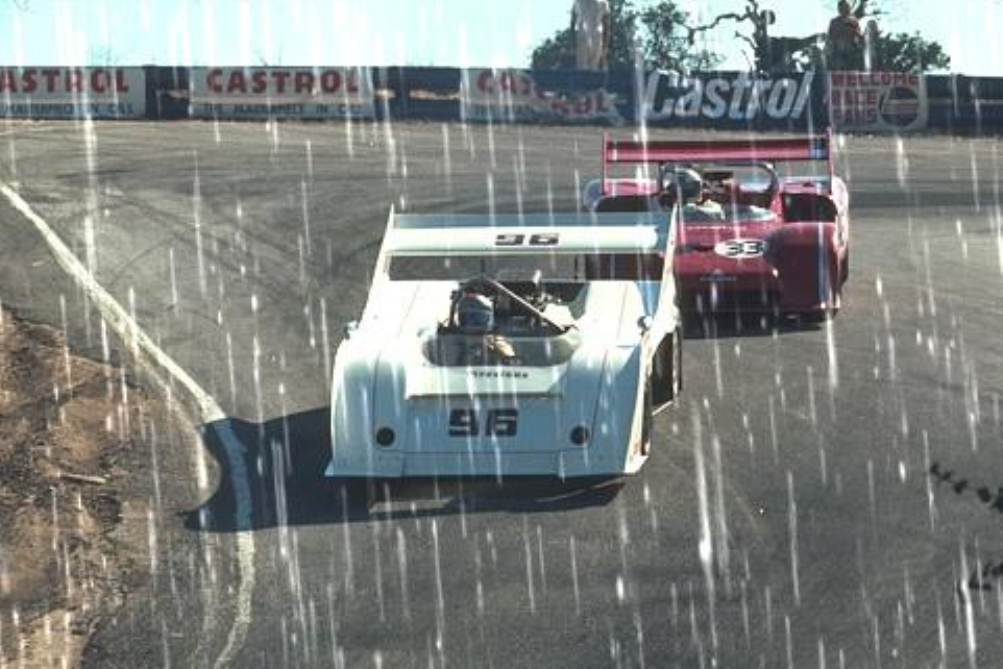}\vspace{-1pt}
		\footnotesize{23.03/0.816}\vspace{5pt}
	\end{minipage}
	\begin{minipage}[t]{0.15\linewidth}
		\centering
		\includegraphics[width=2.6cm]{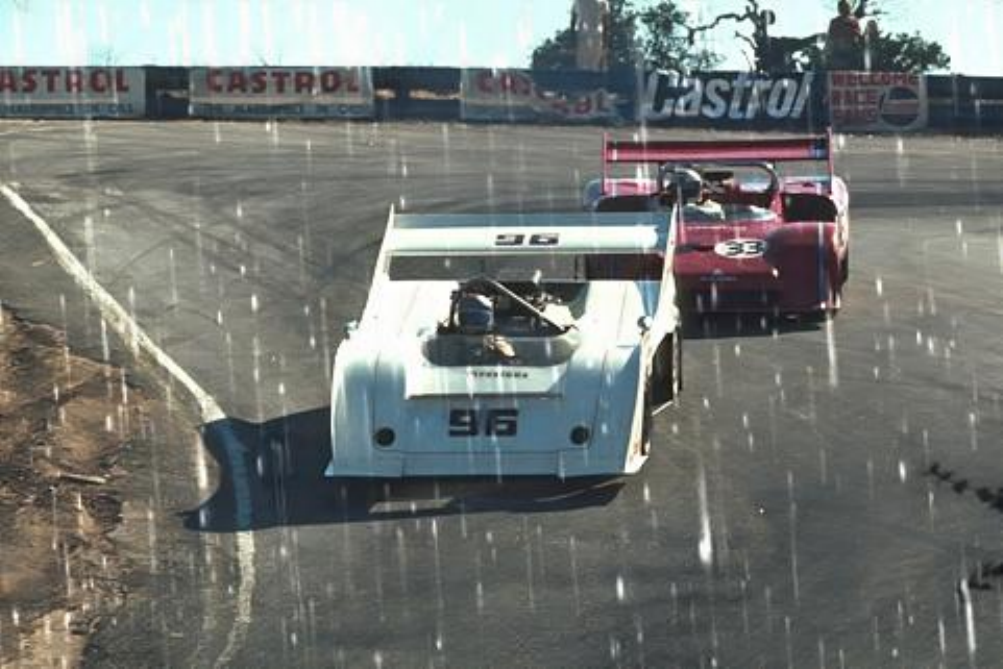}\vspace{-1pt}
		\footnotesize{25.70/0.857}\vspace{5pt}
	\end{minipage} 
	\begin{minipage}[t]{0.15\linewidth}
		\centering
		\includegraphics[width=2.6cm]{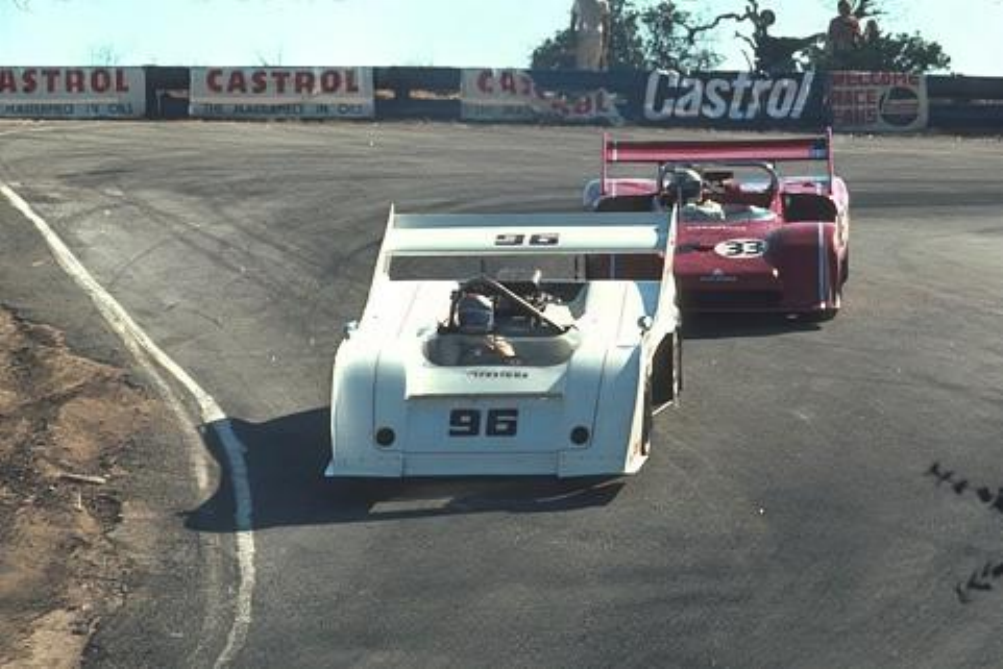}\vspace{-1pt}
		\footnotesize{34.42/0.972} \vspace{5pt}
	\end{minipage}
	\begin{minipage}[t]{0.15\linewidth}
		\centering
		\includegraphics[width=2.6cm]{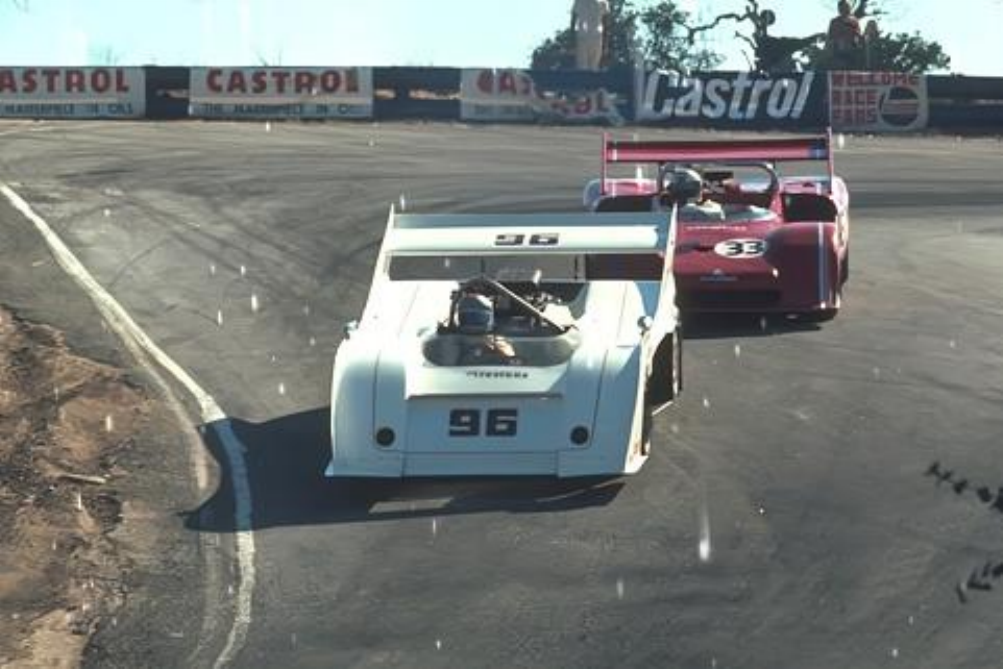}\vspace{-1pt}
		\footnotesize{31.85/0.939}\vspace{5pt}
	\end{minipage}
	\begin{minipage}[t]{0.15\linewidth}
		\centering
		\includegraphics[width=2.6cm]{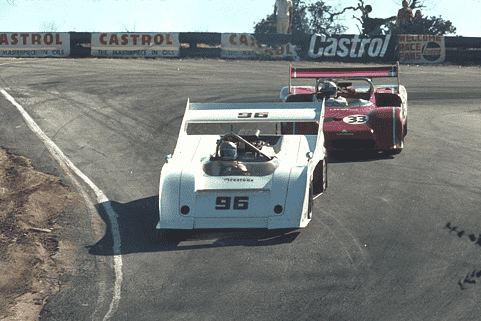}\vspace{-1pt}
		\footnotesize{\textbf{36.78/0.982}}\vspace{5pt}
	\end{minipage}
	\begin{minipage}[t]{0.15\linewidth}
		\centering
		\includegraphics[width=2.6cm]{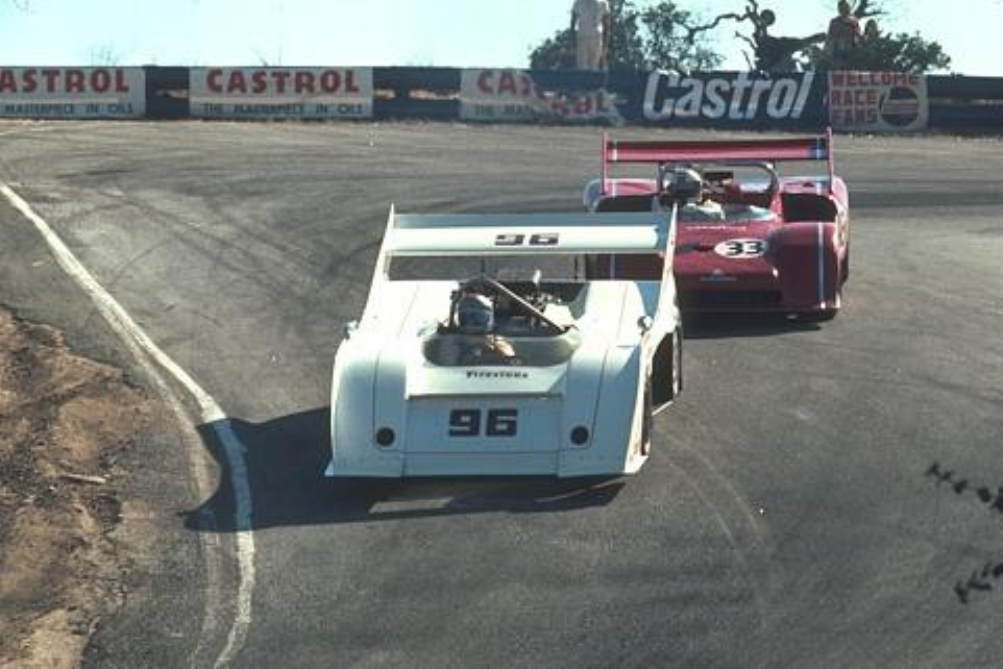}\vspace{-1pt}
		\footnotesize{Inf/1.0}\vspace{5pt}
	\end{minipage} \\ 
	
	\begin{minipage}[t]{0.15\linewidth}
		\centering
		\includegraphics[width=2.6cm]{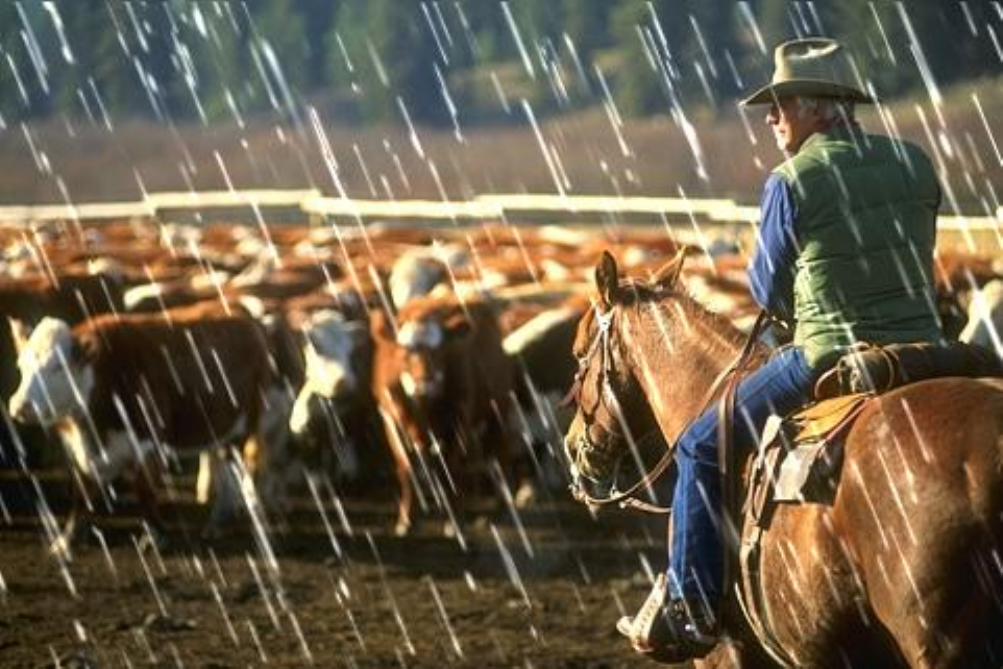}\vspace{-1pt}
		\footnotesize{20.32/0.0.553 \\ Rainy}\vspace{5pt}
	\end{minipage}
	\begin{minipage}[t]{0.15\linewidth}
		\centering
		\includegraphics[width=2.6cm]{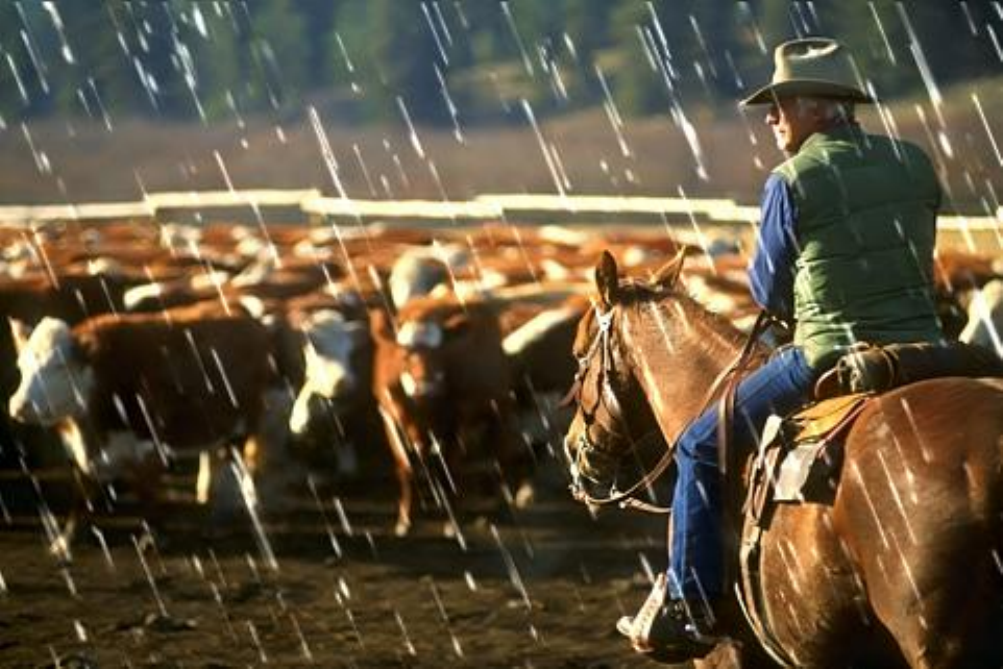}\vspace{-1pt}
		\footnotesize{21.98/0.769 \\ DDN \cite{Fu2017DDN}}\vspace{5pt}
	\end{minipage} 
	\begin{minipage}[t]{0.15\linewidth}
		\centering
		\includegraphics[width=2.6cm]{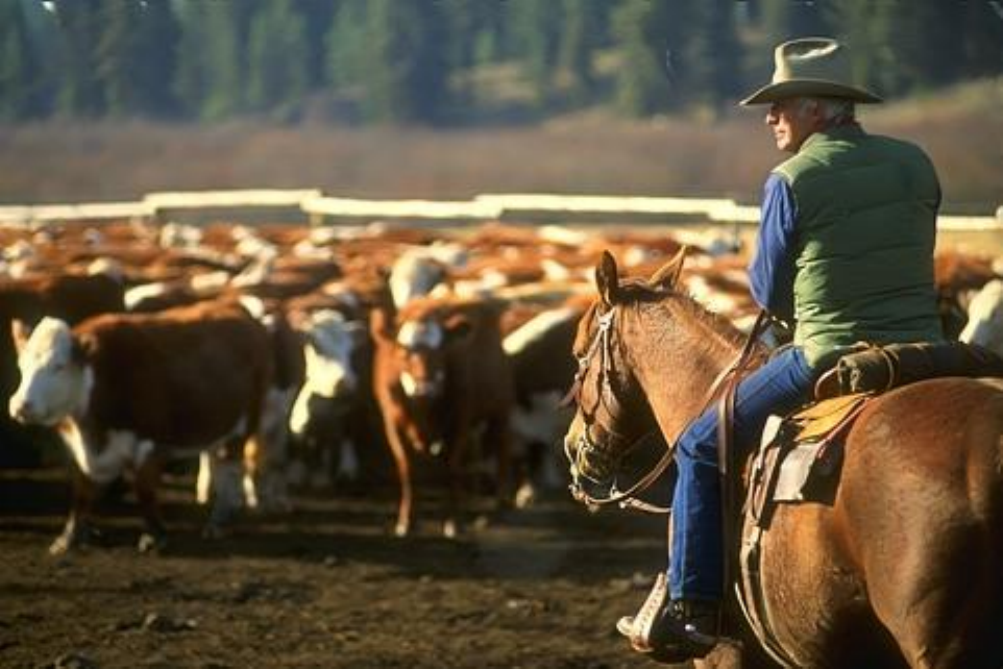}\vspace{-1pt}
		\footnotesize{33.32/0.972 \\ DCSFN \cite{DCSFN} }\vspace{5pt}
	\end{minipage}
	\begin{minipage}[t]{0.15\linewidth}
		\centering
		\includegraphics[width=2.6cm]{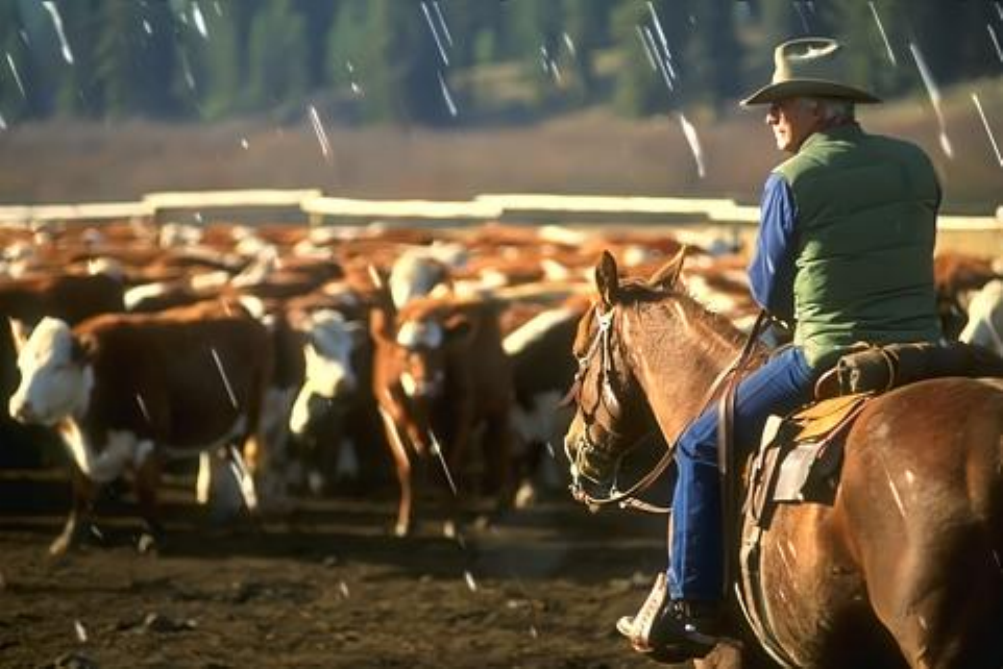}\vspace{-1pt}
		\footnotesize{27.35/0.922 \\ MPRNet~\cite{MPRNet}}\vspace{5pt}
	\end{minipage}
	\begin{minipage}[t]{0.15\linewidth}
		\centering
		\includegraphics[width=2.6cm]{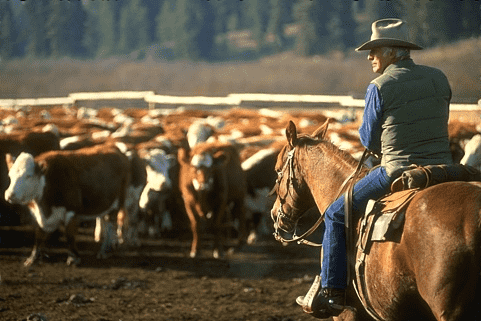}\vspace{-1pt}
		\footnotesize{\textbf{36.78/0.984} \\ MSResNet-TRNR (Ours)}\vspace{5pt}
	\end{minipage}
	\begin{minipage}[t]{0.15\linewidth}
		\centering
		\includegraphics[width=2.6cm]{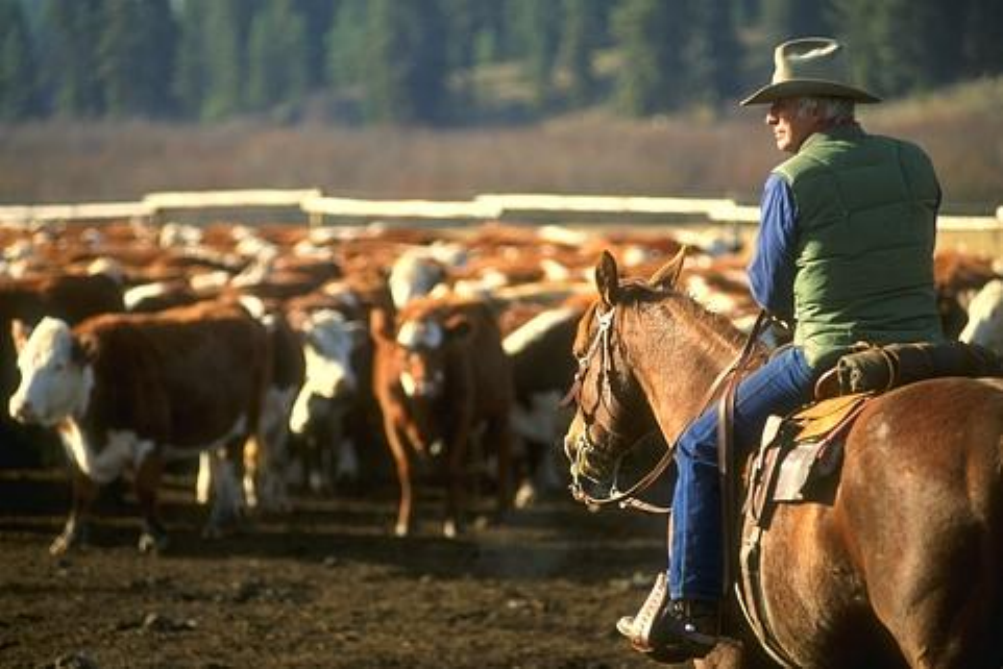}\vspace{-1pt}
		\footnotesize{Inf/1.0 \\ Ground Truth}\vspace{5pt}
	\end{minipage}
	\caption{Image rain removal results of DDN \cite{Fu2017DDN}, DCSFN \cite{DCSFN}, MPReNet$^{\dagger}$ \cite{MPRNet} and proposed method on Rain100L dataset. Values at the bottom of every image indicate the PSNR/SSIM metrics respectively. }
	\label{fig:derain100L}
\end{figure*}

\begin{figure*}[htbp]
	\centering
	\setlength{\abovecaptionskip}{0.01in}
	\begin{minipage}[t]{0.15\linewidth}
		\centering
		\includegraphics[width=2.6cm]{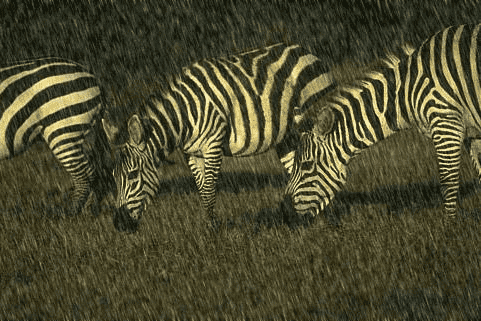}\vspace{-1pt}
		\footnotesize{20.77/0.596}\vspace{5pt}
	\end{minipage}
	\begin{minipage}[t]{0.15\linewidth}
		\centering
		\includegraphics[width=2.6cm]{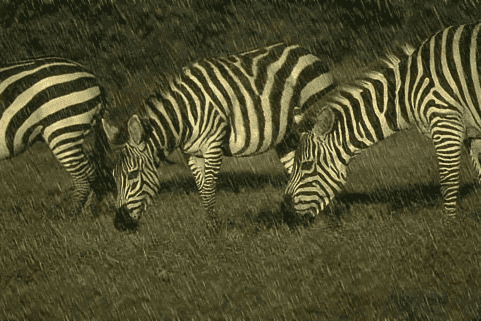}\vspace{-1pt}
		\footnotesize{22.68/0.682}\vspace{5pt}
	\end{minipage} 
	\begin{minipage}[t]{0.15\linewidth}
		\centering
		\includegraphics[width=2.6cm]{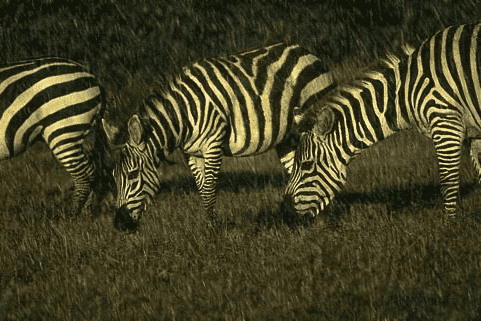}\vspace{-1pt}
		\footnotesize{18.64/0.637}\vspace{5pt}
	\end{minipage}
	\begin{minipage}[t]{0.15\linewidth}
		\centering
		\includegraphics[width=2.6cm]{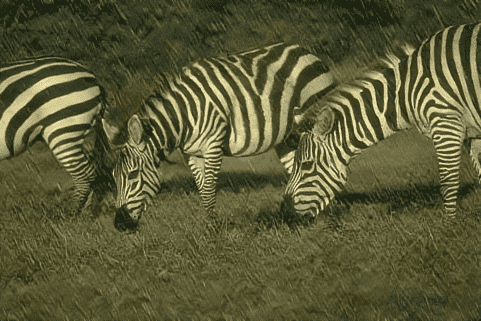}\vspace{-1pt}
		\footnotesize{26.27/0.718}\vspace{5pt}
	\end{minipage}
	\begin{minipage}[t]{0.15\linewidth}
		\centering
		\includegraphics[width=2.6cm]{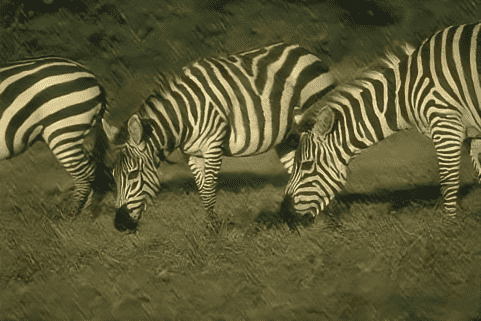}\vspace{-1pt}
		\footnotesize{\textbf{27.51/0.901}}\vspace{5pt}
	\end{minipage}
	\begin{minipage}[t]{0.15\linewidth}
		\centering
		\includegraphics[width=2.6cm]{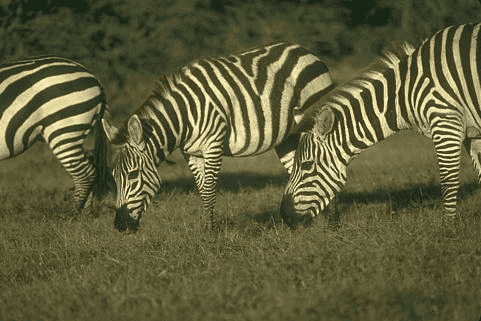}\vspace{-1pt}
		\footnotesize{Inf/1.0}\vspace{5pt}
	\end{minipage} 
	
	\begin{minipage}[t]{0.15\linewidth}
		\centering
		\includegraphics[width=2.6cm]{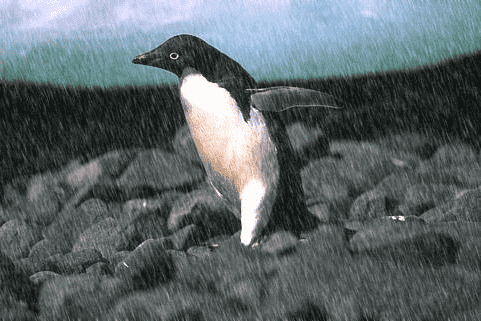}\vspace{-1pt}
		\footnotesize{20.32/0.0.553 \\ Rainy}\vspace{5pt}
	\end{minipage}
	\begin{minipage}[t]{0.15\linewidth}
		\centering
		\includegraphics[width=2.6cm]{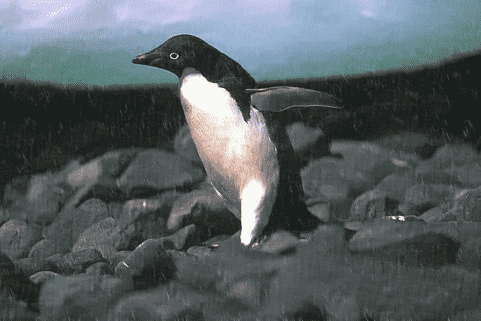}\vspace{-1pt}
		\footnotesize{18.39/0.784 \\ ReHEN \cite{Yang2019acm}}\vspace{5pt}
	\end{minipage} 
	\begin{minipage}[t]{0.15\linewidth}
		\centering
		\includegraphics[width=2.6cm]{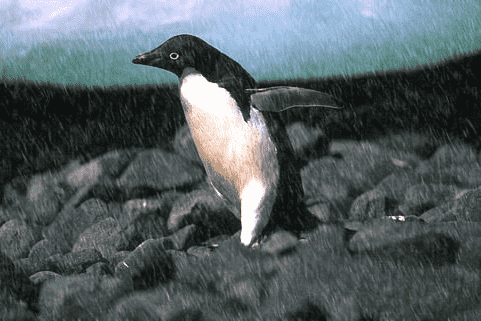}\vspace{-1pt}
		\footnotesize{18.46/0.636 \\ MPRNet~\cite{MPRNet}}\vspace{5pt}
	\end{minipage}
	\begin{minipage}[t]{0.15\linewidth}
		\centering
		\includegraphics[width=2.6cm]{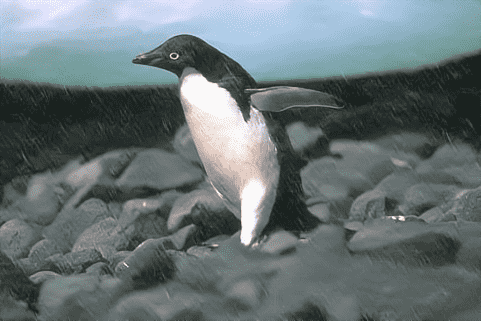}\vspace{-1pt}
		\footnotesize{\textbf{27.30/0.850} \\ MSResNet-RIS (Ours)}\vspace{5pt}
	\end{minipage}
	\begin{minipage}[t]{0.15\linewidth}
		\centering
		\includegraphics[width=2.6cm]{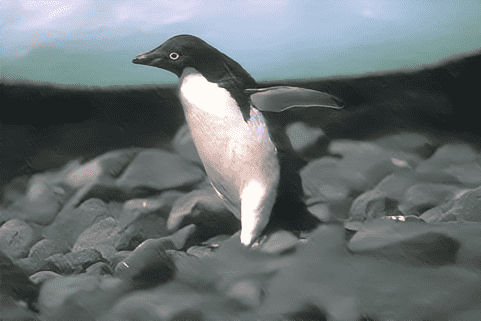}\vspace{-1pt}
		\footnotesize{27.22/0.798 \\ MSResNet-TRNR (Ours)}\vspace{5pt}
	\end{minipage}
	\begin{minipage}[t]{0.15\linewidth}
		\centering
		\includegraphics[width=2.6cm]{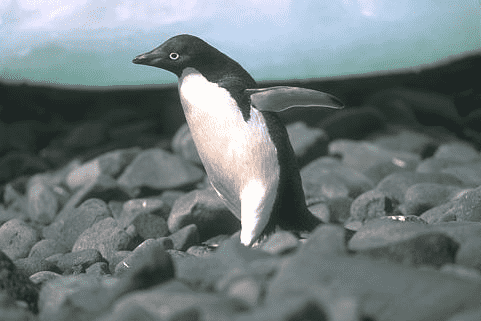}\vspace{-1pt}
		\footnotesize{Inf/1.0 \\ Ground Truth}\vspace{5pt}
	\end{minipage}
	\caption{Image rain removal visual examples of ReHEN \cite{Fu2017DDN}, MPReNet$^{\dagger}$ \cite{MPRNet} and proposed method on Rain800 dataset. Values at the bottom of every image indicate the PSNR/SSIM metrics respectively. ReHEN and MPRNet$^{\dagger}$ both fail to recover the background illuminance. }
	\label{fig:derain800}
\end{figure*}

\subsection{Comparisons on Synthetic Datasets}
\begin{figure*}[htbp]
		\centering
		\begin{minipage}[t]{0.15\linewidth}
			\centering
			\includegraphics[width=2.6cm]{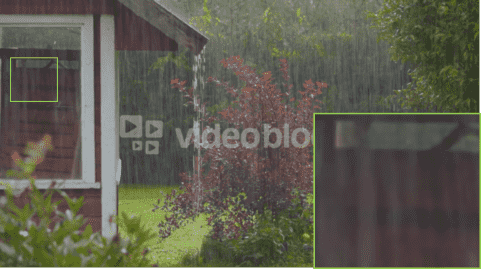}\vspace{4pt}
		\end{minipage}
		\begin{minipage}[t]{0.15\linewidth}
			\centering
			\includegraphics[width=2.6cm]{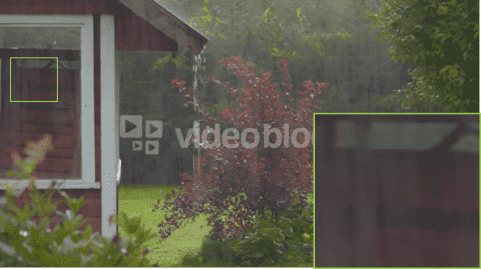}
		\end{minipage}
		\begin{minipage}[t]{0.15\linewidth}
			\centering
			\includegraphics[width=2.6cm]{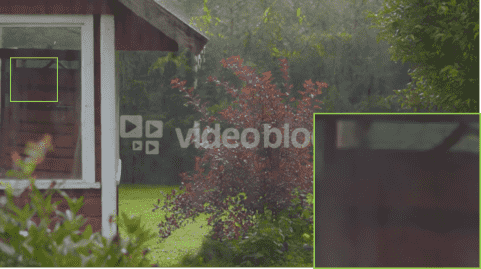}
		\end{minipage}
		\begin{minipage}[t]{0.15\linewidth}
			\centering
			\includegraphics[width=2.6cm]{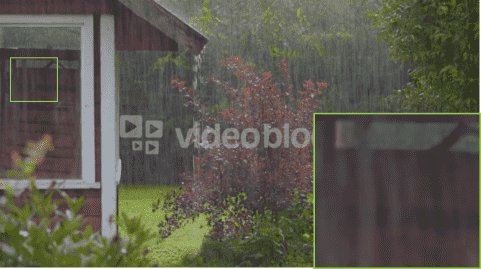}
		\end{minipage}
		\begin{minipage}[t]{0.15\linewidth}
			\centering
			\includegraphics[width=2.6cm]{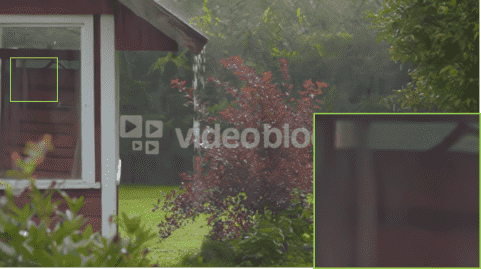}
		\end{minipage}
		\begin{minipage}[t]{0.15\linewidth}
			\centering
			\includegraphics[width=2.6cm]{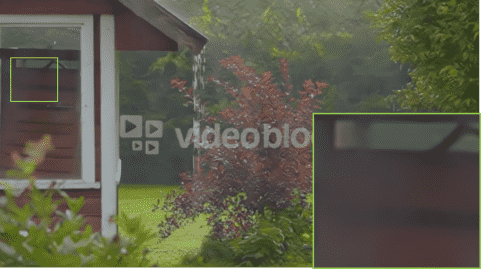}
		\end{minipage} \\ 
		
		\begin{minipage}[t]{0.15\linewidth}
			\centering
			\includegraphics[width=2.6cm]{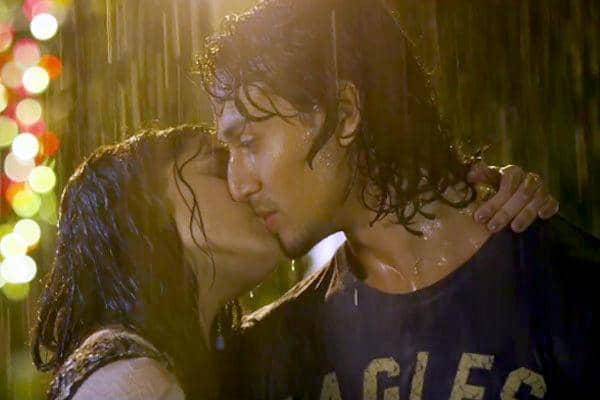}\vspace{4.0pt}
		\end{minipage} 
		\begin{minipage}[t]{0.15\linewidth}
			\centering
			\includegraphics[width=2.6cm]{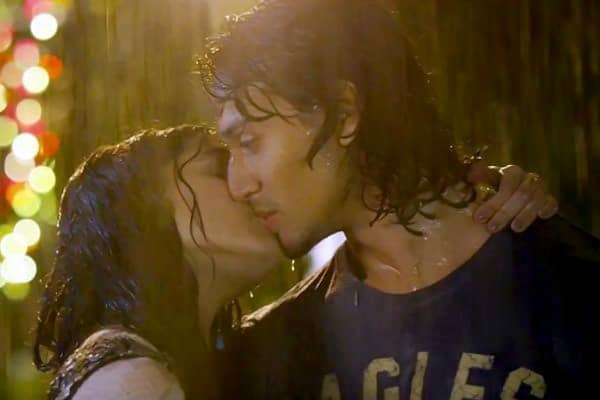}
		\end{minipage}
		\begin{minipage}[t]{0.15\linewidth}
			\centering
			\includegraphics[width=2.6cm]{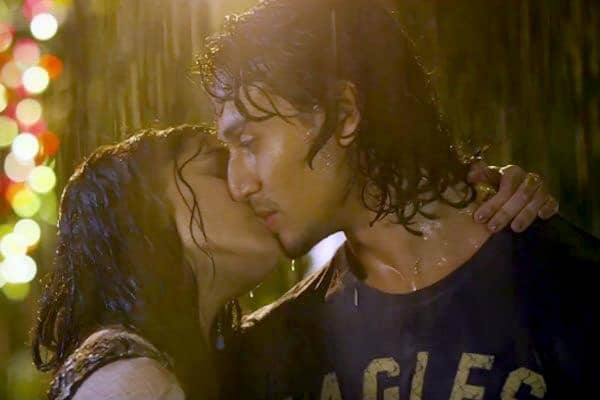}
		\end{minipage}
		\begin{minipage}[t]{0.15\linewidth}
			\centering
			\includegraphics[width=2.6cm]{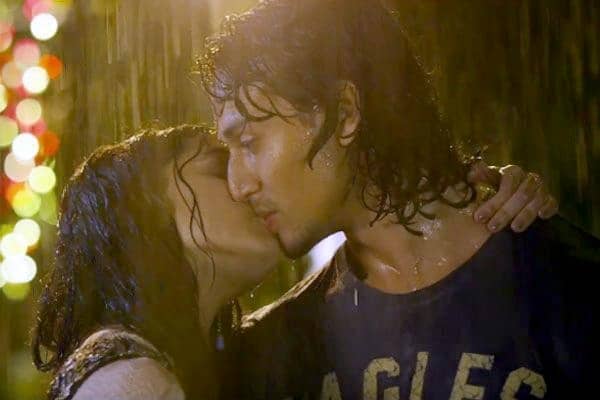}
		\end{minipage}
		\begin{minipage}[t]{0.15\linewidth}
			\centering
			\includegraphics[width=2.6cm]{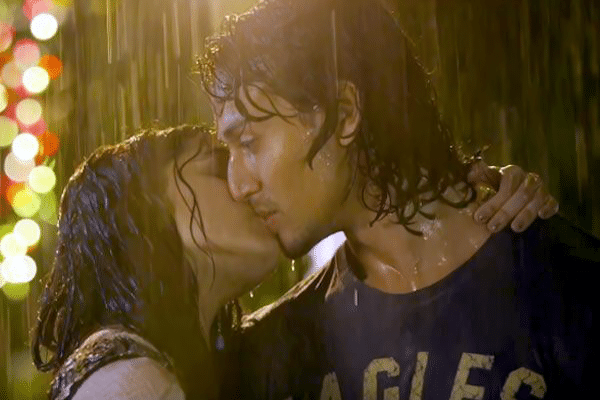}
		\end{minipage} 
		\begin{minipage}[t]{0.15\linewidth}
			\centering
			\includegraphics[width=2.6cm]{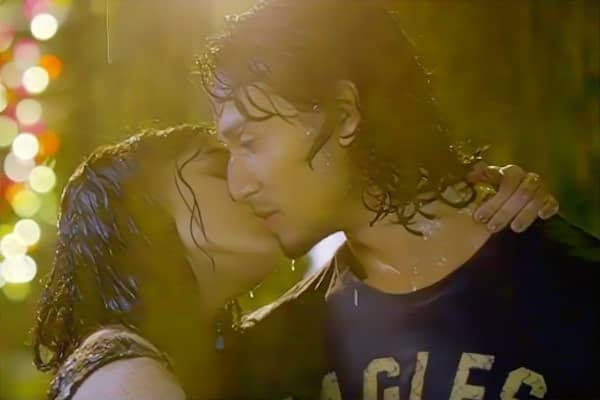}
		\end{minipage}\\ 
		\begin{minipage}[t]{0.15\linewidth}
			\centering
			\includegraphics[width=2.6cm]{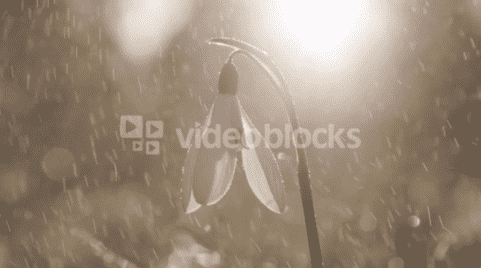}\vspace{4.0pt}
		\end{minipage} 
		\begin{minipage}[t]{0.15\linewidth}
			\centering
			\includegraphics[width=2.6cm]{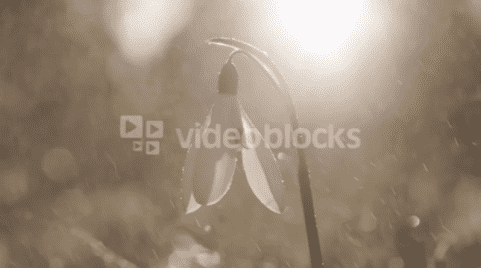}
		\end{minipage}
		\begin{minipage}[t]{0.15\linewidth}
			\centering
			\includegraphics[width=2.6cm]{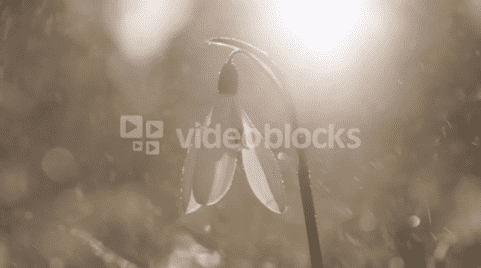}
		\end{minipage}
		\begin{minipage}[t]{0.15\linewidth}
			\centering
			\includegraphics[width=2.6cm]{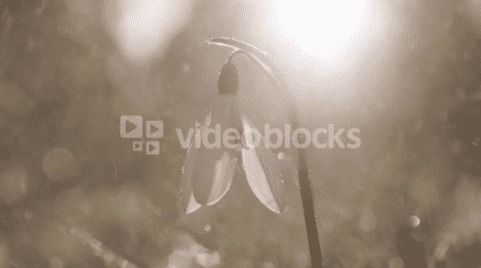}
		\end{minipage}
		\begin{minipage}[t]{0.15\linewidth}
			\centering
			\includegraphics[width=2.6cm]{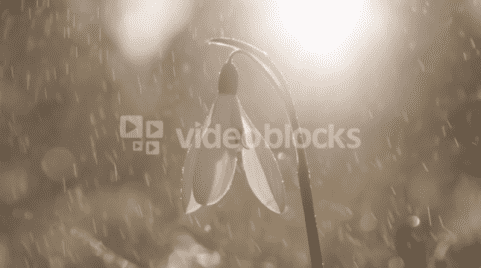}
		\end{minipage}
		\begin{minipage}[t]{0.15\linewidth}
			\centering
			\includegraphics[width=2.6cm]{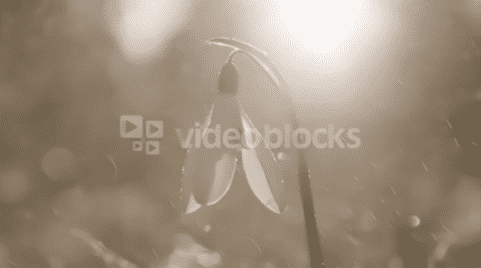}
		\end{minipage} \\ 
		\begin{minipage}[t]{0.15\linewidth}
			\centering
			\includegraphics[width=2.6cm]{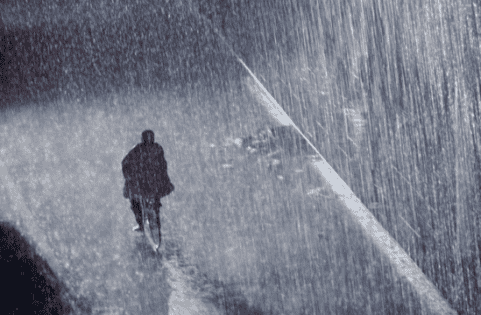}\vspace{4.0pt}
		\end{minipage}
		\begin{minipage}[t]{0.15\linewidth}
			\centering
			\includegraphics[width=2.6cm]{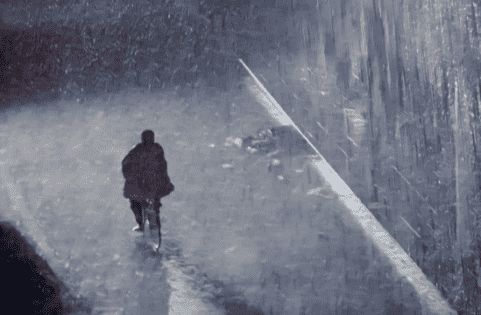}
		\end{minipage}
		\begin{minipage}[t]{0.15\linewidth}
			\centering
			\includegraphics[width=2.6cm]{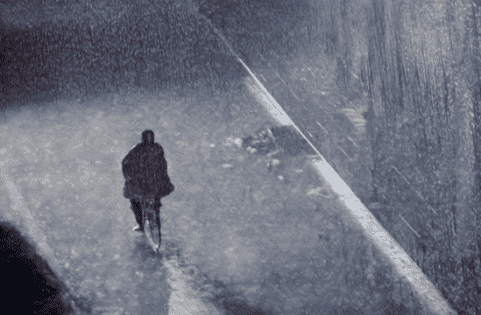}
		\end{minipage}
		\begin{minipage}[t]{0.15\linewidth}
			\centering
			\includegraphics[width=2.6cm]{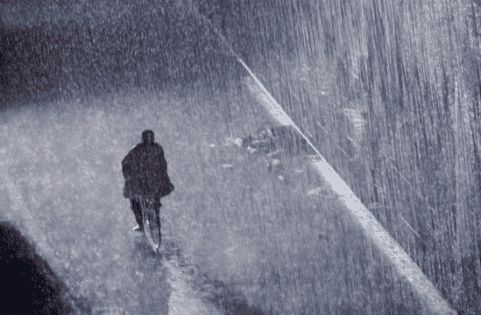}
		\end{minipage}
		\begin{minipage}[t]{0.15\linewidth}
			\centering
			\includegraphics[width=2.6cm]{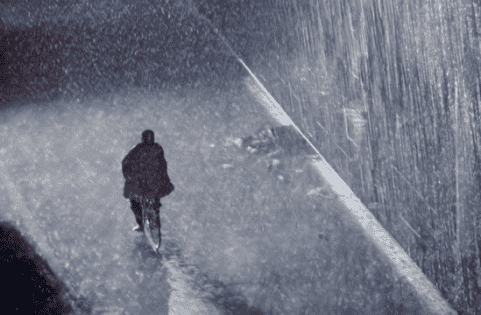}
		\end{minipage}
		\begin{minipage}[t]{0.15\linewidth}
			\centering
			\includegraphics[width=2.6cm]{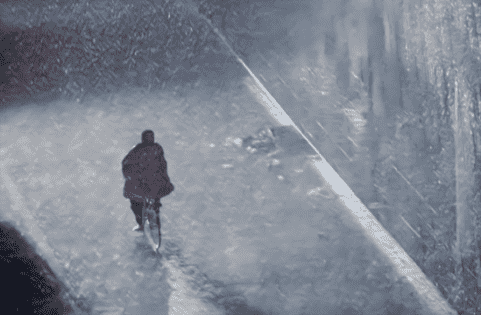}
		\end{minipage} \\
		\begin{minipage}[t]{0.15\linewidth}
			\centering
			\includegraphics[width=2.6cm]{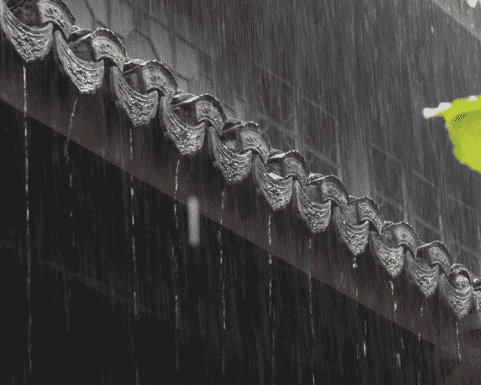}\vspace{4.0pt}
		\end{minipage}
		\begin{minipage}[t]{0.15\linewidth}
			\centering
			\includegraphics[width=2.6cm]{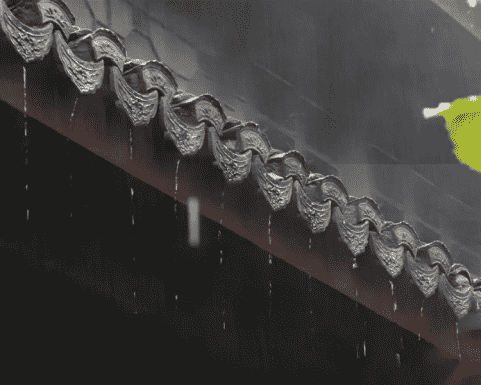}
		\end{minipage}
		\begin{minipage}[t]{0.15\linewidth}
			\centering
			\includegraphics[width=2.6cm]{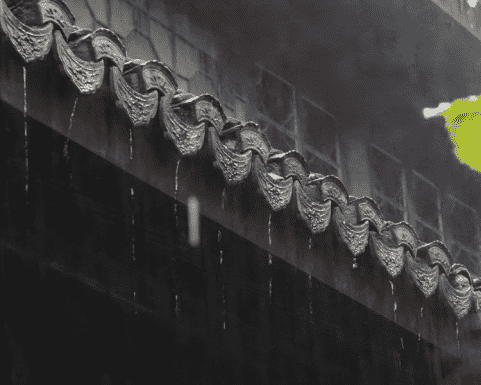}
		\end{minipage}
		\begin{minipage}[t]{0.15\linewidth}
			\centering
			\includegraphics[width=2.6cm]{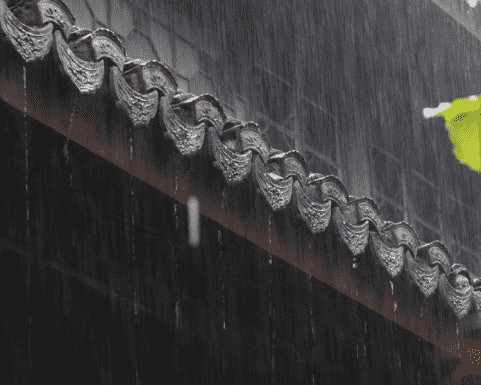}
		\end{minipage}
		\begin{minipage}[t]{0.15\linewidth}
			\centering
			\includegraphics[width=2.6cm]{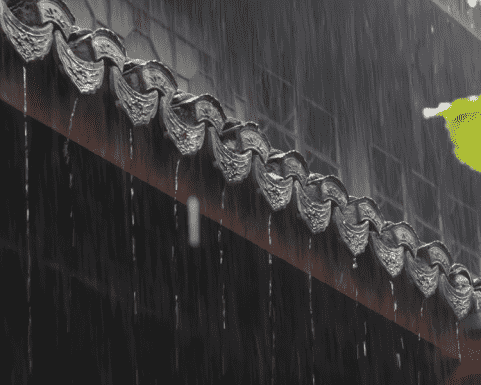}
		\end{minipage}
		\begin{minipage}[t]{0.15\linewidth}
			\centering
			\includegraphics[width=2.6cm]{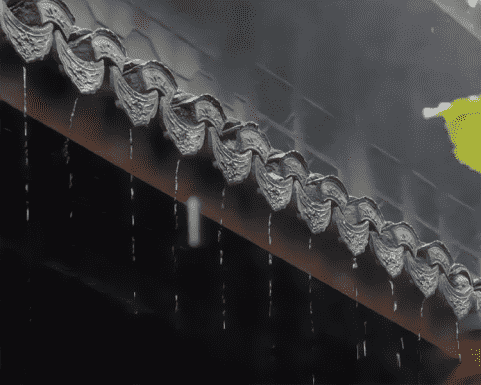}
		\end{minipage} \\
		
		\begin{minipage}[t]{0.15\linewidth}
			\centering
			\includegraphics[width=2.6cm]{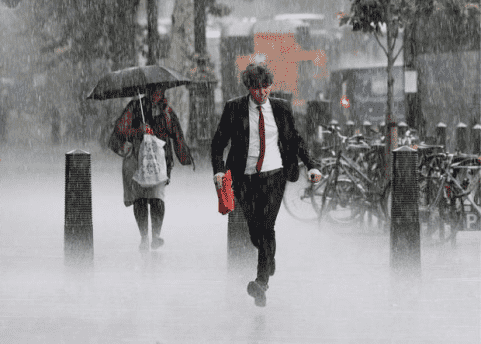}
			\footnotesize{Rainy}
		\end{minipage}
		\begin{minipage}[t]{0.15\linewidth}
			\centering
			\includegraphics[width=2.6cm]{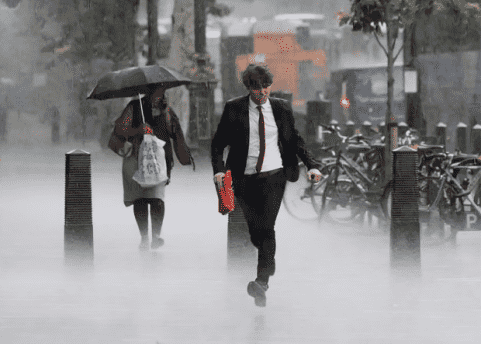}
			\footnotesize{ReHEN \cite{Yang2019acm}}
		\end{minipage}
		\begin{minipage}[t]{0.15\linewidth}
			\centering
			\includegraphics[width=2.6cm]{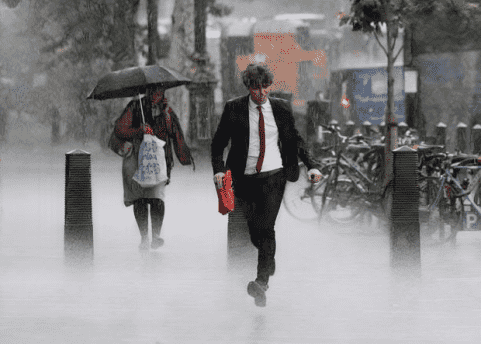}
			\footnotesize{DCSFN \cite{DCSFN}}
		\end{minipage}
		\begin{minipage}[t]{0.15\linewidth}
			\centering
			\includegraphics[width=2.6cm]{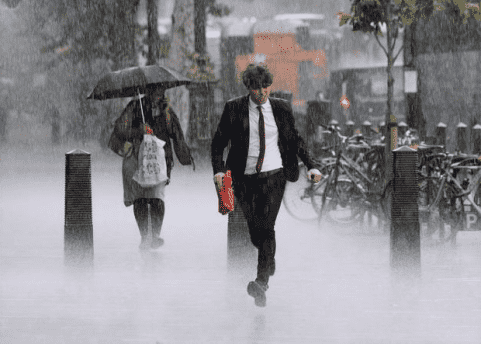}
			\footnotesize{BRN \cite{MPRNet}}
		\end{minipage}
		\begin{minipage}[t]{0.15\linewidth}
			\centering
			\includegraphics[width=2.6cm]{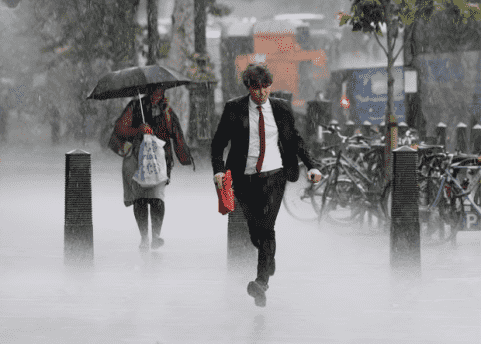}
			\footnotesize{MPRNet~\cite{MPRNet}}
		\end{minipage}
		\begin{minipage}[t]{0.15\linewidth}
			\centering
			\includegraphics[width=2.6cm]{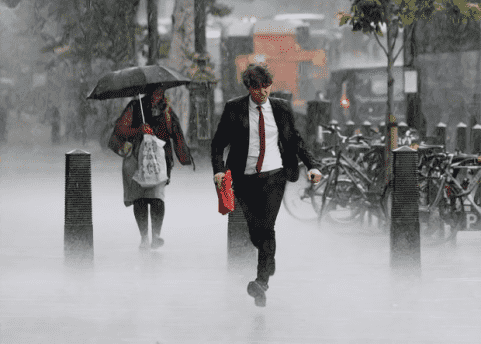}
			\footnotesize{MSResNet-TRNR (ours)}
		\end{minipage} \\
		\caption{Real-world image rain removal examples of ReHEN~\cite{Yang2019acm}, DCSFN \cite{DCSFN}, BRN~\cite{BRN2020TIP}, MPRNet$^{\dagger}$ \cite{MPRNet} and MSResNet-TRNR.}
		\label{fig:real-rain}
	\end{figure*}

\noindent\textbf{Image Rain Removal. } We carry out experiments to verify the effectiveness of proposed TRNR approach. In practice, we train MSResNet-TRNR separately on Rain100L-S, Rain100H-S, Rain800-S, RainLight-S, RainHeavy-S, and Rain1400-S. The MSResNet-TRNR is compared with prior based method JCAS \cite{Gu2017JCAS} and learning methods DerainNet~\cite{clearingthesky}, DDN \cite{Fu2017DDN}, JORDER \cite{Yang2017Jorder}, RESCAN \cite{Xia2018Rescan}, DID-MDN \cite{Zhang2018DID}, DualCNN~\cite{DualCNN}, PReNet \cite{PreNet}, ReHEN \cite{Yang2019acm}, JORDER-E~\cite{JORDERE}, DCSFN \cite{DCSFN}, JDNet \cite{JDNet}, BRN~\cite{BRN2020TIP}, MPRNet \cite{MPRNet}, ADN-64~\cite{ADN}, RLNet~\cite{RLNet}, and JDDGD~\cite{JDDGD}. Due to a large number of parameters ($>$10M) in MPRNet, we use the official pre-trained MPRNet\footnote{\href{https://github.com/swz30/MPRNet}{MPRNet official sources.}} denoted as MPRNet$^{\dagger}$ in this paper for evaluation, which is trained with 13712 examples gathered from multiple datasets. These compared methods except JCAS \cite{Gu2017JCAS} are all trained on large labeled datasets. The quantitative results are provided in Table \ref{tab: de-rain comparison}. From Table \ref{tab: de-rain comparison}, we find that the MSResNet-TRNR outperforms the prior-based method JCAS, and learning-based methods DDN, JORDER, RESCAN, PReNet, and JORDER-E over all compared datasets. Further, MSResNet-TRNR has obtained the first place on Rain100L, and the second place on Rain800, and RainLight when compared with all listed methods. Fig. \ref{fig:derain100L} and Fig. \ref{fig:derain800} have provided 4 image rain removal examples of DDN, DCSFN, MPRNet$^{\dagger}$ and MSResNet-TRNR. Although MSResNet-TRNR has not shown outstanding results on Rain100H, Rain800, and Rain1400 datasets in Table \ref{tab: de-rain comparison}, MSResNet-TRNR has achieved better performance than ReHEN, DCSFN, BRN, and MPRNet$^{\dagger}$ on real-world images as shown in Fig. \ref{fig:real-rain}. 

We further train three recent outstanding models DCSFN, BRN, and ADN-64, highlighted in {\color{magenta}\textit{magenta}} in Table \ref{tab: de-rain comparison}, on small datasets listed in Table \ref{tab: TRNR datasets} according to the instructions in their original papers. 
Quantitative results are tabulated at the bottom of Table \ref{tab: de-rain comparison},  where we observe that DCSFN, BRN, and ADN-64 all suffer performance degradations when trained using small datasets. MSResNet-TRNR, on the other hand, has maintained the performances of MSResNet-RIS across all datasets while remarkably outperforming DCSFN, BRN, and ADN-64 when data is scarce. This observation affirms the superiority of TRNR strategy in data-starved scenarios.

\begin{table*}[h]
	\centering
	\small
	\renewcommand{\arraystretch}{1.2}
	\renewcommand\tabcolsep{0.4pt}
	\caption{Average PSNR/SSIM Metrics for Color Image Blind Gaussian Noise Removal on McMaster, Kodak, BSD68 and Urban100 datasets. Noise Levels are 15, 25, 35, and 50. the Best Results are \textbf{Bolded}, and the Second and Third Results are Highlighted in {\color{red}Red} and {\color{blue}Blue} Colors, Respectively.}
	\resizebox{\textwidth}{!}{
		\begin{tabular}{l c| c c c c c c c c | c c} 
			\bottomrule
			Datasets & \makecell[c]{Noise\\ Level} & BM3D\cite{dabovBM3D} & DnCNN\cite{KaiZhangDnCNN2017} & NLNet\cite{NLNet2017Left} & FFDNet\cite{Zhang2018FFDNet} & BUIFD\cite{BUIFD} & PaCNet\cite{PaCNet} & ADNet~\cite{ADNet} & SMNet-B~\cite{SMNet} & \makecell[c]{MSResNet-\\ RIS} & \makecell[c]{MSResNet-\\ TRNR}\\
			\bottomrule
			\multirow{4}*{McMaster} & 15 & 34.06/0.911 & 34.70/0.922 & 34.10/0.912 & 34.67/0.922& 34.37/0.917 & 34.24/0.916 & 34.93/0.925 & {\color{blue}35.04}/{\color{blue}0.928} & \textbf{35.15}/\textbf{0.929} & {\color{red}35.13}/{\color{red}0.929}\\
			& 25 & 31.66/0.870 & 32.38/0.887 & 31.72/0.872 & 32.37/0.887 & 32.18/0.883 & 32.18/0.884 & 32.56/0.891 & {\color{blue}32.76}/{\color{blue}0.896} & \textbf{32.84}/\textbf{0.897} & {\color{red}32.83}/{\color{red}0.897} \\
			& 35 & 29.92/0.829 & 30.83/0.857 & - & 30.83/0.855 & 30.69/0.853 & - & 31.00/0.862 & {\color{blue}31.24}/{\color{blue}0.869} &  \textbf{31.31}/\textbf{0.869} & {\color{red}31.30}/{\color{red}0.869}\\
			& 50 & 28.28/0.780 & 29.18/0.818 & 28.47/0.791 & 29.20/0.815 & 29.09/0.814 & 29.01/0.813 & 29.36/0.823 & {\color{blue}29.60}/{\color{blue}0.832} & \textbf{29.69}/\textbf{0.833} & {\color{red}29.68}/{\color{red}0.833}\\
			\hline
			\multirow{4}*{Kodak} & 15 & 34.28/0.915 & 34.62/0.922 & 34.53/0.920 & 34.64/0.922 & 34.64/0.921 & 34.69/0.922 & {\color{blue}34.77}/0.923 & 34.77/{\color{blue}0.924} & \textbf{35.00}/\textbf{0.926} & {\color{red}34.99}/{\color{red}0.926}\\
			& 25 & 31.68/0.867 & 32.13/0.879 & 31.97/0.876 & 32.14/0.878 & 32.17/0.879 & 32.14/0.878 & 32.27/0.881 & {\color{blue}32.34}/{\color{blue}0.884} & \textbf{32.53}/{\color{red}0.886} & {\color{red}32.52}/\textbf{0.887}\\
			& 35 & 29.90/0.821 & 30.56/0.842 & - & 30.57/0.841 & 30.61/0.842 & - & 30.69/0.845 & {\color{blue}30.81}/{\color{blue}0.850} & \textbf{30.98}/{\color{red}0.851} & {\color{red}30.97}/\textbf{0.853} \\
			& 50 & 28.23/0.767 & 28.97/0.797 & 28.71/0.789 & 28.98/0.795 & 29.02/0.797 & 28.96/0.794 & 29.10/0.800 & {\color{blue}29.24}/{\color{blue}0.807} & {\color{red}29.39}/{\color{red}0.808} & \textbf{29.39}/\textbf{0.810}\\
			\hline
			\multirow{4}*{BSD68} & 15 & 33.52/0.922 & 33.84/0.929 & 33.79/0.928 & 33.89/0.929 & 33.93/0.929 & 33.95/0.930 & \textbf{33.99}/{\color{blue}0.930} & {\color{blue}33.95}/\textbf{0.931} & {\color{red}33.96}/{\color{red}0.930} & 33.94/0.930\\
			& 25 & 30.71/0.867 & 31.20/0.883 & 31.06/0.880 & 31.23/0.883 & 31.28/0.883 & 31.21/0.882 & 31.31/0.885 & {\color{blue}31.35}/{\color{red}0.886} & \textbf{31.37}/{\color{blue}0.885} & {\color{red}31.35}/\textbf{0.886}\\
			& 35 & 28.89/0.816 & 29.58/0.842 & - & 29.59/0.841 & 29.65/0.843 & - & 29.67/0.845 & {\color{blue}29.75}/{\color{red}0.848} & \textbf{29.78}/{\color{blue}0.846} & {\color{red}29.76}/\textbf{0.848} \\
			& 50 & 27.14/0.753 & 27.96/0.791 & 27.71/0.784 & 27.98/0.789 & 28.02/0.792 & 27.92/0.788 & 28.05/0.794 & {\color{blue}28.15}/\textbf{0.800} &\textbf{28.19}/{\color{blue}0.797} & {\color{red}28.17}/{\color{red}0.799} \\
			\hline
			\multirow{4}*{Urban100} & 15 & 33.95/0.941 & 33.94/0.943 & 33.88/0.942 & 33.84/0.942 & 33.71/0.939 & 33.87/0.941 & {\color{blue}34.20}/{\color{blue}0.945} & 34.19/0.945 &\textbf{34.51}/\textbf{0.948} & {\color{red}34.48}/{\color{red}0.948}\\
			& 25 & 31.38/0.909 & 31.50/0.914 & 31.35/0.910 & 31.40/0.912 & 31.35/0.909 & 31.60/0.913 & 31.73/0.916 & {\color{blue}31.87}/{\color{blue}0.919} &\textbf{32.17}/\textbf{0.922} & {\color{red}32.14}/{\color{red}0.922}\\
			& 35 & 29.27/0.873 & 29.87/0.886 & - &  29.79/0.885 & 29.76/0.881 & - & 30.07/0.890 & {\color{blue}30.33}/{\color{blue}0.896} & \textbf{30.62}/\textbf{0.899} & {\color{red}30.57}/{\color{red}0.899}\\
			& 50 & 27.67/0.832 & 28.10/0.849 & 27.82/0.840 & 28.05/0.848 & 28.04/0.843 & 28.12/0.848 & 28.32/0.854 & {\color{blue}28.65}/{\color{blue}0.863} & \textbf{28.93}/\textbf{0.867} & {\color{red}28.88}/{\color{red}0.867}\\
			\bottomrule
	\end{tabular}}
	\label{tab: colored denoise}
\end{table*}

\begin{table*}[htbp]
	\centering
	\small
	\renewcommand{\arraystretch}{1.3}
	\renewcommand\tabcolsep{1.0pt}
	\caption{Average PSNR/SSIM Metrics for Grayscale Image Gaussian Noise Removal on Set12, BSD68, and Urban100. Noise Levels are 15, 25, 35, and 50. the Best Results are \textbf{Bolded}, and the Second and Third Results are Highlighted in {\color{red}Red} and {\color{blue}Blue} Colors, Respectively.}
			\begin{tabular}{l c| c c c c c c c | c c} 
				\bottomrule
				Datasets & \makecell[c]{Noise\\ Level} & BM3D\cite{dabovBM3D} & DnCNN\cite{KaiZhangDnCNN2017} & NLNet\cite{NLNet2017Left} & FFDNet\cite{Zhang2018FFDNet} & BUIFD\cite{BUIFD} & ADNet~\cite{ADNet} & SMNet-B~\cite{SMNet} & \makecell[c]{MSResNet- \\ RIS} & \makecell[c]{MSResNet-\\ TRNR}\\
				\bottomrule
				\multirow{4}*{Set12} & 15 & 32.37/0.895 & 32.86/0.903 & 32.73/0.901 & 32.75/0.903 & 32.50/0.897 & 32.98/0.905 & {\color{blue}32.99}/{\color{blue}0.905} & \textbf{33.07}/\textbf{0.906} & {\color{red}33.06}/{\color{red}0.906} \\
				& 25 & 29.97/0.850 & 30.48/0.863 & 30.31/0.857 & 30.43/0.864 & 30.19/0.856 & 30.58/0.865 & {\color{blue}30.68}/{\color{blue}0.867} & \textbf{30.73}/\textbf{0.868} & {\color{red}30.72}/{\color{red}0.868}\\
				& 35 & 28.40/0.813 & 28.93/0.830 & - & 28.91/0.831 & 28.66/0.821 & - & {\color{blue}29.17}/{\color{blue}0.836} & {\color{red}29.20}/{\color{red}0.837} & \textbf{29.20}/\textbf{0.837} \\
				& 50 & 26.50/0.757 & 27.28/0.787 & 27.02/0.779 & 27.30/0.790 & 27.02/0.776 & 27.37/0.791 & {\color{blue}27.56}/{\color{blue}0.797} & {\color{red}27.60}/{\color{red}0.797} & \textbf{27.60}/\textbf{0.798}\\
				\hline
				\multirow{4}*{BSD68} & 15 & 31.07/0.872 & 31.69/0.891 & 31.51/0.886 & 31.64/0.891 & 31.50/0.884 & {\color{blue}31.74}/{\color{blue}0.892} & 31.71/0.891 & \textbf{31.80}/\textbf{0.893} & {\color{red}31.80}/{\color{red}0.893} \\
				& 25 & 28.57/0.801 & 29.22/0.828 & 28.99/0.817 & 29.19/0.829 & 29.07/0.821 & 29.25/0.829 & {\color{blue}29.28}/{\color{blue}0.831} & {\color{red}29.33}/{\color{red}0.832} & \textbf{29.33}/\textbf{0.833}\\
				& 35 & 27.09/0.748 & 27.73/0.779 & - & 27.72/0.780 & 27.58/0.770 & - & {\color{blue}27.81}/{\color{blue}0.782} &{\color{red}27.85}/{\color{red}0.784} & \textbf{27.86}/\textbf{0.785}\\
				& 50 & 25.44/0.675 & 26.27/0.721 & 26.02/0.709 & 26.29/0.724 & 26.12/0.711 & 26.29/0.722 & {\color{blue}26.37}/{\color{blue}0.727} & {\color{red}26.41}/{\color{red}0.728} & {\textbf{26.42}}/\textbf{0.730}\\
				\hline
				\multirow{4}*{Urban100} & 15 & 32.32/0.921 & 32.71/0.928 & 32.60/0.926 & 32.40/0.927 & 31.92/0.912 & {\color{blue}32.87}/{\color{blue}0.931} & 32.80/0.930 & \textbf{33.11}/\textbf{0.934} & {\color{red}33.09}/{\color{red}0.933}\\
				& 25 & 29.69/0.877 & 30.13/0.889 & 29.93/0.884 & 29.90/0.888 & 29.47/0.870 & 30.24/0.892 & {\color{blue}30.36}/{\color{blue}0.894} & \textbf{30.60}/\textbf{0.899} & {\color{red}30.57}/{\color{red}0.898}\\
				& 35 & 27.89/0.836 & 28.40/0.853 & - & 28.25/0.853 & 27.82/0.831 & - & {\color{blue}28.75}/{\color{blue}0.862} & \textbf{28.94}/\textbf{0.867} & {\color{red}28.92}/{\color{red}0.866}\\
				& 50 & 25.65/0.766 & 26.54/0.801 & 26.17/0.791 & 26.50/0.805 & 26.06/0.779 & 26.64/0.807 &{\color{blue}27.00}/{\color{blue}0.818}& \textbf{27.16}/\textbf{0.823} & {\color{red}27.16}/{\color{red}0.822}\\
				\bottomrule
			\end{tabular} 
		\label{tab: grayscale denoise}
	\end{table*}
	
	\begin{figure*}[htbp]
		\centering
		\setlength{\abovecaptionskip}{0.01in}
		\begin{minipage}[t]{0.15\linewidth}
			\centering
			\includegraphics[width=2.6cm]{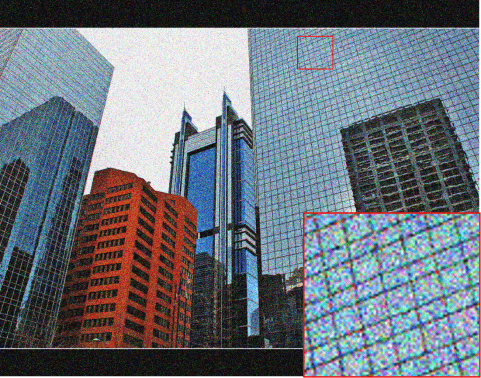}\vspace{-1pt}
			\footnotesize{14.15/0.265}\vspace{5pt}
		\end{minipage}
		\begin{minipage}[t]{0.15\linewidth}
			\centering
			\includegraphics[width=2.6cm]{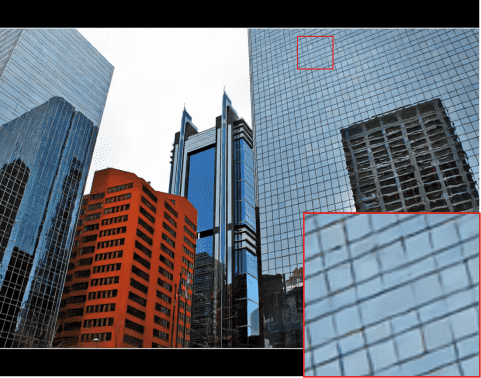}\vspace{-1pt}
			\footnotesize{25.95/0.817}\vspace{5pt}
		\end{minipage} 
		\begin{minipage}[t]{0.15\linewidth}
			\centering
			\includegraphics[width=2.6cm]{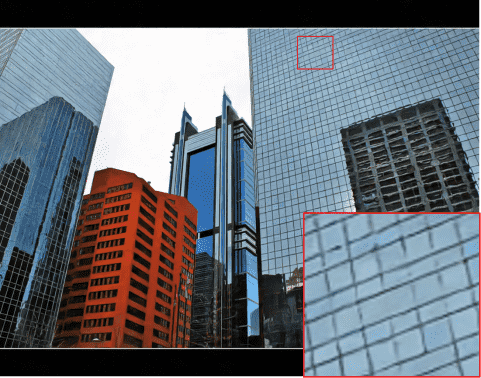}\vspace{-1pt}
			\footnotesize{25.68/0.775}\vspace{5pt}
		\end{minipage}
		\begin{minipage}[t]{0.15\linewidth}
			\centering
			\includegraphics[width=2.6cm]{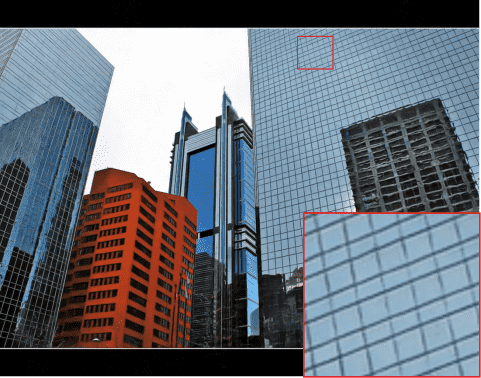}\vspace{-1pt}
			\footnotesize{25.97/0.814}\vspace{5pt}
		\end{minipage}
		\begin{minipage}[t]{0.15\linewidth}
			\centering
			\includegraphics[width=2.6cm]{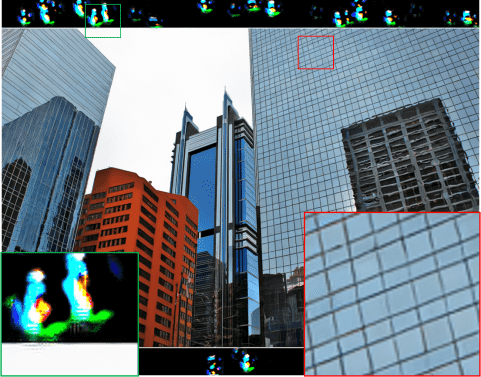}\vspace{-1pt}
			\footnotesize{20.75/0.802}\vspace{5pt}
		\end{minipage}
		\begin{minipage}[t]{0.15\linewidth}
			\centering
			\includegraphics[width=2.6cm]{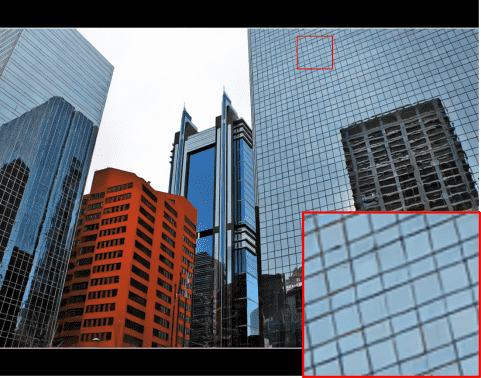}\vspace{-1pt}
			\footnotesize{\textbf{26.53/0.835}}\vspace{5pt}
		\end{minipage} 
		
		\begin{minipage}[t]{0.15\linewidth}
			\centering
			\includegraphics[width=2.6cm]{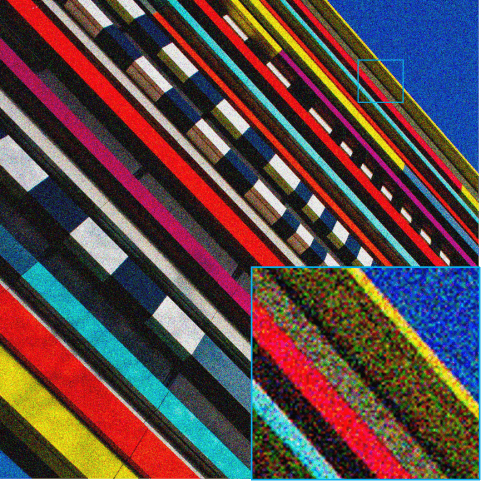}\vspace{-1pt}
			\footnotesize{14.16/0.134 \\Noisy}\vspace{5pt}
		\end{minipage}
		\begin{minipage}[t]{0.15\linewidth}
			\centering
			\includegraphics[width=2.6cm]{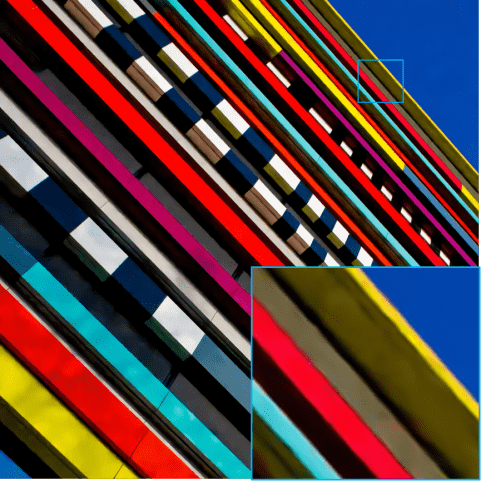}\vspace{-1pt}
			\footnotesize{31.97/0.876 \\ FFDNet~\cite{Zhang2018FFDNet}}\vspace{5pt}
		\end{minipage} 
		\begin{minipage}[t]{0.15\linewidth}
			\centering
			\includegraphics[width=2.6cm]{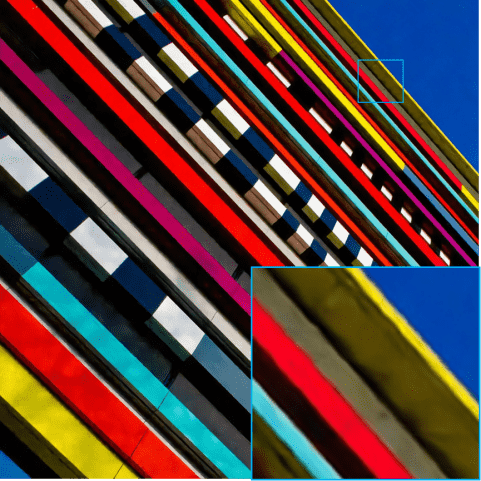}\vspace{-1pt}
			\footnotesize{31.13/0.834 \\ BUIFD~\cite{BUIFD}}\vspace{5pt}
		\end{minipage}
		\begin{minipage}[t]{0.15\linewidth}
			\centering
			\includegraphics[width=2.6cm]{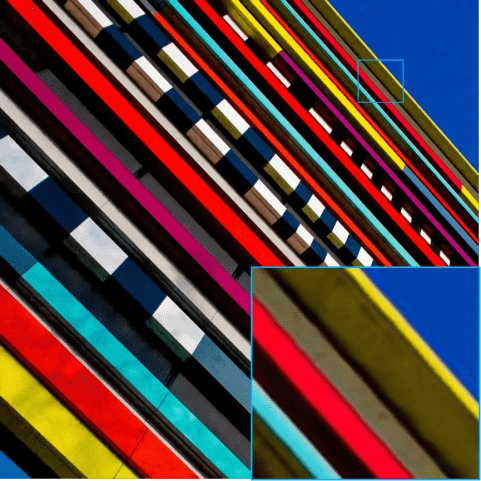}\vspace{-1pt}
			\footnotesize{31.60/0.814\\PaCNet~\cite{PaCNet}}\vspace{5pt}
		\end{minipage}
		\begin{minipage}[t]{0.15\linewidth}
			\centering
			\includegraphics[width=2.6cm]{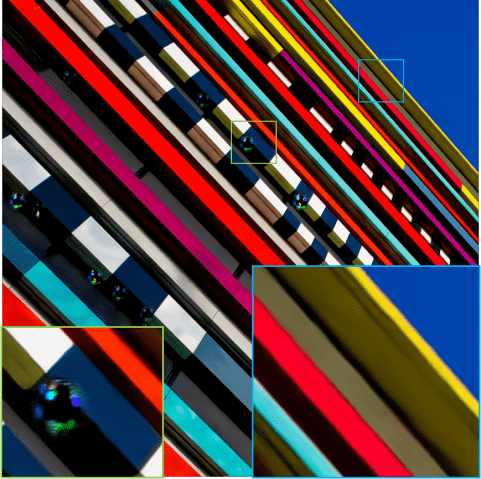}\vspace{-1pt}
			\footnotesize{29.36/0.894\\DNNet-RIS (Ours)}\vspace{5pt}
		\end{minipage}
		\begin{minipage}[t]{0.15\linewidth}
			\centering
			\includegraphics[width=2.6cm]{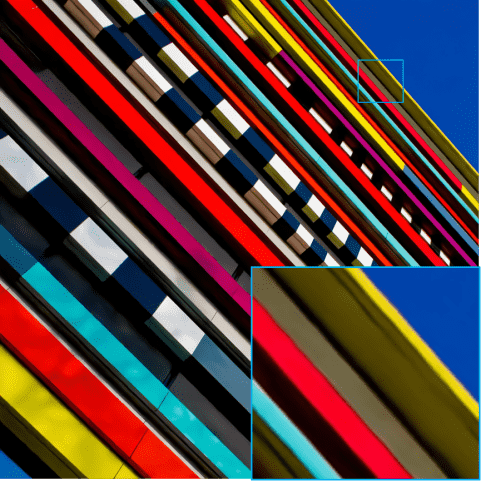}\vspace{-1pt}
			\footnotesize{\textbf{33.24}/\textbf{0.908}\\ DNNet-TRNR (Ours)}\vspace{5pt}
		\end{minipage} 
		\caption{Color image noise removal visual examples of FFDNet~\cite{Zhang2018FFDNet}, BUIFD~\cite{BUIFD}, PaCNet~\cite{PaCNet} and proposed DDNet-RIS, DDNet-TRNR on Urban100 dataset. Noise level is 50. Values at the bottom of every image indicate the PSNR/SSIM metrics respectively.}
		\label{fig:color-denoise}
	\end{figure*}
	
	\begin{figure*}[htbp]
		\centering
		\setlength{\abovecaptionskip}{0.01in}
		\begin{minipage}[t]{0.17\linewidth}
			\centering
			\includegraphics[width=2.8cm]{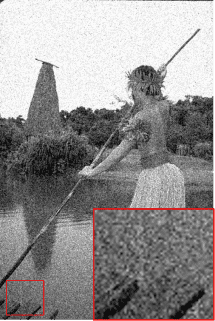}\vspace{-1pt}
			\footnotesize{20.17/0.294}\vspace{5pt}
		\end{minipage}
		\begin{minipage}[t]{0.17\linewidth}
			\centering
			\includegraphics[width=2.8cm]{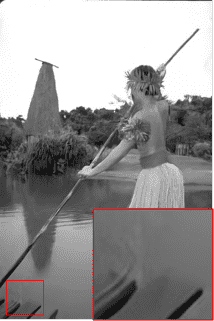}\vspace{-1pt}
			\footnotesize{30.23/0.874}\vspace{5pt}
		\end{minipage} 
		\begin{minipage}[t]{0.17\linewidth}
			\centering
			\includegraphics[width=2.8cm]{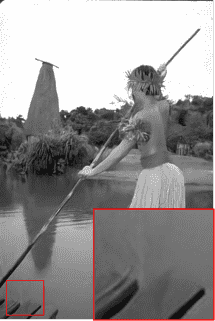}\vspace{-1pt}
			\footnotesize{30.17/0.867}\vspace{5pt}
		\end{minipage}
		\begin{minipage}[t]{0.17\linewidth}
			\centering
			\includegraphics[width=2.8cm]{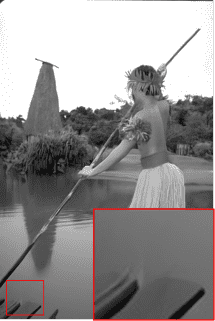}\vspace{-1pt}
			\footnotesize{\textbf{30.41}/\textbf{0.878}}\vspace{5pt}
		\end{minipage}
		\begin{minipage}[t]{0.17\linewidth}
			\centering
			\includegraphics[width=2.8cm]{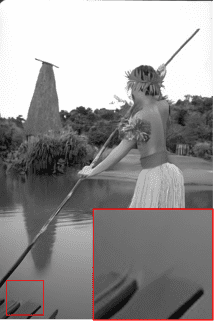}\vspace{-1pt}
			\footnotesize{30.37/0.878}\vspace{5pt}
		\end{minipage}
		
		\begin{minipage}[t]{0.17\linewidth}
			\centering
			\includegraphics[width=2.8cm]{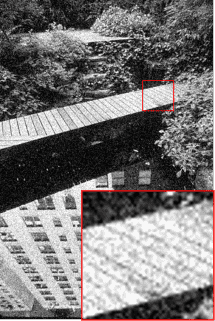}\vspace{-1pt}
			\footnotesize{20.19/0.574 \\Noisy}\vspace{5pt}
		\end{minipage}
		\begin{minipage}[t]{0.17\linewidth}
			\centering
			\includegraphics[width=2.8cm]{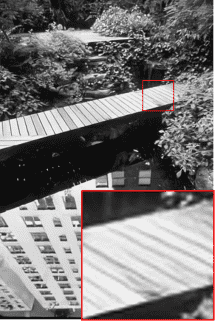}\vspace{-1pt}
			\footnotesize{26.38/0.871 \\ FFDNet~\cite{Zhang2018FFDNet}}\vspace{5pt}
		\end{minipage} 
		\begin{minipage}[t]{0.17\linewidth}
			\centering
			\includegraphics[width=2.8cm]{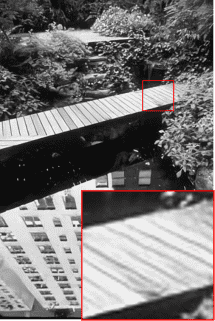}\vspace{-1pt}
			\footnotesize{26.35/0.869 \\ BUIFD~\cite{BUIFD}}\vspace{5pt}
		\end{minipage}
		\begin{minipage}[t]{0.17\linewidth}
			\centering
			\includegraphics[width=2.8cm]{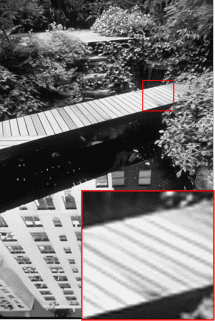}\vspace{-1pt}
			\footnotesize{\textbf{26.51}/\textbf{0.876}\\DNNet-RIS (Ours)}\vspace{5pt}
		\end{minipage}
		\begin{minipage}[t]{0.17\linewidth}
			\centering
			\includegraphics[width=2.8cm]{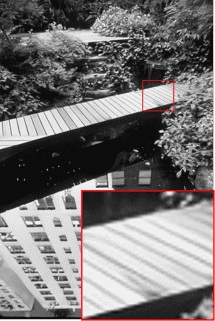}\vspace{-1pt}
			\footnotesize{26.48/0.875\\DNNet-TRNR (Ours)}\vspace{5pt}
		\end{minipage}
		\caption{Grayscale image noise removal visual examples of FFDNet~\cite{Zhang2018FFDNet}, BUIFD~\cite{BUIFD}, SMNet-B~\cite{SMNet}, and proposed MSResNet-RIS, MSResNet-TRNR on BSD68 dataset. Noise level is 25. Values at the bottom of every image indicate the PSNR/SSIM metrics respectively.}
		\label{fig:gray-denoise}
	\end{figure*}
	
	\noindent\textbf{Color Image Gaussian Noise Removal. } To further investigate the proposed TRNR for image Gaussian noise removal, we compare MSResNet with recent prior-based method BM3D \cite{dabovBM3D}, as well as learning-based methods DnCNN \cite{KaiZhangDnCNN2017}, NLNet \cite{NLNet2017Left}, FFDNet \cite{Zhang2018FFDNet}, BUIFD~\cite{BUIFD}, PaCNet~\cite{PaCNet}, ADNet~\cite{ADNet}, and SMNet-B~\cite{SMNet} on color image Gaussian noise removal with noise levels set to 15, 25, 35 and 50. In practice, we train MSResNet-RIS on WaterlooBSD and MSResNet-TRNR on WaterlooBSD-S. The quantitative results are presented in Table \ref{tab: colored denoise}. The PSNR/SSIM metrics for NLNet and PaCNet on noise level 35 are vacant due to lack of corresponding pretrained models. According to Table \ref{tab: colored denoise}, MSResNet-RIS has achieved the \textit{state-of-the-art} performance on McMaster, Kodak, BSD68, and Urban100. Furthermore, MSResNet-TRNR trained using WaterlooBSD-S (9.7\% of WaterlooBSD) becomes comparable to MSResNet-RIS. Fig. \ref{fig:color-denoise} provides 2 noise removal examples in Urban100 dataset when noise level equals 50. The compared methods in Fig. \ref{fig:color-denoise} are FFDNet, BUIFD, PaCNet, MSResNet-RIS and MSResNet-TRNR. As shown in Fig. \ref{fig:color-denoise}, MSResNet-RIS and MSResNet-TRNR can better restore textures and remove noise.
	
	\noindent\textbf{Grayscale Image Gaussian Noise Removal. }To demonstrate TRNR's effectiveness in grayscale image noise removal, we compare MSResNet with the prior-based method BM3D \cite{dabovBM3D}, as well as learning-based methods DnCNN \cite{KaiZhangDnCNN2017}, NLNet \cite{NLNet2017Left}, FFDNet \cite{Zhang2018FFDNet}, BUIFD~\cite{BUIFD}, ADNet~\cite{ADNet}, and SMNet-B~\cite{SMNet}. We train MSResNet-RIS on WaterlooBSD and MSResNet-TRNR on WaterlooBSD-S. And then we use Set12, BSD68, and Urban100 for grayscale image Gaussian noise removal evaluation with noise levels set to 15, 25, 35, and 50. Table \ref{tab: grayscale denoise} reports the grayscale image noise removal results. Both MSResNet-RIS and MSResNet-TRNR have shown the best performance across all datasets and noise levels. As for MSResNet-TRNR trained using 9.7\% of WaterlooBSD, it has outperformed recent extraordinary methods, e.g. DnCNN, FFDNet, BUIFD, ADNet, and SMNet-B.
	It is worth noticing that MSResNet-TRNR has surpassed MSResNet-RIS when noise levels are 35, and 50 on Set12 and BSD68 datasets, which demonstrates the potential of MSResNet-TRNR in restoring heavily degraded images. We present 2 visual examples of grayscale noise removal in Fig. \ref{fig:gray-denoise} to show the superiority of TRNR. The results on the first row of Fig. \ref{fig:gray-denoise} demonstrate that MSResNet-RIS and MSResNet-TRNR can remove noise from images better than FFDNet and BUIFD. While the results in the second row illustrate that MSResNet-RIS and MSResNet-TRNR can recover more details.
	
	\subsection{Comparisons on Real-World Dataset} 
	Although the proposed MSResNet trained utilizing the TRNR has shown great performance on synthetic datasets while requiring fewer images for training, the generalization ability of MSResNet on real-world images is still unknown. Therefore, we compare MSResNet-TRNR with recent image rain removal methods on real-world rainy images collected from \cite{Yang2017Jorder,Wang2019SPAnet}, and \cite{CGAN}. MSResNet-TRNR tested on real-world image rain removal is trained on Rain800-S.
	Fig. \ref{fig:real-rain} provides 7 real-word image rain removal examples. As can be seen from Fig. \ref{fig:real-rain}, MSResNet-TRNR can better remove rain streaks and recover background compared with ReHEN, DCSFN, BRN, and MPReNet. Note that MSResNet-TRNR is trained on Rain800-S with only 280 images. Hence, 
	the real-world image rain removal results prove that MSResNet-TRNR can generalize well on real-world occasions.
	
	\section{Conclusion}
	\label{sec: conclusion}
	Recent deep learning methods for image rain and noise removal require large labeled datasets for training. To alleviate the reliance on large labeled datasets, we have proposed the task-driven image rain and noise removal (TRNR) by introducing the patch analysis strategy. In TRNR, we first use the patch analysis strategy to improve the utilization of image patches, then we propose the N-frequency-K-shot learning tasks to enable the deep models to learn from abundant tasks when data is scarse. In this paper, we have conducted sufficient experiments on image rain removal and Gaussian noise removal to validate the proposed TRNR. Experiments on both synthetic datasets and real-world datasets have revealed the superiority and effectiveness of TRNR when training data is limited. It is worth noting that there is still room for further study of TRNR. One direction would be to employ TRNR in real scenarios where we can only collect a few examples and apply TRNR to other image restoration tasks, e.g. image super-resolution, and image dehazing. Another direction is building N-frequency-K-shot learning tasks by dynamically clustering image patches from data batches instead of pre-clustering all image patches. Because clustering $N^{(p)}$ image patches into $C$ clusters requires $O(CN^{(p)})$ time complexity, which is prohibitively expensive when $N^{(p)}$ is large. Furthermore, TRNR makes it possible to learn a unified model for a variety of image processing tasks due to a task-driven learning strategy. Finally, TRNR can also be improved by incorporating superior meta-learning algorithms.
	
	\appendix[a Theoretical Analysis of the Proposed TRNR]
	In this section, we present a theoretical analysis of the proposed TRNR to show that TRNR can improve the generalization ability of models. Following the notations in \S\ref{subsec: task-driven}, let $\mathcal{L}(\theta, \mathcal{D})$ be the loss function where we omit $f$ for simplicity, and $\mathcal{D}^{(train)}_{\tau_j}, \mathcal{D}^{(val)}_{\tau_j}, \mathcal{D}^{(test)}_{\tau_j}$ be the training set, validation set and testing set respectively of the N-frequency-K-shot learning task $\tau_j$.
	
	We first show that the data-driven training strategy cannot guarantee the generalization ability of models. Suppose we feed the neural network $f$ with a batch of training data $D^{(train)}_B$. Then by gradient descent, we obtain the updated parameters $\theta^{\prime}$ formulated in Eq. (\ref{eq: data-driven}) with lower loss value on $D^{(train)}_B$, where $\alpha$ is a step size hyperparameter.
	
	\begin{equation}
		\theta^{\prime} = \theta-\alpha\frac{\partial\mathcal{L}(\theta, D^{(train)}_B)}{\partial \theta}=\theta-\alpha \mathbf{g}^{(train)}_B
		\label{eq: data-driven}
	\end{equation}
	
	Eq. (\ref{eq: data-driven}) indicates that $\theta^{\prime}$ is independent of $D^{(val)}$ and $D^{(test)}$, therefore the loss value on validation set and testing set can only rely on the \emph{i.i.d} hypothesis on dataset. For example, we can take a first-order approximation on $\mathcal{L}(\theta^{\prime}, \mathcal{D}^{(val)})$ in Eq. (\ref{eq: first-order}), where $\mathbf{g}^{(val)}$ means the gradients of $\theta$ on validation set. It can be seen that the validation loss decreases when $\mathbf{g}^{(train)}_B$ and $\mathbf{g}^{(val)}$ are in the same direction (which is true when \emph{i.i.d} hypothesis holds), and it is the same case on the testing set. Since $\mathcal{D}^{(val)}$ and $\mathcal{D}^{(test)}$ do not contribute to $\mathbf{g}^{(train)}_{B}$ in Eq. (\ref{eq: data-driven}), the reduction of validation loss and testing loss is only related to the \emph{i.i.d} hypothesis.
	
	\begin{equation}
		\mathcal{L}(\theta^{\prime}, \mathcal{D}^{(val)})\approx \mathcal{L}(\theta, \mathcal{D}^{(val)})-\alpha (\mathbf{g}^{(train)}_B)^{T}\mathbf{g}^{(val)}
		\label{eq: first-order}
	\end{equation}
	
	As for the proposed TRNR, we first perform gradient descent on the training set of sampled $R$ N-frequency-K-shot tasks. As shown in Eq. (\ref{eq: gradient-descents}), we obtain $\theta_j$ for task $\tau_j$ by performing gradient descent on $\mathcal{D}^{(train)}_{\tau_j}$. 
	
	\begin{equation}
		\theta_j = \theta-\alpha\frac{\partial\mathcal{L}(\theta, \mathcal{D}^{(train)}_{\tau_j})}{\partial \theta}=\theta-\alpha \mathbf{g}^{(train)}_j
		\label{eq: gradient-descents}
	\end{equation}
	
	Then, in order to make a significant update to $\theta$, TRNR computes the gradients of $\theta_j$ as $\mathbf{g}^{mul}_{j}$ on the validation set $D^{(val)}_{\tau_j}$ following
	
	\begin{equation}
		\mathbf{g}^{mul}_{j}=\frac{\partial\mathcal{L}(\theta_j, \mathcal{D}^{(val)}_{\tau_j})}{\partial \theta}=\frac{\partial \mathcal{L}(\theta-\alpha\mathbf{g}^{(train)}_{j}, \mathcal{D}^{(val)}_{\tau_j})}{\partial \theta}.
		\label{eq: multi-step}
	\end{equation} 
	
	Finally, By combining $\mathbf{g}_j^{mul}$ from all $R$ tasks, TRNR updates $\theta$ to $\theta^{\prime}$ following Eq. (\ref{eq: sig-update})
	
	\begin{equation}
		\theta^{\prime} = \theta - \beta\frac{1}{R}\sum_j\mathbf{g}^{mul}_{j},
		\label{eq: sig-update}
	\end{equation} 
	
	\noindent where $\beta$ is a step size hyperparameter. As presented in Eq. (\ref{eq: approx-gmul}), we take an approximation of $\mathbf{g}^{mul}_{j}$ in Eq. (\ref{eq: multi-step}), where $H_j^{(train)}$ means the Hessian matrix of $\theta$ on training set $\mathcal{D}^{(train)}_{\tau_j}$ and $\mathbf{g}^{(val)}_j$ indicates the gradients of $\theta$ on validation set $\mathcal{D}^{(val)}_{\tau_j}$.
	
	\begin{equation}
		\begin{split}
			\mathbf{g}^{mul}_{j}&= \frac{\partial\mathcal{L}(\theta_j, \mathcal{D}^{(val)}_{\tau_j})}{\partial \theta}=\frac{\partial \theta_j}{\partial \theta} \frac{\partial\mathcal{L}(\theta_j, \mathcal{D}_{\tau_j}^{(val)})}{\partial \theta_j} \\
			&=\left(\mathbf{1}-\alpha H_j^{(train)}\right)\frac{\partial\mathcal{L}(\theta_j, \mathcal{D}_{\tau_j}^{(val)})}{\partial \theta_j} \\
			&\approx \mathbf{g}^{(val)}_j -\alpha H_j^{(train)}\mathbf{g}^{(val)}_j
		\end{split}
		\label{eq: approx-gmul}
	\end{equation}
	
	By combining Eq. (\ref{eq: sig-update}) and Eq. (\ref{eq: approx-gmul}), we obtain an approximation of updated parameters $\theta^{\prime}$ in Eq. (\ref{eq: sig-update-approx}) 
	
	\begin{equation}
		\theta^{\prime} \approx \theta-\beta\frac1{R}\sum_j \mathbf{g}^{(val)}_j+\beta\frac1{R}\sum_j\alpha H^{(train)}_j\mathbf{g}^{(val)}_j.
		\label{eq: sig-update-approx}
	\end{equation}
	
	Note that Eq. (\ref{eq: sig-update-approx}) can be regarded as a gradient descent step with respect to a specific loss function $\mathcal{L}^{TRNR}(\theta, \mathcal{D}^{(train)}_{\tau_j}, \mathcal{D}^{(val)}_{\tau_j})$ as below
	
	\begin{align}
		\mathcal{L}^{TRNR}&=\frac1{R}\sum_j\mathcal{L}(\theta, \mathcal{D}^{(val)}_{\tau_j}) \notag \\
		& -\alpha\frac1{R}\sum_j (\mathbf{g}^{(train)}_j)^T\phi(\mathbf{g}^{(val)}_j),
		\label{eq: trnr-loss}
	\end{align}
	
	\noindent where $\phi(\cdot)$ means to remove the derivative on $\theta$. Eq. (\ref{eq: trnr-loss}) demonstrates that TRNR can explicitly decrease the loss value on validation sets, while simultaneously aligning the gradients on training sets and validation sets. Therefore, compared with the data-driven approach, TRNR can improve the generalization ability of models.
	
	\bibliographystyle{IEEEtran}
	\bibliography{references.bib}
	\vfill
	
\end{document}